\documentclass{IEEEtran}
\usepackage{cite}
\usepackage{amsmath,amssymb,amsfonts}
\usepackage{algorithmic}
\usepackage{graphicx}
\usepackage{bbm}
\usepackage{mathtools}
\usepackage{adjustbox}
\usepackage{multirow}
\usepackage{enumerate}
\usepackage{rotating}
\usepackage{makecell}
\usepackage{diagbox}
\usepackage{subcaption}
\usepackage{hyperref}
\usepackage{textcomp}
\usepackage{algorithm,algorithmic}
\newcommand{\etal}{\textit{et al}.}
\hypersetup{
    colorlinks=true,
    linkcolor=blue,
    filecolor=magenta,      
    urlcolor=cyan,
    pdftitle={Overleaf Example},
    pdfpagemode=FullScreen,
    }

\makeatletter
\newcommand\notsotiny{\@setfontsize\notsotiny\@vipt\@viipt}
\makeatother

\def\BibTeX{{\rm B\kern-.05em{\sc i\kern-.025em b}\kern-.08em
    T\kern-.1667em\lower.7ex\hbox{E}\kern-.125emX}}
\begin{document}
\title{Improving Network Slimming with Nonconvex Regularization}
\author{Kevin Bui,
Fredrick Park, Shuai Zhang, Yingyong Qi, and Jack Xin}

\maketitle

\begin{abstract}
Convolutional neural networks (CNNs) have developed to become powerful models for various computer vision tasks ranging from object detection to semantic segmentation. However, most of the state-of-the-art CNNs cannot be deployed directly on edge devices such as smartphones and drones, which need low latency under limited power and memory bandwidth. 
One popular, straightforward approach to compressing CNNs is network slimming, which imposes $\ell_1$ regularization on the channel-associated scaling factors via the batch normalization layers during training. Network slimming thereby identifies insignificant channels that can be pruned for inference. In this paper, we propose replacing the $\ell_1$ penalty with an alternative nonconvex, sparsity-inducing penalty in order to yield a more compressed and/or accurate CNN architecture. We investigate $\ell_p (0 < p < 1)$, transformed $\ell_1$ (T$\ell_1$), minimax concave penalty (MCP), and smoothly clipped absolute deviation (SCAD) due to their recent successes and popularity in solving sparse optimization problems, such as compressed sensing and variable selection. We demonstrate the effectiveness of network slimming with nonconvex penalties on three neural network architectures -- VGG-19, DenseNet-40, and ResNet-164 -- on standard image classification datasets. Based on the numerical experiments, T$\ell_1$ preserves model accuracy against channel pruning, $\ell_{1/2, 3/4}$ yield better compressed models with similar accuracies after retraining as $\ell_1$, and MCP and SCAD provide more accurate models after retraining with similar compression as $\ell_1$. Network slimming with T$\ell_1$ regularization also outperforms the latest Bayesian modification of network slimming in compressing a CNN architecture in terms of memory storage while preserving its model accuracy after channel pruning.
\end{abstract}

\section{Introduction}
\label{sec:introduction}
In the past years, convolutional neural networks (CNNs) have evolved into superior models for various computer vision tasks, such as image classification~\cite{he2016deep,krizhevsky2012imagenet,simonyan2014very}, image segmentation~\cite{chen2017deeplab,long2015fully,ronneberger2015u}, and object detection \cite{girshick2014rich, huang2017speed, ren2015faster}. Unfortunately, training a highly accurate CNN is computationally demanding. State-of-the-art CNNs such as ResNet~\cite{he2016deep} can have up to at least a hundred layers and thus require millions of parameters to train and billions of floating-point-operations to execute. Consequently, deploying CNNs in low-memory devices, such as mobile smartphones, is difficult, making their real-world applications limited. 

To make CNNs more practical, many works suggest several different directions to compress large CNNs or to learn smaller, more efficient models from scratch.  Low-rank approximation~\cite{denton2014exploiting,jaderberg2014speeding,wen2017coordinating,xu2018trained,xu2020trp} minimizes network redundancy by approximating the network's weight matrices with low-rank matrices. Weight quantization ~\cite{chen2015compressing,courbariaux2015binaryconnect,li2016ternary,zhu2016trained,yin2018binaryrelax} replaces the floating-point weights with quantized weights, such as binary weights $\{-1,+1\}$ and ternary weights $\{-1, 0, +1\}$. Pruning~\cite{aghasi2017net,han2015learning,li2016pruning,hu2016network} determines which weights, filters, and/or channels are unnecessary and removes them from the network. Lastly, another popular direction is to sparsify the CNN while training it~\cite{alvarez2016learning,changpinyo2017power,scardapane2017group,wen2016learning}. Sparsity can be imposed on various types of structures existing in CNNs, such as filters and channels~\cite{wen2016learning}. 

One interesting yet straightforward approach in sparsifying CNNs is \textit{network slimming}~\cite{liu2017learning}. This method imposes $\ell_1$ regularization on the scaling factors in the batch normalization layers. Due to $\ell_1$ regularization, scaling factors corresponding to insignificant channels are pushed towards zeroes, narrowing down the important channels to retain, while the CNN model is being trained. Once the insignificant channels are pruned, the compressed model may need to be retrained since pruning can degrade its original accuracy. Overall, network slimming yields a compressed model with low run-time memory and number of computing operations. Since its inception, network slimming helps develop lightweight CNNs for various image classification tasks, such as traffic sign classification \cite{zhang2020lightweight}, facial expression recognition \cite{ma2021lightweight}, and semantic segmentation. \cite{he2021cap}.

To improve the performance of network slimming, we propose replacing $\ell_1$ regularization with an alternative regularization that promotes better sparsity and/or accuracy. Typically, better sparsity-promoting regularizers are nonconvex. Hence,  we examine the $\ell_{p}$ penalty~\cite{chartrand2007exact,chartrand2008iteratively,xu2012l}, transformed $\ell_1$ (T$\ell_1$) penalty~\cite{zhang2014minimization,zhang2018minimization}, the minimax concave penalty (MCP) \cite{zhang2010nearly}, and the smoothly clipped absolute deviation (SCAD) penalty \cite{fan2001variable} due to their recent successes and popularity.  These four regularizers have explicit formulas for their subgradients, which allow us to directly perform subgradient descent~\cite{shor2012minimization} when training CNNs. 

Preliminary work in the conference version \cite{bui2020nonconvex} of this paper demonstrated that T$\ell_1$ regularization preserves the CNN's accuracy after pruning, and $\ell_p$ regularization yields a more compressed CNN than $\ell_1$ with similar accuracy after retraining. This extended work includes discussion on the application of MCP and SCAD as additional regularization options for network slimming. Moreover, we provide more numerical results and analyses to validate the improvement in network slimming by using nonconvex regularization.
\section{Related Works}
\subsection{Compression Techniques for CNN}
\textbf{Low-rank decomposition.} Low-rank decomposition aims to reduce weight matrices to their low-rank structures for faster computation and more efficient storage. One set of methods focuses on decomposing pre-trained weight tensors. Denton \etal~\cite{denton2014exploiting} compressed the weight tensors of convolutional layers using singular value decomposition to approximate them. Jaderberg \etal~\cite{jaderberg2014speeding} exploited the redundancy between different feature channels and filters to approximate a full-rank filter bank in CNNs by combinations of a rank-one filter basis.  On the other hand, there are methods that train CNNs with low-rank weight matrices from scratch. Tai \etal \cite{tai2015convolutional} incorporated low-rank tensor decomposition into their CNN training algorithm. Wen \etal~\cite{wen2017coordinating} proposed \textit{force regularization} to train a CNN towards having a low-rank representation. Xu \etal~\cite{xu2018trained,xu2020trp}  developed trained rank pruning, an optimization scheme that incorporates low-rank decomposition into the training process. Trained rank pruning was further strengthened by nuclear norm regularization.

\textbf{Weight Quantization.} Quantization aims to represent weights with low-precision values ($\leq$8 bits arithmetic). The simplest form of quantization is binarization, constraining weights to only two values. Courbariaux \etal~\cite{courbariaux2015binaryconnect} proposed BinaryConnect, a method that trains deep neural networks (DNNs) with strictly binary weights. Neural networks with ternary weights have also been developed and investigated. Li \etal~\cite{li2016ternary} created ternary weight networks, where the weights are only $-1,0$, or $+1$. Zhu \etal~\cite{zhu2016trained} proposed Trained Ternary Quantization that constrains the weights to more general values $-W^n, 0$, and $W^p$, where $W^n$ and $W^p$ are parameters learned through the training process. For more general quantization, Yin \etal~\cite{yin2018binaryrelax} developed BinaryRelax, which relaxes the quantization constraint into a continuous regularizer for the optimization problem needed to be solved in CNNs. Later, Bai \etal \cite{bai2018proxquant} proposed Proxquant, a stochastic proximal gradient method for quantizing networks while training them.

\textbf{Pruning.} Pruning methods identify which weights, filters, and/or channels in CNNs are redundant and remove them from the networks. Early works focus on pruning weights. Han \etal~\cite{han2015learning} proposed a three-step framework to first train a CNN, prune weights if their norms are below a fixed threshold, and retrain the compressed CNN. Aghasi \etal~\cite{aghasi2017net,aghasi2020fast} proposed using convex optimization to determine which weights to prune while preserving model accuracy. For CNNs, channel or filter pruning is preferred over individual weight pruning since the former significantly eliminates more unnecessary weights. Li \etal~\cite{li2016pruning} calculated the sum of absolute weights for each filter of the CNN and pruned the filters with the lowest sums. On the other hand, Hu \etal~\cite{hu2016network} proposed a metric that measures the redundancies in channels to determine which to prune. Network slimming~\cite{liu2017learning} is also another method of channel pruning since it prunes channels with the smallest associated scaling factors. Zhao \etal \cite{zhao2019variational} improved network slimming by incorporating a variational Bayesian framework.

\textbf{Sparse optimization.} Sparse optimization methods introduce a sparse regularizer term to the loss function of the CNN so that the CNN is trained to have a  compressed structure from scratch. BinaryRelax~\cite{yin2018binaryrelax} and network slimming~\cite{liu2017learning} are examples of sparse optimization methods for CNNs. Alvarez and Salzmann~\cite{alvarez2016learning} and Scardapane \etal~\cite{scardapane2017group} applied group lasso~\cite{yuan2006model} and sparse group lasso~\cite{scardapane2017group} to CNNs to obtain group-sparse networks. Nonconvex regularizers have also been examined recently. Xue and Xin~\cite{xue2019learning} used $\ell_0$ and  T$\ell_1$ regularization in three-layer CNNs that classify shaky vs. normal handwriting. Both Ma \etal~\cite{ma2019transformed} and Pandit \etal \cite{pandit2021learning} proposed a regularizer that combines group sparsity and T$\ell_1$ and applied it to CNNs for image classification. Bui \etal \cite{bui2021structured} generalized sparse group lasso to incorporate nonconvex regularizers and applied it to various CNN architectures. Li \etal \cite{Li_2020_CVPR} introduced sparsity-inducing matrices into CNNs and imposed group sparsity on the rows or columns via $\ell_1$ or other nonconvex regularizers to prune filters and/or channels.

\begin{figure*}[t!!]
\centering
\begin{subfigure}{0.4\columnwidth}
\includegraphics[width=\columnwidth]{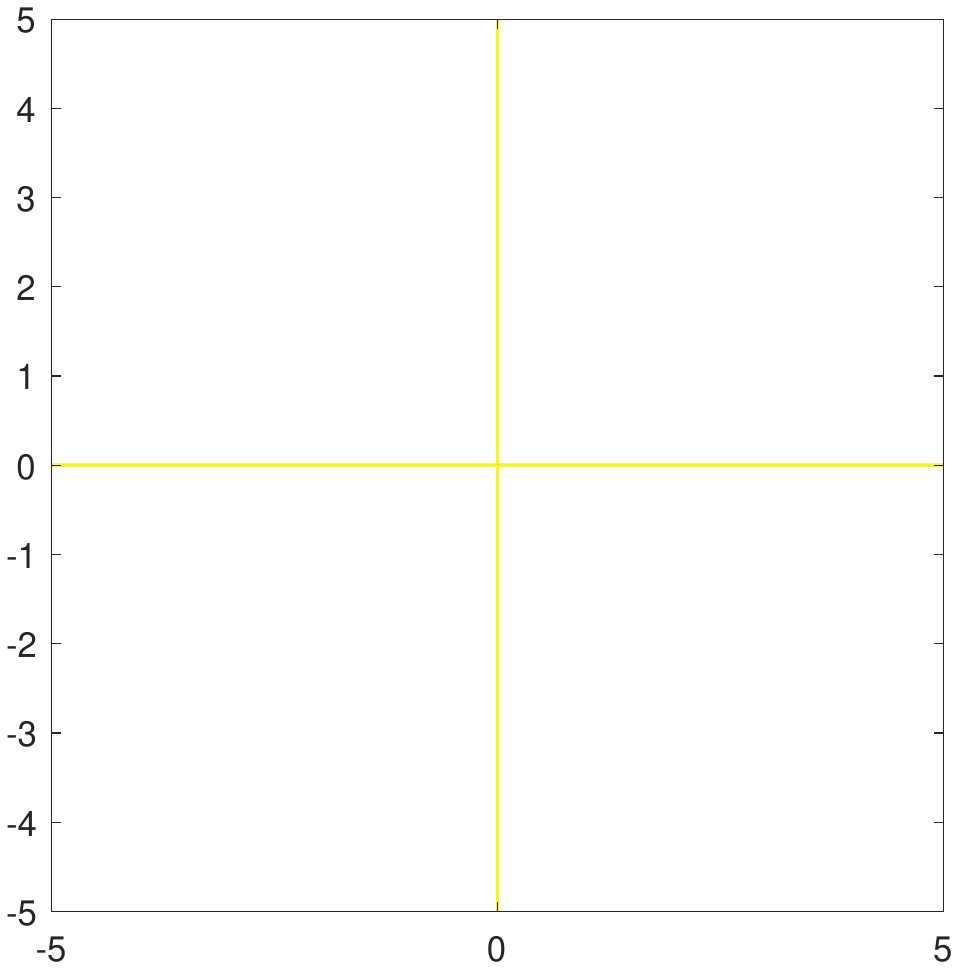}%
\caption{$\ell_0$}%
\label{subfig:l0}%
\end{subfigure}
\centering
\begin{subfigure}{0.4\columnwidth}
\includegraphics[width=\columnwidth]{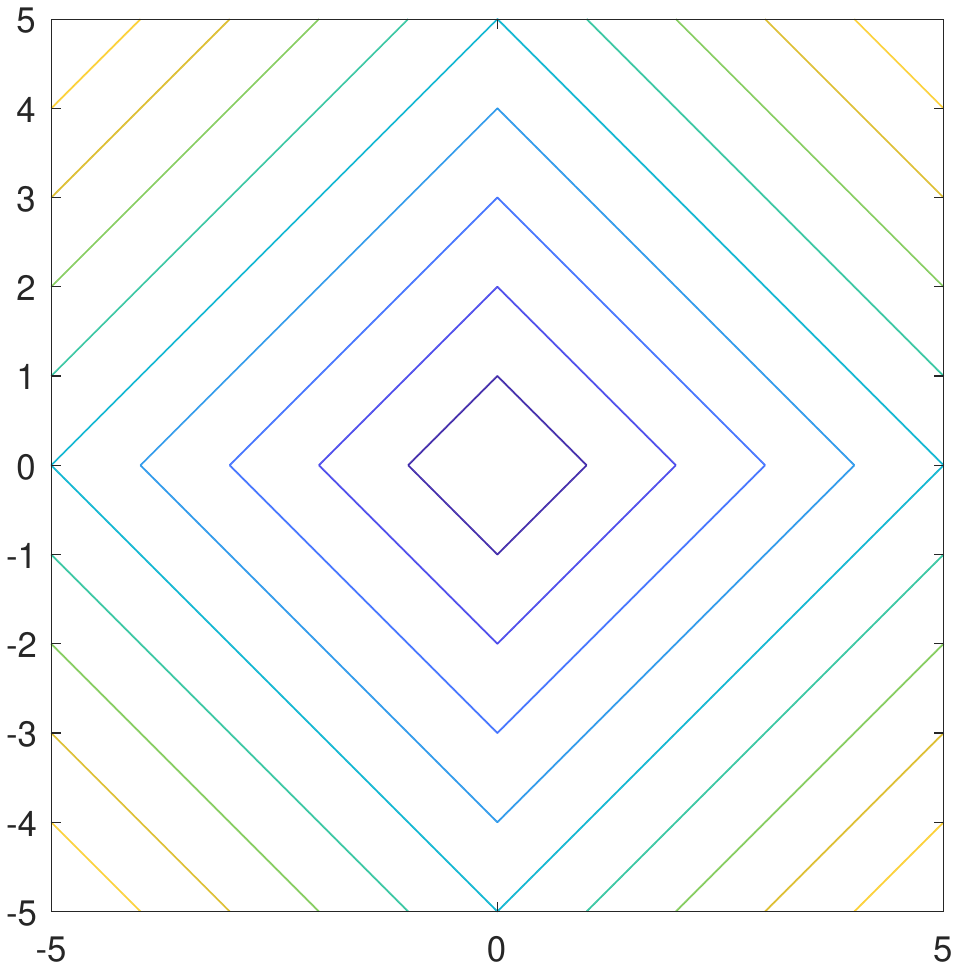}
\caption{$\ell_1$}%
\label{subfig:l1}%
\end{subfigure}
\centering
\begin{subfigure}{0.4\columnwidth}
\includegraphics[width=\columnwidth]{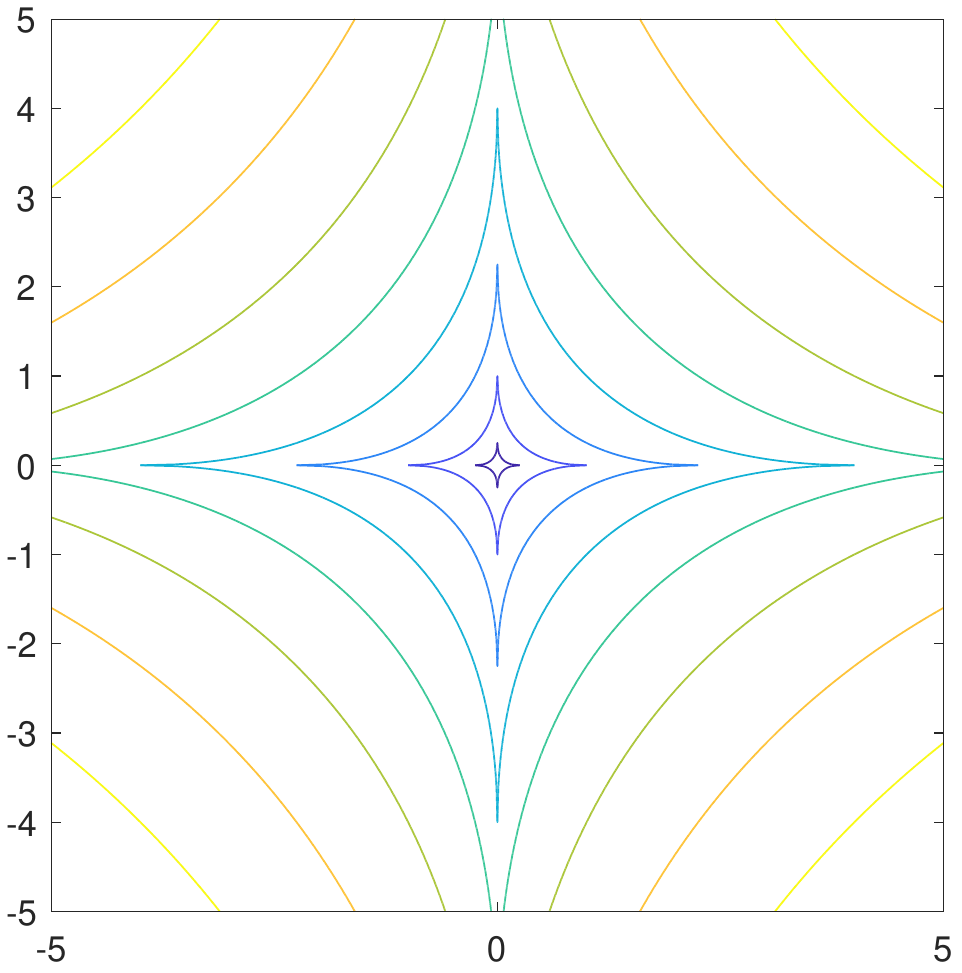}%
\caption{$\ell_{1/2}$}%
\label{subfig:l12}%
\end{subfigure}
\centering
\begin{subfigure}{0.4\columnwidth}
\includegraphics[width=\columnwidth]{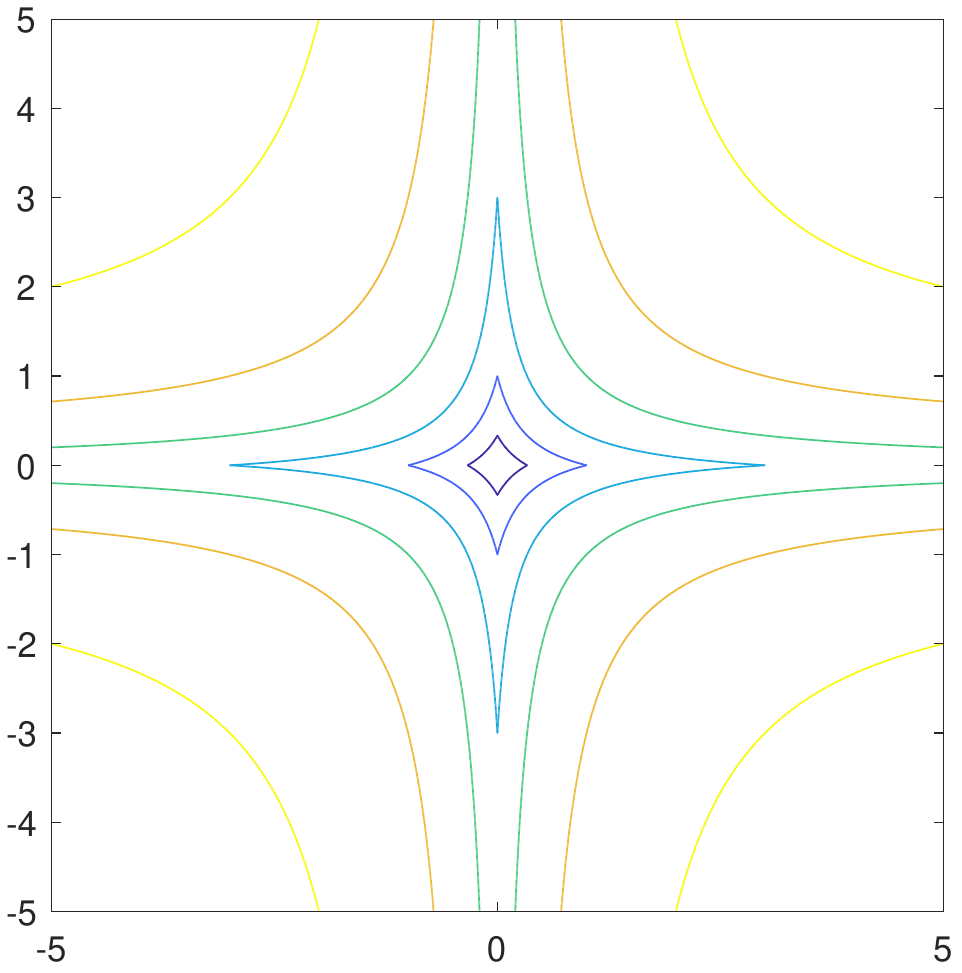}%
\caption{$\text{T}\ell_{1}, a = 1$}%
\label{subfig:tl1}%
\end{subfigure}\centering
\begin{subfigure}{0.4\columnwidth}
\includegraphics[width=\columnwidth]{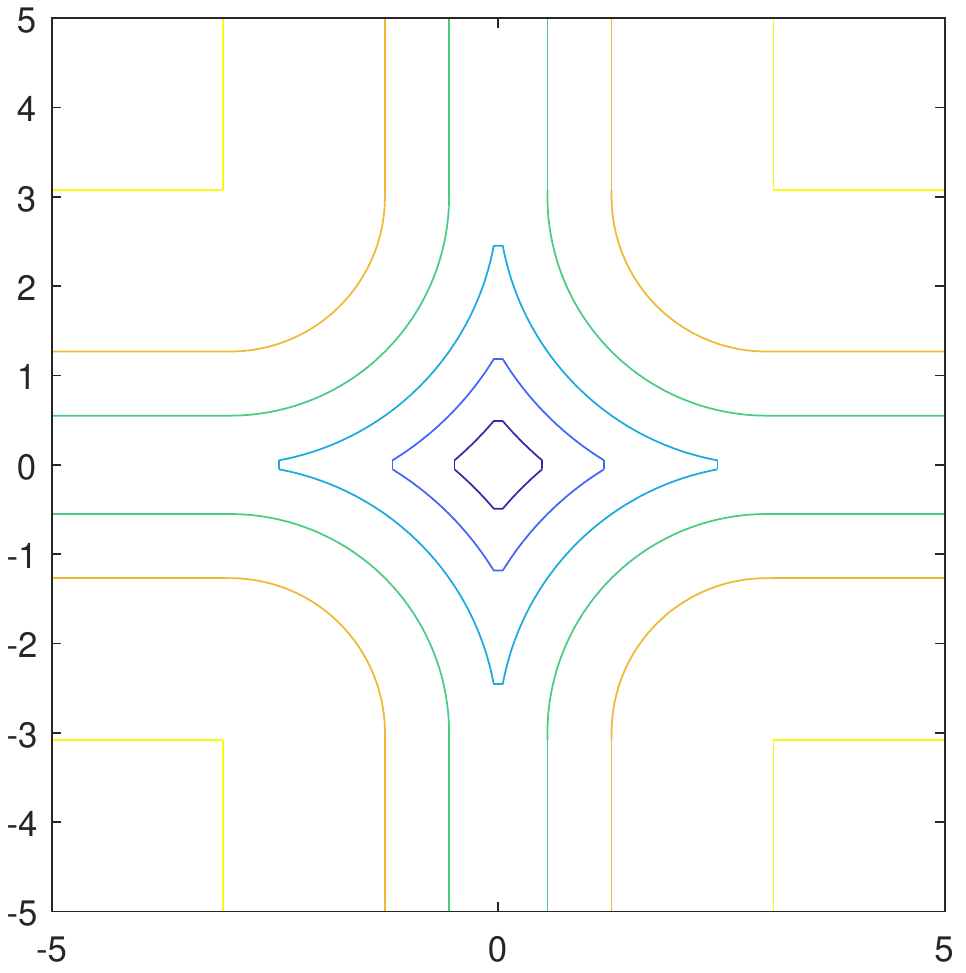}%
\caption{$\text{MCP}, \lambda = 1, a =3$}%
\label{subfig:MCP}%
\end{subfigure}\centering
\begin{subfigure}{0.4\columnwidth}
\includegraphics[width=\columnwidth]{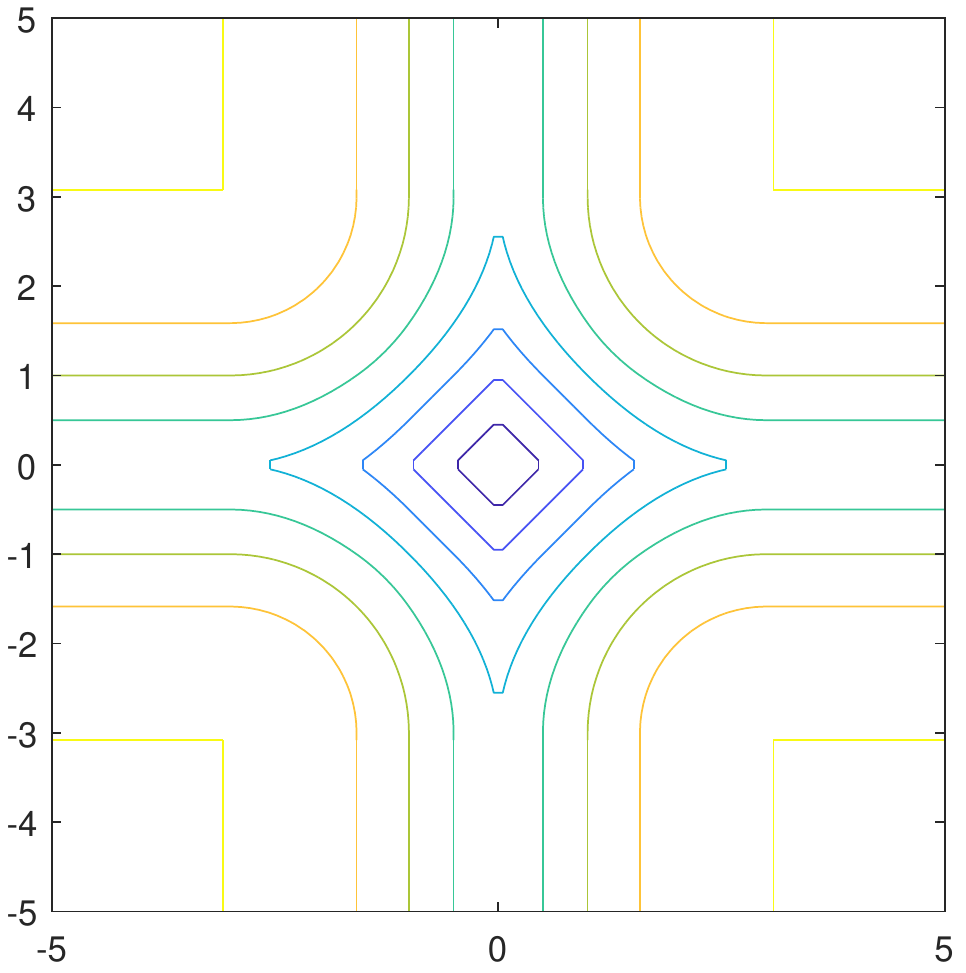}%
\caption{SCAD, $\lambda = 1, a =3$}%
\label{subfig:SCAD}%
\end{subfigure}
\caption{Contour plots of sparse regularizers.}
\label{fig:penalty}
\vspace{-4mm}
\end{figure*}

\subsection{Regularization Penalty}
Let $z = (z_1, \ldots, z_n) \in \mathbb{R}^n$. The $\ell_1$ penalty is described by
\begin{align}
    \|z\|_1 = \sum_{i=1}^n |z_i|,
\end{align}
while the $\ell_0$ penalty is described by
\begin{align}
    \|z\|_0 = \sum_{i=1}^n \mathbbm{1}_{\{z_i \neq 0\}}, \quad \text{where}  \quad   \mathbbm{1}_{\{z_i \neq 0 \}} = \begin{cases}
    1 & \text{ if } z_i \neq 0 \\
    0 & \text{ if } z_i = 0.
    \end{cases}
\end{align}
Although $\ell_1$ regularization is popular in sparse optimization in various applications such as compressed sensing~\cite{candes-2006,candes2006robust,yin2008bregman} and compressive imaging~\cite{jung2007improved,lustig2007sparse}, it may not actually yield the sparsest solution~\cite{chartrand2007exact,lou-2015-cs,lou2015computational,xu2012l,zhang2018minimization}. Moreover, it is sensitive to outliers and it may yield biased solutions~\cite{fan2001variable}.

A nonconvex alternative to the $\ell_1$ penalty is the $\ell_{p}$ penalty
\begin{align}
    \|z\|_{p} = \left(\sum_{i=1}^n |z_i|^{p}\right)^{1/p}
\end{align}
for $p \in (0, 1)$. The $\ell_p$ penalty interpolates $\ell_0$ and $\ell_1$ because as $p \rightarrow 0^+$, we have $\ell_p \rightarrow \ell_0$, and as $p \rightarrow 1^-$, we have $\ell_p \rightarrow \ell_1$. It recovers sparser solution than $\ell_1$ for certain compressed sensing problems~\cite{chartrand2008iteratively,chartrand2008restricted}. Empirical studies~\cite{chartrand2008iteratively,zong2012representative} demonstrate that for $p \in [1/2,1)$, as $p$ decreases, the solution becomes sparser by $\ell_p$ minimization, but for $p \in (0,1/2)$, the performance becomes no longer significant. Moreover, it is used in image deconvolution~\cite{krishnan2009fast,cao2013fast}, hyperspectral unmixing~\cite{qian2011hyperspectral}, computed topography reconstruction \cite{miao2015general}, and image segmentation~\cite{li2020tv, wu2021two}. Numerically, in compressed sensing, a small value $\epsilon$ is added to $z_i$ to avoid blowup in the subgradient when $z_i = 0$. In this work, we will examine across different values of $p$ since $\ell_p$ regularization may work differently in deep learning than in other areas.

Although $\ell_p$ may yield sparser solutions than $\ell_1$, it is still biased because parameters with large weights could be overpenalized \cite{fan2004nonconcave}. Hence, a better regularizer should also be unbiased. In fact, Fan and Li \cite{fan2001variable} suggested three properties that a regularizer should have: (1) continuity to avoid model instability; (2) sparsity to reduce model complexity; and (3) unbiasedness to avoid modeling bias due to overpenalization of large parameters. Hence, we consider regularizers that have all three properties, such as T$\ell_1$, MCP, and SCAD.

The $\text{T}\ell_1$ penalty is formulated as
\begin{align}
    P_a(z) = \sum_{i=1}^n \frac{(a+1)|z_i|}{a+|z_i|}
\end{align}
for $a > 0$. T$\ell_1$ interpolates $\ell_0$ and $\ell_1$ because as $a \rightarrow 0^+$, we have T$\ell_1 \rightarrow \ell_0$, and as $a \rightarrow + \infty$, we have T$\ell_1 \rightarrow \ell_1$. It was validated to have the three aforementioned properties \cite{lv2009unified}. The $\text{T}\ell_1$ penalty outperforms $\ell_1$ and $\ell_p$ in compressed sensing problems with both coherent and incoherent sensing matrices~\cite{zhang2014minimization,zhang2018minimization}. Additionally, the $\text{T}\ell_1$ penalty yields satisfactory, sparse solutions in matrix completion~\cite{zhang2015transformed} and deep learning~\cite{ma2019transformed}.

The MCP penalty \cite{zhang2010nearly} is provided by
\begin{equation}
\begin{split}
    &p_{\lambda, a}(z)\\
    &= \sum_{i=1}^n \left[\left(\lambda |z_i| - \frac{z_i^2}{2a}\right)\mathbbm{1}_{\{|z_i| \leq a \lambda \}} + \frac{a \lambda^2}{2} \mathbbm{1}_{\{|z_i| > a\lambda\}} \right],
    \end{split}
\end{equation}
where $\lambda \geq 0$ and $a >1$. The parameter $\lambda$ acts as a regularization parameter while the parameter $a$ controls the level of sparsity, where the smaller $a$ is, the sparser the solution becomes. In fact, $a$ allows MCP to roughly interpolate between $\ell_0$ and $\ell_1$. Originally, MCP is developed for variable selection \cite{zhang2010nearly}, but it has been utilized in various other applications such as image restoration \cite{you2019nonconvex} and matrix completion \cite{jin2016alternating}.

Lastly, the SCAD penalty \cite{fan2001variable} is given by
\begin{align}
\begin{split}
    &\tilde{p}_{\lambda,a}(z)\\
    &=\sum_{i=1}^n \bigg [ \lambda|z_i| \mathbbm{1}_{\{|z_i| \leq \lambda\}} + \frac{2 a \lambda |z_i| - z_i^2 - \lambda^2}{2(a-1)} \mathbbm{1}_{\{\lambda < |z_i| \leq a \lambda\}}\\& + \frac{\lambda^2(a+1)}{2} \mathbbm{1}_{\{|z_i| > a \lambda\}} \bigg],
\end{split}
\end{align}
where $\lambda \geq 0$ is the regularization parameter and $a>2$ controls the level of sparsity similarly to MCP. In both linear and logistic regression problems, SCAD outperforms $\ell_1$ in variable selection \cite{fan2001variable}. Beyond variable selection, it is applied in compressed sensing \cite{mehranian2013smoothly}, bioinformatics \cite{breheny2011coordinate,wang2007group}, image processing \cite{gu2017tvscad}, and wavelet approximation \cite{antoniadis2001regularization}.

Figure~\ref{fig:penalty} displays the contour plots of the aforementioned regularizers. With $\ell_1$ regularization, the solution tends towards one of the corners of the rotated squares, making it sparse. Compared with $\ell_1$, the level lines of the nonconvex regularizers bend more inward towards the axes, encouraging the solutions to coincide with one of the corners. In addition, the contour plots of the nonconvex regularizers appear more similar to the contour plot of $\ell_0$. Therefore, solutions tend to be sparser with nonconvex regularization than with $\ell_1$ regularization. 

For further discussion on the aforementioned nonconvex regularizers, \cite{ahn2017difference} and \cite{wen2018survey} provide detailed survey, application, and analysis.

Throughout the rest of the paper, we define $\lambda p_{1,a}(\cdot) \coloneqq p_{\lambda, a}(\cdot)$ and $\lambda \tilde{p}_{1,a}(\cdot) \coloneqq \tilde{p}_{\lambda, a}(\cdot)$.
\section{Proposed Method}
\subsection{Batch Normalization Layer}
Batch normalization~\cite{ioffe2015batch} has been instrumental in speeding the convergence and improving generalization of many deep learning models, especially CNNs~\cite{szegedy2016rethinking,he2016deep}. In most state-of-the-arts CNNs, a convolutional layer is always followed by a batch normalization layer. Within a batch normalization layer, features generated by the preceding convolutional layer are normalized by their mean and variance within the same channel. Afterward, a linear transformation is applied to compensate for the loss of their representative abilities. 
\begin{figure}[t!]
    \centering
    \includegraphics[scale =0.25]{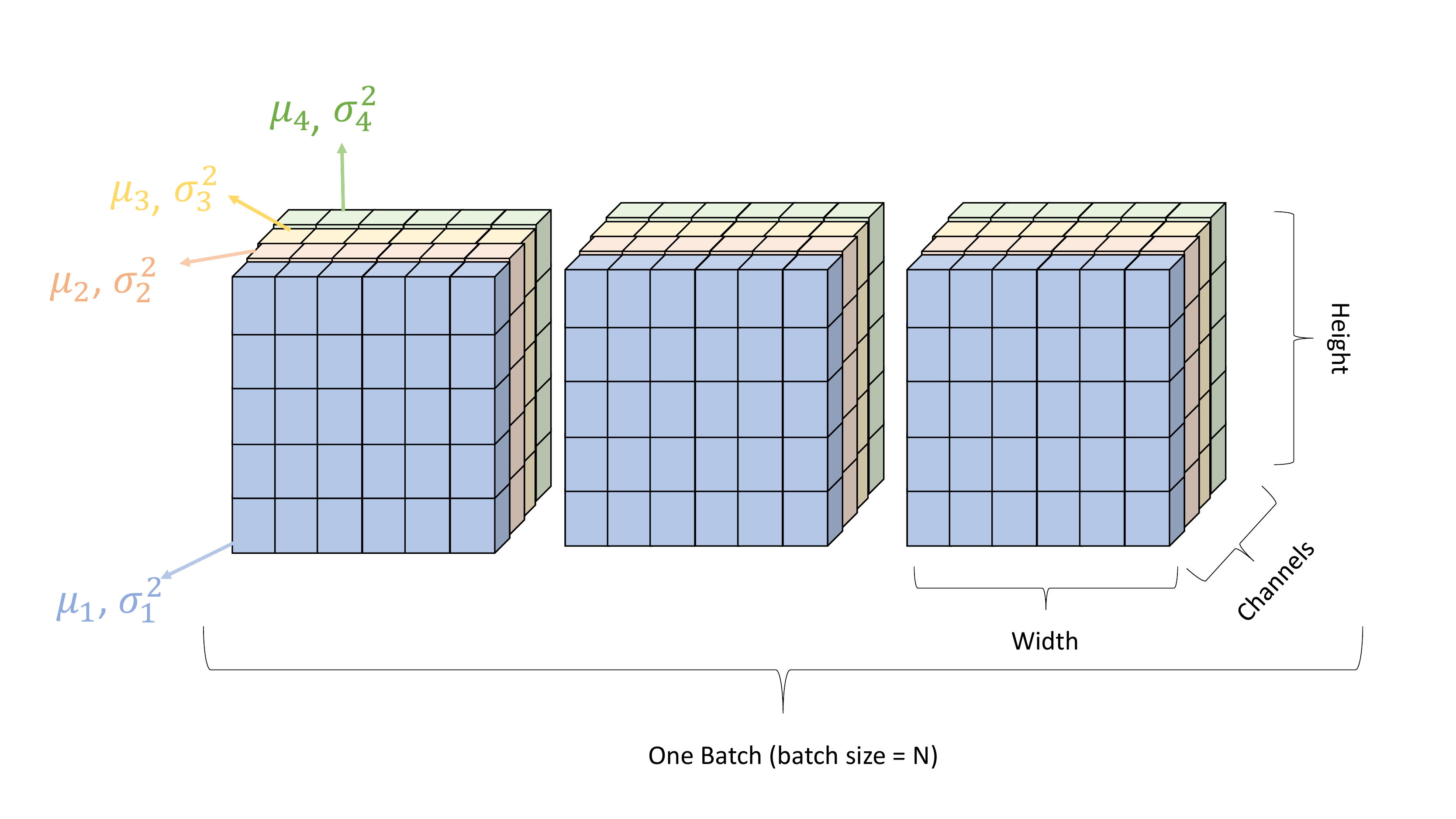}
    \caption{Visualization of batch normalization on a feature map. The mean and variance of the values of the pixels of the same colors corresponding to the channels are computed and are used to normalize these pixels.}
    \label{fig:BN}
    \vspace{-4mm}
\end{figure}

We mathematically describe the process of the batch normalization layer. First we suppose that we are working with 2D images. Let $x'$ be a feature computed by a convolutional layer. Each entry of $x'$ is denoted by $x_i'$, where $i = (i_N,i_C, i_H, i_W)$ indexes the features in $(N, C, H, W)$ order. Here, $N$ is the batch axis, $C$ is the image channel axis, $H$ is the image height axis, and $W$ is the image width axis. We define the index set $S_i = \{k = (k_N, k_C, k_H, k_W): k_C = i_C\}$,
where $k_C$ and $i_C$ are the respective subindices of $k$ and $i$ along the $C$ axis. In other words, the index set consists of pixels that belong to the same channel. The mean $\mu_i$ and variance $\sigma_i^2$ are computed as follows:
\begin{align}
    \mu_i &= \frac{1}{|S_i|} \sum_{k \in S_i} x_k', \quad \sigma_i^2 = \frac{1}{|S_i|} \sum_{k \in S_i} (x_k'-\mu_i)^2 + \epsilon
\end{align}
for some small value $\epsilon > 0$, where $|\mathcal{A}|$ denotes the cardinality of the set $\mathcal{A}$.  Then we normalize $x_i'$ by $\hat{x}_i = \frac{x_i' - \mu_i}{\sigma_i}$
for each index $i$. In short, the mean and variance are computed from pixels of the same channel index and are used to normalize them. Visualization is provided in Figure~\ref{fig:BN}. Lastly, the output of the batch normalization layer is computed as a linear transformation of the normalized features:
\begin{align} \label{eq:lin_transform}
    z_i = \gamma_{i_C} \hat{x}_i + \beta_{i_C},
\end{align}
where $\gamma_{i_C}, \beta_{i_C} \in \mathbb{R}$ are trainable parameters. Additionally, $\gamma_{i_C}$ is defined to be the scaling factor related to the channel $i_C$. 
\subsection{Network Slimming with Nonconvex Sparse Regularization}
 \begin{algorithm}
 \caption{Algorithm for minimizing \eqref{eq:min_problem}}
 \label{alg:sgd}
 \begin{algorithmic}[1]
 \renewcommand{\algorithmicrequire}{\textbf{Input:}}
 \renewcommand{\algorithmicensure}{\textbf{Output:}}
 \REQUIRE Regularization parameter $\lambda$, learning rate $\eta$, sparse regularizer $\mathcal{R}$
 \\ Initialize $\mathcal{W}^0$, excluding $\{\gamma_l\}_{l=1}^L$, with random values.\\
 Initialize $\{\gamma_l^0\}_{l=1}^L$ with entries 0.5.
  \FOR {each epoch $t=1, \ldots, T$}
   \STATE $\mathcal{W}^{t} = \mathcal{W}^{t-1} - \displaystyle \frac{\eta}{N} \sum_{i=1}^N \nabla \mathcal{L}(h(x_i, \mathcal{W}^{t-1}), y_i)$  by stochastic gradient descent or variant.
   \STATE $\gamma_l^{t} = \gamma_l^{t-1} - \eta \lambda \partial \mathcal{R}(\gamma_l^{t-1})$ for $l=1, \ldots, L$.
  \ENDFOR
 \end{algorithmic} 
 \end{algorithm}
\begin{table*}[t!]
\scriptsize
\centering
\caption{Sparse regularizers and their (limiting) subgradients.}
\label{tab:reg}
\begin{tabular}{|l|c|c|}
\hline
\textbf{Name} & $\mathcal{R}(z)$ & $\partial{R(z)}$\\ \hline & &\\
$\ell_1$ & $\|z\|_1 = \displaystyle \sum_{i=1}^n |z_i|$ & $\partial\|z\|_1 = \left\{\zeta \in \mathbb{R}^n: \zeta_i = \begin{cases} \text{sgn}(z_i) &\text{ if } z_i \neq 0 \\
\zeta_i \in [-1,1] &\text{ if } z_i = 0
\end{cases} \right\}$\\ & &\\
\hline & & \\
$\ell_{p}$ & $\|z\|_{p}^{p} = \displaystyle \sum_{i=1}^n |z_i|^{p}$& $\partial \|z\|_{p}^{p} = \left\{\zeta \in \mathbb{R}^n: \zeta_i = \begin{cases} \displaystyle \frac{p\cdot \text{sgn}(z_i)}{|z_i|^{1-p}} &\text{ if } z_i \neq 0 \\ 
\zeta_i \in \mathbb{R} &\text{ if } z_i = 0
\end{cases} \right\}$\\ & & \\
\hline
$\text{T}\ell_1$ & $P_a(z) = \displaystyle \sum_{i=1}^n \frac{(a+1)|z_i|}{a+|z_i|}$ & $\partial P_a(z) = \left\{\zeta \in \mathbb{R}^n: \zeta_i = \begin{cases} \displaystyle  \frac{a(a+1)\text{sgn}(z_i)}{(a+|z_i|)^2} &\text{ if } z_i \neq 0 \\ 
\zeta_i \in \left[-\frac{a+1}{a}, \frac{a+1}{a} \right] &\text{ if } z_i = 0
\end{cases} \right\}$\\
\hline 
MCP & $p_{\lambda, a}(z) =  \displaystyle \sum_{i=1}^n \left[\left(\lambda |z_i| - \frac{z_i^2}{2a}\right)\mathbbm{1}_{\{|z_i| \leq a \lambda \}} + \frac{a \lambda^2}{2} \mathbbm{1}_{\{|z_i| > a\lambda\}} \right]$ & $\partial p_{\lambda,a}(z) = \left\{\zeta \in \mathbb{R}^n: \zeta_i = \begin{cases} 0 &\text{ if } |z_i| > a \lambda \\
\lambda \text{sgn}(z_i) - \displaystyle \frac{z_i}{a} & \text{ if } 0 < |z_i| \leq a \lambda \\
\zeta_i \in \left[-\lambda,\lambda \right] &\text{ if } z_i = 0
\end{cases} \right\}$\\
\hline 
SCAD & \makecell{$\tilde{p}_{\lambda, a}(z)$ =  $\displaystyle \sum_{i=1}^n \bigg [ \lambda|z_i| \mathbbm{1}_{\{|z_i| \leq \lambda\}} + \frac{2 a \lambda |z_i| - z_i^2 - \lambda^2}{2(a-1)} \mathbbm{1}_{\{\lambda < |z_i| \leq a \lambda\}}$\\ $+ \displaystyle \frac{\lambda^2(a+1)}{2} \mathbbm{1}_{\{|z_i| > a \lambda\}} \bigg]$} & $\partial \tilde{p}_{\lambda, a}(z) = \left\{\zeta \in \mathbb{R}^n: \zeta_i = \begin{cases} \displaystyle 0 &\text{ if } |z_i| > a \lambda \\ 
\displaystyle\frac{a \lambda \text{sgn}(z_i)-z_i}{a-1} & \text{ if } \lambda < |z_i| \leq a \lambda \\
\lambda \text{sgn}(z_i) &\text{ if } 0< |z_i| \leq \lambda\\ 
\zeta_i \in \left[-\lambda, \lambda \right] &\text{ if } z_i = 0
\end{cases} \right\}$\\
\hline 
\end{tabular}
\vspace{-4mm}
\end{table*}
Since the scaling factors $\gamma_{i_C}$'s in \eqref{eq:lin_transform} are associated with the channels of a convolutional layer, we aim to penalize them with a sparse regularizer in order to identify which channels are irrelevant to the compressed CNN model. Suppose we have a training dataset that consists of $N$ input-output pairs $\{(x_i, y_i)\}_{i=1}^{N}$ and a CNN with $L$ convolutional layers, where each is followed by a batch normalization layer. Then we have two sets of vectors $\{\gamma_l\}_{l=1}^L$ and $\{\beta_l\}_{l=1}^L$, where $\gamma_l = (\gamma_{l,1}, \ldots, \gamma_{l, C_l})$ and 
    $\beta_l = (\beta_{l,1}, \ldots, \beta_{l, C_l})$
with $C_l$ being the number of channels in the $l$th convolutional layer. Let $\mathcal{W}$ be the weight parameters that include  $\{\gamma_l\}_{l=1}^L$ and $\{\beta_l\}_{l=1}^L$. Hence, the trainable parameters $\mathcal{W}$ of the CNN are learned by minimizing the following objective function:
\begin{align}\label{eq:min_problem}
    \frac{1}{N} \sum_{i=1}^N \mathcal{L}(h(x_i, \mathcal{W}), y_i) + \lambda \sum_{l=1}^L \mathcal{R}(\gamma_l), 
\end{align}
where  $h(\cdot, \cdot)$ is the output of the CNN used for prediction, $\mathcal{L}(\cdot, \cdot)$ is a loss function, $\mathcal{R}(\cdot)$ is a sparse regularizer, and $\lambda>0$ is a regularization parameter for $\mathcal{R}(\cdot)$. When $\mathcal{R}(\cdot) = \|\cdot\|_1$, we have the original network slimming method. As mentioned earlier, since $\ell_1$ regularization may not yield the sparsest solution and it could potentially be biased, we investigate the method with a nonconvex regularizer, where $\mathcal{R}(\cdot)$ is $\|\cdot\|_{p}^{p}$, $P_a(\cdot)$, $p_{1,a}(\cdot)$, or $\tilde{p}_{1,a}(\cdot)$. 

To minimize \eqref{eq:min_problem}, stochastic gradient descent is applied to the loss function term while subgradient descent is applied to the regularizer term~\cite{shor2012minimization}. The algorithm is summarized in Algorithm \ref{alg:sgd}.  Subgradient descent is applicable to the nonconvex regularizers $\mathcal{R}(z)$ for $z \in \mathbb{R}^n$ as it is for $\ell_1$. Like $\ell_1$, the nonconvex regularizers are of the form $\sum_{i=1}^n r(z_i)$, where $r: \mathbb{R} \rightarrow \mathbb{R}$ has the following properties:
\begin{enumerate}[(i)]
    \item $r(0) = 0$;
    \item $r$ is an even, proper, and continuous function;
    \item $r$ is increasing on $[0, +\infty)$;
    \item $r$ is differentiable on $(-\infty, 0) \cup (0, +\infty)$.
\end{enumerate}
These properties ensure that $r$ is differentiable everywhere except at $0$ and $0$ is the global minimum of $r$ while being its only local minimum. As a result, the regularizers are differentiable when $z_i \neq 0$ for all $i=1, \ldots, n$. Hence, subgradient descent becomes gradient descent at these points. If $z_i = 0$ for at least one index $i$, then we need to compute its (limiting) subgradient \cite[Definition~6.1]{penot2012calculus} and decide a candidate descent direction. Fortunately, because $\mathcal{R}(z) = \sum_{i=1}^n r(z_i)$, we have
\begin{align*}
    \partial{\mathcal{R}}(z) = \left(\partial r(z_1), \partial r(z_2), \ldots, \partial r(z_n) \right)
\end{align*}
by \cite[Proposition~6.17(e)]{penot2012calculus}. This means that at each component $r(z_i)$, we can compute its subgradient $\partial r(z_i)$ individually and select a descent direction from the set. Since $0$ is a local minimum of $r$, we have $0 \in \partial r(0)$, so we can select $0$ as a descent direction for simplicity. Table 1 presents the subgradients of the regularizers. 

After the CNN is trained with \eqref{eq:min_problem} using Algorithm \ref{alg:sgd}, we prune the channels whose scaling factors are small in magnitude, giving us a compressed model. However, the compressed model may lose its original accuracy, so it may need to be retrained but without the sparse regularizer in order to attain its original accuracy or better. 

\begin{table*}[p]
  \rotatebox{90}{\begin{minipage}{\textheight}
\caption{Effect of channel pruning on the mean pruned parameter / FLOPs percentages (\%) on VGG-19 trained on (a) CIFAR 10, (b) CIFAR 100, and (c) SVHN. The mean is computed from five runs for each regularizer. For each channel pruning ratio, \textbf{bold} indicates outperforming $\ell_1$; * indicates best value; and NA indicates at least one of the five models is over-pruned.}
\label{tab:vgg_result}
\scriptsize
\scalebox{0.95}{
\begin{tabular}{|c|c|c|c|c|c|c|c|c|c|}\hline
    (a) &\multicolumn{9}{c|}{CIFAR 10}  \\
     \hline
     \makecell{Channel Pruning \\ Ratio}& 0.10 & 0.20 & 0.30 & 0.40 & 0.50 & 0.60 & 0.70 & 0.80 & 0.90 \\ \hline
     $\ell_1$& 21.10\% / 11.83\% & 39.34\% / 22.18\% & 54.88\% / 31.09\% & 67.56\% / 38.51\% & 77.53\% / 44.78\% & 84.71\% / 49.68\% & 88.81\% / 51.95\% & NA  & NA\\ \hline
     $\ell_{3/4}$& \textbf{21.14\%} / \textbf{12.11\%} & \textbf{39.55\%} / \textbf{22.71\%} & \textbf{55.13\%} / \textbf{32.06\%} & \textbf{67.96\%} / \textbf{39.97\%} & \textbf{77.99\%} / \textbf{46.08\%}& \textbf{85.24\%} / \textbf{51.37\%} & \textbf{89.69\%} / \textbf{54.96\%}& NA & NA\\ \hline 
     $\ell_{1/2}$& \textbf{21.12\%} / \textbf{12.38\%} & \textbf{39.62\%} / \textbf{22.65\%}& \textbf{55.29\%}* / \textbf{32.06\%} & \textbf{68.09\%}* / \textbf{39.75\%} & \textbf{78.10\%} / \textbf{46.45\%} & \textbf{85.43\%} / \textbf{52.02\%} & \textbf{90.01\%} / \textbf{56.12\%} & \textbf{93.66\%} / \textbf{65.52\%}& NA\\ \hline
     $\ell_{1/4}$& 19.95\% / \textbf{15.49\%}* & 37.66\% / \textbf{29.82\%}* & 52.99\% / \textbf{42.93\%}*& 66.20\% / \textbf{54.39\%}* & 77.07\% / \textbf{64.47\%}* & \textbf{85.76\%} / \textbf{73.44\%}*& \textbf{92.14\%}* / \textbf{81.89\%}* & \textbf{96.54\%}* / \textbf{91.07\%}*& \textbf{99.05\%}* / \textbf{98.32\%}*\\ \hline
     T$\ell_1 (a=10.0)$& \textbf{21.15\%} / \textbf{11.92\%} & \textbf{39.41\%} / \textbf{22.59\%} & \textbf{54.95\%} / \textbf{31.85\%} & \textbf{67.71\%} / \textbf{39.39\%} & \textbf{77.69\%} / \textbf{45.77\%} & \textbf{84.89\%} / \textbf{50.47\%} & \textbf{89.06\%} / \textbf{52.75\%} & NA & NA\\ \hline
     T$\ell_1 (a=1.0)$ & \textbf{21.16\%} / \textbf{12.12\%} & \textbf{39.35\%} / \textbf{23.13\%} & \textbf{54.94\%} / \textbf{32.60\%} & \textbf{67.88\%} / \textbf{40.69\%} & \textbf{78.06\%} / \textbf{47.94\%} & \textbf{85.55\%} / \textbf{53.40\%} & \textbf{90.34\%} / \textbf{57.43\%} & NA  & NA  \\ \hline
     T$\ell_1 (a=0.5)$ & 20.94\% / \textbf{12.59\%}& 39.29\% / \textbf{23.66\%} & \textbf{54.92\%} / \textbf{33.61\%} & \textbf{67.83\%} / \textbf{42.34\%} & \textbf{78.22\%}* / \textbf{49.58\%} & \textbf{85.92\%}* / \textbf{55.36\%} & \textbf{90.88\%} / \textbf{59.84\%} & NA & NA\\ \hline
     MCP$(a=15000)$& \textbf{21.18\%} / 11.56\% & \textbf{39.48\%} / 21.43\% & \textbf{54.96\%} / 30.31\% & \textbf{67.62\%} / 37.75\% & 77.53\% / 43.76\% & 84.58\% / 48.05\% & 88.58\% / 50.69\%  & NA & NA\\ \hline
     MCP$(a=10000)$ & 20.99\% / 11.24\% & \textbf{39.35\%} / 21.23\% & \textbf{54.97\%} / 29.94\% & \textbf{67.64\%} / 37.71\% & \textbf{77.55\%} / 43.42\% & 84.61\% / 47.88\% & 88.63\% / 50.49\% & NA & NA\\ \hline
     MCP$(a=5000)$& \textbf{21.20\%} / 10.97\% & \textbf{39.71\%}* / 20.92\% & \textbf{55.24\%} / 29.31\% & \textbf{67.87\%} / 36.32\% & \textbf{77.61\%} / 41.97\% & 84.53\% / 46.08\% & 88.56\% / 49.49\% & NA  & NA  \\ \hline
     SCAD$(a=15000)$& 21.10\% / 11.70\% & \textbf{39.42\%} / 21.83\% & \textbf{54.97\%} / 30.75\% & \textbf{67.68\%} / 37.83\% & \textbf{77.56\%} / 43.62\% & 84.66\% / 48.09\% & 88.71\% / 50.70\%  & NA  & NA\\ \hline
     SCAD$(a=10000)$& \textbf{21.21\%} / 11.23\% & \textbf{39.45\%} / 20.95\% & \textbf{54.95\%} / 29.89\% & \textbf{67.60\%} / 37.33\% & 77.44\% / 43.23\% & 84.53\% / 47.52\% & 88.57\% / 50.43\% &  NA & NA  \\ \hline
     SCAD$(a=5000)$ & \textbf{21.24\%}* / 11.25\% & \textbf{39.65\%} / 21.08\% & \textbf{55.16\%} / 29.64\% & \textbf{67.77\%} / 36.77\% & \textbf{77.58\%} / 42.69\% & 84.58\% / 46.96\% & 88.62\% / 50.19\%  & NA  & NA \\ \hline \hline
    (b) &\multicolumn{9}{c|}{CIFAR 100}  \\
     \hline
     \makecell{Channel Pruning \\ Ratio}& 0.10 & 0.20 & 0.30 & 0.40 & 0.50 & 0.60 & 0.70 & 0.80 &0.90 \\ \hline
     $\ell_1$&  21.91\% / 12.44\% & 40.36\% / 22.89\%* & 55.83\% / 28.19\% & 67.45\% / 31.75\% & 75.35\% / 36.08\% & NA  &  NA &  NA&NA\\ \hline
     $\ell_{3/4}$& \textbf{21.98\%} / 10.92\% & \textbf{40.64\%} / 20.75\% & \textbf{56.04\%} / \textbf{28.64\%} & \textbf{68.01\%} / \textbf{34.77\%} & \textbf{76.54\%} / \textbf{38.40\%} & NA  &NA  &NA & NA\\ \hline 
     $\ell_{1/2}$& \textbf{22.02\%} / 10.89\% & \textbf{40.85\%}* / 20.04\% & \textbf{56.29\%}* / 27.83\% & \textbf{68.41\%} / \textbf{34.23\%} & \textbf{77.19\%} / \textbf{39.40\%} & \textbf{83.07\%} / \textbf{43.82\%} &  NA& NA&NA \\ \hline
     $\ell_{1/4}$& \textbf{22.00\%} / 10.80\% & \textbf{40.71\%} / 20.52\% & \textbf{56.21\%} / \textbf{29.10\%} & \textbf{68.53\%}* / \textbf{36.59\%} & \textbf{78.16\%}* / \textbf{44.28\%}* & \textbf{85.71\%}* / \textbf{54.15\%}* & \textbf{91.48\%}* / \textbf{68.94\%}* & \textbf{96.54\%}* / \textbf{86.86\%}* & NA\\ \hline
     T$\ell_1 (a=10.0)$&  21.83\% / 12.41\% & 40.34\% / 22.62\% & 55.62\% / \textbf{29.71\%} & \textbf{67.80\%} / \textbf{32.97\%} & \textbf{76.07\%} / \textbf{37.00\%} &    NA& NA  & NA & NA\\ \hline
     T$\ell_1 (a=1.0)$ & 21.87\% / 11.21\% & \textbf{40.47\%} / 20.85\% & \textbf{55.99\%} / \textbf{29.04\%} & \textbf{68.22\%} / \textbf{36.22\%} & \textbf{77.18\%} / \textbf{40.47\%} & \textbf{82.90\%} / \textbf{43.94\%} &   NA& NA &NA \\ \hline
     T$\ell_1 (a=0.5)$ & 21.67\% / 11.42\% & 40.33\% / 21.52\% & \textbf{55.97\%} / \textbf{30.16\%}* & \textbf{68.49\%} / \textbf{37.52\%}* & \textbf{77.99\%} / \textbf{43.24\%} & \textbf{84.09\%} / \textbf{47.15\%} & NA   &NA & NA\\ \hline
     MCP$(a=15000)$&21.86\% / 12.37\% & 40.28\% / 22.40\% & 55.67\% / 28.17\% & 67.29\% / 31.62\% & \textbf{75.38\%} / 35.91\% & NA  & NA & NA & NA \\ \hline
     MCP$(a=10000)$ & 21.90\% / 12.40\% & 40.24\% / 22.60\% & 55.73\% / 27.94\% & 67.28\% / 31.60\% & 75.20\% / 35.95\%  & NA  & NA  &  NA&NA\\ \hline
     MCP$(a=5000)$&\textbf{22.03\%}* / 11.90\% & \textbf{40.49\%} / 21.45\% & \textbf{55.94\%} / 26.46\% & 67.35\% / 30.43\% & 75.03\% / 34.78\% & NA & NA   &  NA &  NA \\ \hline
     SCAD$(a=15000)$& \textbf{21.96\%} / \textbf{12.48\%}* & \textbf{40.42\%} / 22.36\% & 55.83\% / 28.04\% & \textbf{67.50\%} / 31.70\% & 75.34\% / 35.91\% & NA & NA  & NA & NA  \\ \hline
     SCAD$(a=10000)$& 21.90\% / 11.76\% & 40.28\% / 21.82\% & 55.71\% / 27.34\% & 67.18\% / 31.07\% & 75.02\% / 35.56\% &  NA &  NA  & NA &  NA  \\ \hline
     SCAD$(a=5000)$ & \textbf{22.01\%} / 11.59\% & \textbf{40.49\%} / 20.60\% & 55.75\% / 25.63\% & 66.91\% / 29.91\% & 74.50\% / 34.47\% & NA  & NA & NA  & NA  \\ \hline
     \hline
     (c) &\multicolumn{9}{c|}{SVHN}  \\
     \hline
     \makecell{Channel Pruning \\ Ratio}& 0.10 & 0.20 & 0.30 & 0.40 & 0.50 & 0.60 & 0.70 & 0.80 & 0.90 \\ \hline
     $\ell_1$& 19.92\% / 15.90\% & 37.51\% / 30.99\% & 52.91\% / 43.85\% & 66.08\% / 55.25\% & 76.98\% / 65.06\% & 85.60\% / 73.63\% & 92.00\% / 80.80\% & 96.12\% / 86.39\% & NA \\ \hline
     $\ell_{3/4}$& \textbf{19.96\%} / \textbf{16.04\%} & \textbf{37.60\%} / 30.91\% & \textbf{52.96\%} / 43.82\% & 66.00\% / \textbf{55.74\%} & 76.84\% / \textbf{65.93\%} & 85.52\% / \textbf{74.49\%} & 92.00\% / \textbf{81.53\%} & \textbf{96.23\%} / \textbf{87.23\%} &  NA\\ \hline 
     $\ell_{1/2}$& 19.80\% / \textbf{16.80\%} & 37.35\% / \textbf{31.70\%} & 52.74\% / \textbf{44.89\%} & 65.84\% / \textbf{56.70\%} & 76.79\% / \textbf{67.00\%} & 85.52\% / \textbf{75.75\%} & \textbf{92.01\%} / \textbf{83.14\%} & \textbf{96.31\%} / \textbf{88.36\%} & \textbf{98.94\%} / \textbf{95.45\%}\\ \hline
     $\ell_{1/4}$& 19.36\% / \textbf{17.72\%}* & 36.77\% / \textbf{32.96\%}* & 52.07\% / \textbf{47.02\%}* & 65.13\% / \textbf{59.17\%}* & 76.01\% / \textbf{70.57\%}* & 84.86\% / \textbf{79.92\%}* & 91.59\% / \textbf{87.85\%}* & \textbf{96.36\%} / \textbf{93.72\%}* & \textbf{99.08\%}* / \textbf{98.15\%}*\\ \hline
     T$\ell_1 (a=10.0)$& \textbf{19.94\%} / 15.90\% & 37.39\% / 30.91\% & 52.71\% / \textbf{44.18\%} & 65.94\% / \textbf{55.64\%} & 76.83\% / \textbf{65.68\%} & 85.53\% / \textbf{73.99\%} & 91.96\% / \textbf{80.96\%} & \textbf{96.19\%} / \textbf{86.70\%} & \textbf{98.60\%} / \textbf{93.70\%}\\ \hline
     T$\ell_1 (a=1.0)$ & 19.71\% / \textbf{17.01\%} & 37.17\% / \textbf{32.34\%} & 52.55\% / \textbf{45.74\%} & 65.70\% / \textbf{57.43\%} & 76.70\% / \textbf{67.21\%} & 85.44\% / \textbf{76.19\%} & \textbf{92.02\%}* / \textbf{83.11\%} & \textbf{96.38\%} / \textbf{88.58\%} & NA \\ \hline
     T$\ell_1 (a=0.5)$ & \textbf{19.99\%} / \textbf{16.20\%} & \textbf{37.52\%} / \textbf{31.27\%} & 52.70\% / \textbf{44.95\%} & 65.70\% / \textbf{57.29\%} & 76.68\% / \textbf{67.60\%} & 85.40\% / \textbf{76.66\%} & 91.98\% / \textbf{83.79\%} & \textbf{96.43\%}* / \textbf{89.53\%} & \textbf{98.73\%} / \textbf{94.41\%}\\ \hline
     MCP$(a=15000)$&
     \textbf{20.14\%} / 15.43\% & \textbf{37.86\%} / 29.52\% & \textbf{53.15\%} / 42.46\% & \textbf{66.23\%} / 53.76\% & \textbf{77.07\%} / 63.40\% & \textbf{85.67\%} / 71.68\% & 91.92\% / 78.98\% & 95.87\% / 84.33\% & \textbf{98.61\%} / \textbf{93.64\%}\\ \hline
     MCP$(a=10000)$ & \textbf{20.13\%} / 15.60\% & \textbf{37.91\%} / 29.62\% & \textbf{53.41\%} / 41.85\% & \textbf{66.40\%} / 52.96\% & \textbf{77.20\%} / 62.47\% & \textbf{85.72\%} / 70.84\% & 91.88\% / 77.91\% & 95.68\% / 83.63\% & NA\\ \hline
     MCP$(a=5000)$& \textbf{20.34\%}* / 14.91\% & \textbf{38.20\%}* / 28.64\% & \textbf{53.63\%} / 40.96\% & \textbf{66.72\%} / 51.97\% & \textbf{77.46\%} / 61.03\% & \textbf{85.76\%} / 68.51\% & 91.68\% / 75.01\% & 95.41\% / 82.79\% &  NA \\ \hline
     SCAD$(a=15000)$& 19.92\% / \textbf{16.25\%} & \textbf{37.55\%} / 30.27\% & \textbf{53.11\%} / 42.75\% & \textbf{66.19\%} / 54.23\% & \textbf{77.06\%} / 63.81\% & 85.57\% / 72.47\% & 91.91\% / 79.38\% & 95.88\% / 84.81\% &  NA\\ \hline
     SCAD$(a=10000)$& \textbf{19.97\%} / 15.32\% & \textbf{37.60\%} / 29.41\% & \textbf{53.22\%} / 41.88\% & \textbf{66.31\%} / 52.80\% & \textbf{77.15\%} / 62.58\% & \textbf{85.66\%} / 70.73\% & 91.81\% / 77.63\% & 95.65\% / 83.85\% &  NA \\ \hline
     SCAD$(a=5000)$&
     \textbf{20.28\%} / 15.07\% & \textbf{38.10\%} / 28.72\% & \textbf{53.66\%}* / 40.75\% & \textbf{66.82\%}* / 51.56\% & \textbf{77.50\%}* / 61.09\% & \textbf{85.79\%}* / 69.09\% & 91.73\% / 75.80\% & 95.47\% / 83.12\% & NA \\ \hline
\end{tabular}}
\end{minipage}}
\end{table*}
\begin{figure*}
\centering
    \includegraphics[width=\textwidth]{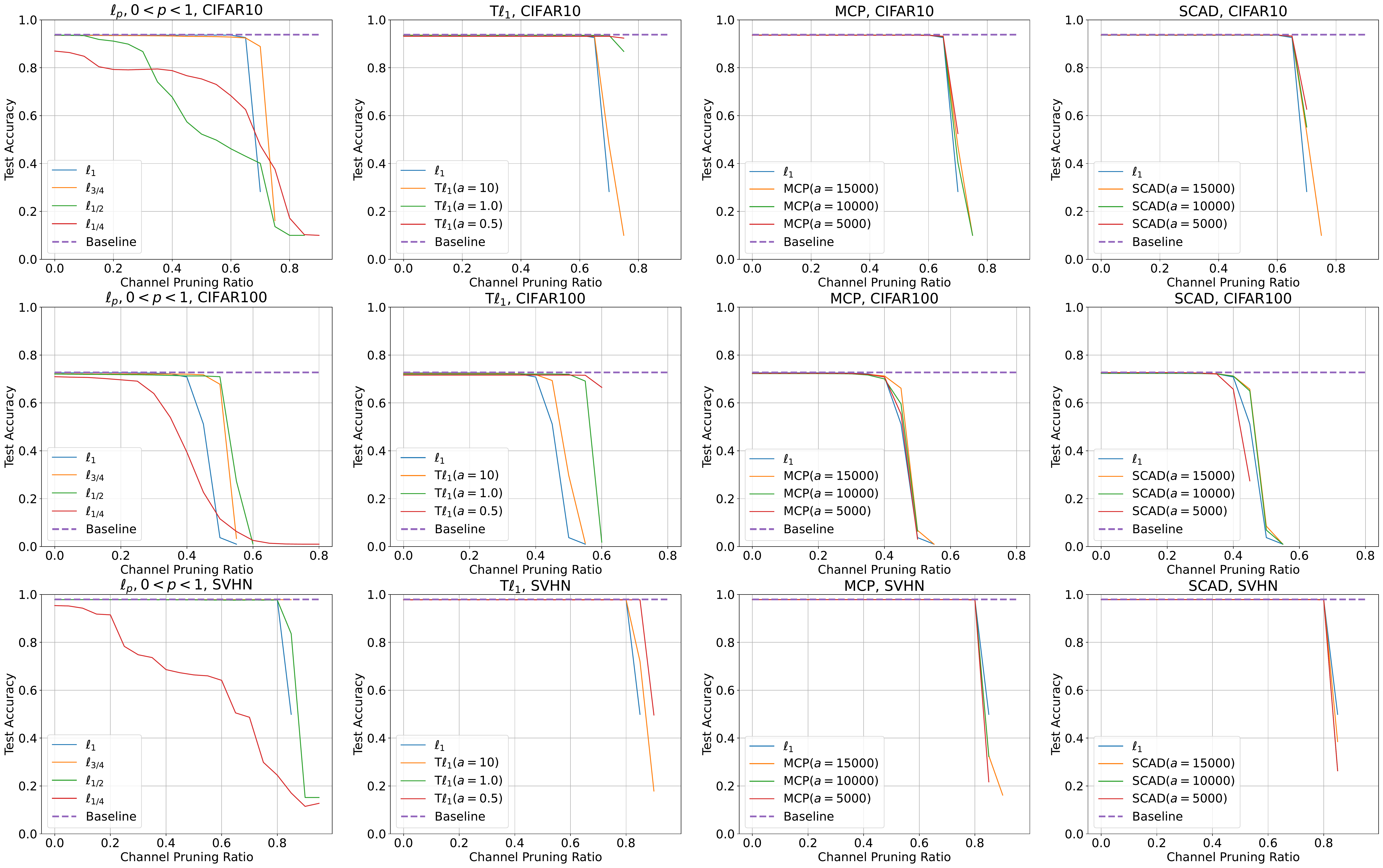}
    \caption{Effect of channel pruning on the mean test accuracy of five runs of VGG-19 on CIFAR 10/100 and SVHN. Baseline refers to the mean test accuracy of the unregularized model that is not pruned. Baseline accuracies are 93.83\% for CIFAR 10, 72.73\% for CIFAR 100, and 97.91\% for SVHN.}
    \label{fig:vgg_result}
\end{figure*}

\section{Experimental Results}
We apply the proposed nonconvex network slimming using $\ell_p (0<p<1)$, T$\ell_1$, MCP, and SCAD regularization on various networks and datasets and compare their results against the original network slimming with $\ell_1$ regularization as the baseline.

Code for the experiments is available at \url{https://github.com/kbui1993/NonconvexNetworkSlimming}. 

\subsection{Datasets}
\textbf{CIFAR 10/100.} The CIFAR 10/100 dataset \cite{krizhevsky2009learning} consists of 50k training color images and 10k test color images with 10/100 classes total. The resolution of each image is $32 \times 32$. To preprocess the dataset, we apply the data augmentation techniques (horizontal flipping and translation by 4 pixels) that have been standard in practice \cite{huang2016deep, lin2013network, goodfellow2013maxout, he2016deep, liu2017learning} followed by global contrast normalization and ZCA whitening \cite{goodfellow2013maxout}. These preprocessing techniques help improve the classification accuracy of CNNs on CIFAR 10 and 100 as demonstrated in \cite{goodfellow2013maxout, lin2013network}.

\textbf{SVHN.} The SVHN dataset \cite{Netzer2011ReadingDI} consists of $32 \times 32$ color images. The entire training set has 604,388 images and the test set has has 26,032 images. Before training on the dataset, each image is normalized by the channel means and standard deviations. 

We evaluate the proposed methods on VGG-19 \cite{simonyan2014very}, DenseNet-40 \cite{huang2017densely}, and ResNet-164 \cite{he2016deep}, three networks that were examined in \cite{liu2017learning}. More specifically, we use a variation of VGG-19 from \url{https://github.com/szagoruyko/cifar.torch}, a 40-layer DenseNet with a growth rate of 12, and a 164-layer pre-activation ResNet with a bottleneck structure.

\subsection{Implementation Details}
\textbf{Training the Network.} To perform a fair comparison between the original network slimming and the proposed nonconvex network slimming, we emulate most of the training settings in the original work \cite{liu2017learning}. All networks are trained from scratch using stochastic gradient descent.  The initial learning rate is set at 0.1, and it is reduced by a factor of 10 at the 50\% and 75\% of the total number of epochs. In addition, we use weight decay of $10^{-4}$ and Nesterov momentum \cite{sutskever2013importance} of 0.9 without dampening. On CIFAR 10/100, we train for 160 epochs, while on SVHN, we train for 20 epochs. On both datasets, the training batch size is 64. Weight initialization is based on \cite{he2015delving} and scaling factor initialization is set to 0.5 as done in \cite{liu2017learning}. We examine the following regularizers for network slimming: $\ell_1$, $\ell_p (p=0.25, 0.5, 0.75)$, T$\ell_1 (a=0.5, 1.0, 10.0)$, MCP $(a=5000, 10000, 15000)$, and SCAD $(a=5000,10000, 15000)$. The examined parameter values for these regularizers are chosen because they attain similar model accuracy as the baseline model without scaling factor regularization and they can prune a model by at least 40\% of its channels.  Lastly, we have the regularization parameter $\lambda = 10^{-4}$ for VGG-19 and DenseNet-40 and $\lambda = 5 \times 10^{-5}$ for ResNet-164. The regularization parameter is chosen by trying to balance between model accuracy and channel sparsity.  

\textbf{Pruning the Network.} After a model is trained, its channels are pruned globally. For example, we specify a channel pruning ratio to be 0.35 or a channel pruning percentage to be 35\% and determine the 35th percentile among all magnitudes of the scaling factors of the model. The 35th percentile is set as the threshold. Any channels whose scaling factors are below the threshold in magnitude are pruned. 

Since the channels are pruned globally, there is a threshold specific for each model: if the pruning ratio is above a certain value, a model becomes over-pruned. That is, the model cannot be used for inference because at least one of its layers has all of its channels removed. 

\textbf{Retraining the Network.} We retrain the pruned model without regularization on the scaling factors with the same optimization setting as the first time training it. The purpose of retraining is to at least recover the compressed model's original accuracy prior to pruning.

\textbf{Performance Metrics.} We compare the regularizers' performances based on test accuracy and compression of their respective models.

After pruning a network by its channels, we measure its compression by the remaining number of parameters and floating point operations (FLOPs). The number of parameters relates to the storage cost while the number of FLOPS relates to the computational cost. In our experiments, we report the following percentages:
\begin{align*}
\tiny
    \text{Percentage of parameters pruned}   = \left(1-\frac{\text{\# parameters remaining}}{\text{total \# network parameters} } \right)\times 100\%
\end{align*}
and
\begin{align*}
\tiny
        \text{Percentage of FLOPs pruned}   = \left(1-\frac{\text{\# FLOPs remaining}}{\text{total \# network FLOPs} } \right)\times 100\%.
\end{align*}

Since CNNs are highly nonconvex, each run of the same model and regularizer with the same hyperparameters will give a different result. Hence, we train each model of one regularizer five times and compute the mean. Therefore, the mean test accuracies and mean ratios/percentages of parameters/FLOPs pruned are computed from five runs each.

\subsection{Channel Pruning Results}
\textbf{VGG-19.} VGG-19 has about 20 million parameters and $7.97 \times 10^8$ FLOPs. Table \ref{tab:vgg_result} shows the relationships between channel pruning ratios and mean percentages of parameters/FLOPs pruned. Figure \ref{fig:vgg_result} shows the effect of channel pruning on mean test accuracies. 
\begin{table*}[p]
  \rotatebox{90}{\begin{minipage}{\textheight}
\caption{Effect of channel pruning on the mean pruned parameter / FLOPs percentages (\%) on DenseNet-40 trained on (a) CIFAR 10, (b) CIFAR 100, and (c) SVHN. The mean is computed from five runs for each regularizer. For each channel pruning ratio, \textbf{bold} indicates outperforming $\ell_1$; * indicates best value; and NA indicates at least one of the five models is over-pruned. }
\label{tab:DenseNet_result}
\scriptsize
\centering
\scalebox{0.95}{
\begin{tabular}{|c|c|c|c|c|c|c|c|c|c|}\hline
    (a) &\multicolumn{9}{c|}{CIFAR 10}  \\
     \hline
     \makecell{Channel Pruning \\ Ratio}& 0.10 & 0.20 & 0.30 & 0.40 & 0.50 & 0.60 & 0.70 & 0.80 & 0.90 \\ \hline
     $\ell_1$& 9.22\% / 8.40\% & 18.35\% / 16.63\% & 27.57\% / 24.91\% & 36.73\% / 33.02\% & 45.95\% / 41.49\% & 55.15\% / 49.75\% & 64.38\% / 58.10\% & 73.75\% / 68.18\% & 83.76\% / 79.75\%  \\ \hline
     $\ell_{3/4}$& \textbf{9.32}\% / \textbf{8.53\%} & \textbf{18.64\%} / \textbf{16.79\%} & \textbf{27.87\%} / \textbf{25.62\%} & \textbf{37.14\%} / \textbf{34.10\%} & \textbf{46.42\%} / \textbf{42.85\%} & \textbf{55.62\%} / \textbf{51.27\%} & \textbf{64.90\%} / \textbf{59.71\%} & \textbf{74.25\%} / \textbf{68.63\%} & \textbf{84.02\%} / \textbf{80.07\%} \\ \hline 
     $\ell_{1/2}$&  \textbf{9.33\%} / \textbf{8.65\%} & \textbf{18.59\%} / \textbf{17.08\%} & \textbf{27.97\%} / \textbf{25.96\%} & \textbf{37.26\%} / \textbf{34.71\%} & \textbf{46.62\%} / \textbf{43.33\%} & \textbf{55.88\%} / \textbf{51.85\%} & \textbf{65.12\%} / \textbf{60.32\%} & \textbf{74.47\%} / \textbf{69.14\%} & \textbf{84.36\%} / \textbf{80.13\%}\\ \hline
     $\ell_{1/4}$&  \textbf{9.35\%}* / \textbf{8.83\%}* & \textbf{18.71\%} / \textbf{17.63\%}* & \textbf{28.13\%}* / \textbf{26.39\%}* & \textbf{37.52\%}* / \textbf{35.27\%}* & \textbf{47.05\%}* / \textbf{44.74\%}* & \textbf{56.69\%}* / \textbf{54.33\%}* & \textbf{66.56\%}* / \textbf{64.34\%}* & \textbf{77.02\%}* / \textbf{75.42\%}* &  NA\\ \hline
     T$\ell_1 (a=10.0)$& 9.20\% / 8.31\% & 18.34\% / \textbf{16.83\%} & \textbf{27.59\%} / \textbf{25.32\%} & \textbf{36.82\%} / \textbf{33.65\%} & \textbf{46.08\%} / \textbf{41.97\%} & \textbf{55.27\%} / \textbf{50.17\%} & \textbf{64.54\%} / \textbf{58.19\%} & \textbf{73.89\%} / 68.01\% & \textbf{83.89\%} / 79.72\%\\ \hline
     T$\ell_1 (a=1.0)$ &  \textbf{9.35\%}* / \textbf{8.67\%} & \textbf{18.63\%} / \textbf{17.09\%} & \textbf{27.85\%} / \textbf{25.39\%} & \textbf{37.17\%} / \textbf{34.04\%} & \textbf{46.41\%} / \textbf{42.32\%} & \textbf{55.73\%} / \textbf{50.93}\% & \textbf{65.14\%} / \textbf{59.70\%} & \textbf{74.46\%} / \textbf{68.57\%} & \textbf{84.23\%} / \textbf{80.19\%} \\ \hline
     T$\ell_1 (a=0.5)$ & \textbf{9.35\%}* / \textbf{8.45\%} & \textbf{18.72\%}* / \textbf{16.99\%} & \textbf{28.08\%} / \textbf{25.82\%} & \textbf{37.39\%} / \textbf{34.47\%} & \textbf{46.73\%} / \textbf{43.13\%} & \textbf{56.16\%} / \textbf{52.18\%} & \textbf{65.49\%} / \textbf{60.60\%} & \textbf{74.88\%} / \textbf{69.28\%} & \textbf{84.45\%} / \textbf{80.70\%} \\ \hline
     MCP$(a=15000)$&
    9.19\% / 8.01\% & \textbf{18.37\%} / 16.21\% & \textbf{27.59\%} / 24.47\% & \textbf{36.79\%} / 32.96\% & \textbf{45.97\%} / 40.97\% & 55.15\% / 49.24\% & 64.35\% / 57.64\% & \textbf{73.77\%} / 68.13\% & 83.72\% / 79.37\%\\ \hline
     MCP$(a=10000)$ &\textbf{9.29\%} / 8.23\% & \textbf{18.45\%} / 16.28\% & \textbf{27.71\%} / 24.60\% & \textbf{36.93\%} / \textbf{33.05\%} & \textbf{46.07\%} / 41.26\% & \textbf{55.22\%} / 49.32\% & \textbf{64.40\%} / 57.57\% & \textbf{73.91\%} / \textbf{68.23\%} & \textbf{83.85\%} / 79.39\%\\ \hline
     MCP$(a=5000)$& 9.17\% / 8.19\% & 18.25\% / 16.11\% & 27.45\% / 24.25\% & 36.57\% / 32.22\% & 45.75\% / 40.39\% & 54.94\% / 48.56\% & 64.13\% / 56.70\% & 73.75\% / 67.59\% & \textbf{83.92\%} / 79.17\%  \\ \hline
     SCAD$(a=15000)$& 9.21\% / 8.11\% & \textbf{18.36\%} / 16.21\% & 27.54\% / 24.43\% & \textbf{36.75\%} / 32.57\% & 45.94\% / 40.90\% & 55.12\% / 49.18\% & \textbf{64.41\%} / 57.68\% & \textbf{73.85\%} / 68.10\% & \textbf{83.80\%} / 79.42\% \\ \hline
     SCAD$(a=10000)$& 9.18\% / 8.16\% & \textbf{18.36\%} / 16.54\% & \textbf{27.60\%} / 24.83\% & \textbf{36.77\%} / 32.77\% & 45.94\% / 41.05\% & 55.10\% / 49.04\% & 64.30\% / 57.21\% & \textbf{73.79\%} / 67.98\% & 83.75\% / 79.27\%  \\ \hline
     SCAD$(a=5000)$&
     9.06\% / 7.78\% & 18.22\% / 15.88\% & 27.40\% / 23.97\% & 36.54\% / 32.07\% & 45.66\% / 39.87\% & 54.87\% / 48.02\% & 64.08\% / 56.01\% & 73.71\% / 67.25\% & \textbf{83.84\%} / 78.76\% \\ \hline \hline
     (b)&\multicolumn{9}{c|}{CIFAR 100}  \\
     \hline
     \makecell{Channel Pruning \\ Ratio}& 0.10 & 0.20 & 0.30 & 0.40 & 0.50 & 0.60 & 0.70 & 0.80 & 0.90 \\ \hline
     $\ell_1$&9.18\% / 7.46\% & 18.34\% / 15.21\% & 27.53\% / 22.91\% & 36.69\% / 30.44\% & 45.84\% / 37.84\% & 54.98\% / 45.36\% & 64.12\% / 54.09\% & 73.39\% / 65.92\% &    \\ \hline
     $\ell_{3/4}$& \textbf{9.19\%} / \textbf{8.20\%} & \textbf{18.39\%} / \textbf{16.12\%} & \textbf{27.57\%} / \textbf{24.04\%} & \textbf{36.76\%} / \textbf{31.88\%} & \textbf{45.95\%} / \textbf{39.91\%} & \textbf{55.13\%} / \textbf{47.74\%} & \textbf{64.33\%} / \textbf{55.84\%} & \textbf{73.56\%} / \textbf{66.30\%} & \textbf{83.34\%} / \textbf{79.89\%}  \\ \hline 
     $\ell_{1/2}$& \textbf{9.20\%} / \textbf{8.23\%} & \textbf{18.41\%} / \textbf{16.41\%} & \textbf{27.62\%} / \textbf{24.37\%} & \textbf{36.85\%} / \textbf{32.34\%} & \textbf{46.06\%} / \textbf{40.58\%} & \textbf{55.26\%} / \textbf{48.91\%} & \textbf{64.44\%} / \textbf{56.97\%} & \textbf{73.67\%} / \textbf{66.98\%} & NA\\ \hline
     $\ell_{1/4}$&  \textbf{9.26\%}* / \textbf{8.33\%}* & \textbf{18.53\%}* / \textbf{16.76\%}* & \textbf{27.85\%}* / \textbf{25.00\%}* & \textbf{37.17\%}* / \textbf{33.70\%}* & \textbf{46.51\%}* / \textbf{43.03\%}* & \textbf{55.94\%}* / \textbf{52.75\%}* & \textbf{65.73\%}* / \textbf{63.59\%}* & \textbf{76.28\%}* / \textbf{76.02\%}* & NA \\ \hline
     T$\ell_1 (a=10.0)$& \textbf{9.19\%} / \textbf{7.80\%} & \textbf{18.35\%} / 15.19\% & \textbf{27.55\%} / \textbf{23.01\%} & \textbf{36.72\%} / \textbf{30.60\%} & \textbf{45.92\%} / \textbf{38.42\%} & \textbf{55.08\%} / \textbf{45.82\%} & \textbf{64.24\%} / 53.94\% & \textbf{73.49\%} / 65.90\% &  NA \\ \hline
     T$\ell_1 (a=1.0)$ & \textbf{9.26\%}* / \textbf{8.00\%} & \textbf{18.46\%} / \textbf{15.92\%} & \textbf{27.72\%} / \textbf{23.79\%} & \textbf{36.91\%} / \textbf{31.49\%} & \textbf{46.15\%} / \textbf{39.49\%} & \textbf{55.35\%} / \textbf{47.34\%} & \textbf{64.55\%} / \textbf{55.62\%} & \textbf{73.78\%} / \textbf{66.24\%} & \textbf{83.48\%} / \textbf{80.01\%} \\ \hline
     T$\ell_1 (a=0.5)$ &  \textbf{9.25\%} / \textbf{8.11\%} & \textbf{18.49\%} / \textbf{15.98\%} & \textbf{27.75\%} / \textbf{24.15\%} & \textbf{36.98\%} / \textbf{32.22\%} & \textbf{46.24\%} / \textbf{40.44\%} & \textbf{55.46\%} / \textbf{48.33\%} & \textbf{64.71\%} / \textbf{56.39\%} & \textbf{73.92\%} / \textbf{66.20\%} & \textbf{83.60\%}* / \textbf{80.15\%}* \\ \hline
     MCP$(a=15000)$&
    \textbf{9.19\%} / \textbf{7.72\%} & \textbf{18.35\%} / \textbf{15.52\%} & 27.52\% / \textbf{23.29\%} & 36.67\% / \textbf{30.99\%} & 45.81\% / \textbf{38.42\%} & \textbf{54.99\%} / \textbf{46.02\%} & \textbf{64.14\%} / \textbf{55.17\%} & \textbf{73.46\%} / \textbf{66.30\%} & \textbf{83.35\%} / \textbf{79.72\%} \\ \hline
     MCP$(a=10000)$ &9.16\% / \textbf{7.50\%} & 18.31\% / 15.09\% & 27.46\% / 22.84\% & 36.61\% / \textbf{30.61\%} & 45.79\% / \textbf{38.36\%} & 54.94\% / \textbf{45.92\%} & 64.10\% / \textbf{55.76\%} & 73.37\% / \textbf{66.94\%} & \textbf{83.19\%} / \textbf{79.68\%}\\ \hline
     MCP$(a=5000)$& 9.16\% / \textbf{7.53\%} & 18.32\% / 15.00\% & 27.46\% / 22.51\% & 36.64\% / 30.01\% & 45.78\% / \textbf{37.87\%} & 54.93\% / \textbf{46.00\%} & 64.12\% / \textbf{56.81\%} & \textbf{73.42\%} / \textbf{67.34\%} & \textbf{83.46\%} / \textbf{79.52\%}  \\ \hline
     SCAD$(a=15000)$& \textbf{9.19\%} / \textbf{7.85\%} & \textbf{18.36\%} / \textbf{15.50\%} & 27.52\% / \textbf{23.12\%} & 36.68\% / \textbf{30.65\%} & 45.84\% / \textbf{38.33\%} & \textbf{54.99\%} / \textbf{46.03\%} & \textbf{64.16\%} / \textbf{55.37\%} & \textbf{73.45\%} / \textbf{66.81\%} & \textbf{83.33\%} / \textbf{79.72\%}\\ \hline
     SCAD$(a=10000)$& 9.15\% / \textbf{7.72\%} & 18.30\% / \textbf{15.46\%} & 27.47\% / \textbf{23.15\%} & 36.63\% / \textbf{30.66\%} & 45.76\% / \textbf{38.43\%} & 54.94\% / \textbf{46.14\%} & \textbf{64.14\%} / \textbf{56.10\%} & \textbf{73.44\%} / \textbf{67.22\%} & \textbf{83.36\%} / \textbf{79.61\%} \\ \hline
     SCAD$(a=5000)$&
    9.15\% / 7.37\% & 18.31\% / 14.96\% & 27.44\% / 22.27\% & 36.59\% / 29.79\% & 45.76\% / 37.50\% & 54.91\% / \textbf{45.40\%} & 64.11\% / \textbf{55.93\%} & \textbf{73.42\%} / \textbf{67.19\%} & \textbf{83.53\%} / \textbf{79.75\%} \\ \hline \hline
     (c)&\multicolumn{9}{c|}{SVHN}  \\
     \hline
     \makecell{Channel Pruning \\ Ratio}& 0.10 & 0.20 & 0.30 & 0.40 & 0.50 & 0.60 & 0.70 & 0.80 & 0.90 \\ \hline
     $\ell_1$&9.51\% / 9.29\% & 19.12\% / 18.86\% & 28.61\% / 27.83\% & 38.25\% / 37.38\% & 47.84\% / 46.82\% & 57.48\% / 55.88\% & 67.09\% / 65.54\% & 76.67\% / 75.30\% & 86.15\% / 85.06\%  \\ \hline
     $\ell_{3/4}$&  \textbf{9.63\%} / \textbf{9.69\%} & \textbf{19.27\%} / \textbf{19.42\%} & \textbf{28.79\%} / \textbf{28.88\%} & \textbf{38.42\%} / \textbf{38.25\%} & \textbf{48.04\%} / \textbf{47.95\%} & \textbf{57.69\%} / \textbf{57.71\%} & \textbf{67.25\%} / \textbf{67.11\%} & \textbf{76.79\%} / \textbf{76.51\%} & \textbf{86.38\%} / \textbf{85.87\%}  \\ \hline 
     $\ell_{1/2}$& \textbf{9.62\%} / \textbf{9.38\%} & \textbf{19.21\%} / \textbf{19.20\%} & \textbf{28.81\%} / \textbf{28.68\%} & \textbf{38.44\%} / \textbf{38.46\%} & \textbf{48.08\%} / \textbf{47.85\%} & \textbf{57.75\%} / \textbf{57.54\%} & \textbf{67.43\%} / \textbf{67.40\%} & \textbf{77.05\%} / \textbf{76.96\%} & \textbf{86.68\%} / \textbf{86.41\%} \\ \hline
     $\ell_{1/4}$&  \textbf{9.68\%}* / \textbf{9.88\%}* & \textbf{19.34\%} / \textbf{19.46\%}* & \textbf{29.05\%}* / \textbf{29.40\%}* & \textbf{38.74\%}* / \textbf{39.33\%}* & \textbf{48.42\%}* / \textbf{49.00\%}* & \textbf{58.12\%}* / \textbf{58.71\%}* & \textbf{67.92\%}* / \textbf{68.69\%}* & \textbf{77.81\%}* / \textbf{78.96\%}* & \textbf{87.81\%}* / \textbf{89.44\%}*\\ \hline
     T$\ell_1 (a=10.0)$& \textbf{9.57\%} / \textbf{9.48\%} & \textbf{19.13\%} / \textbf{19.05\%} & \textbf{28.72\%} / \textbf{28.73\%} & \textbf{38.36\%} / \textbf{38.13\%} & \textbf{47.87\%} / \textbf{47.49\%} & \textbf{57.51\%} / \textbf{56.91\%} & 67.06\% / \textbf{66.32\%} & 76.64\% / \textbf{75.74\%} & \textbf{86.29\%} / \textbf{85.54\%}   \\ \hline
     T$\ell_1 (a=1.0)$ &\textbf{9.58\%} / \textbf{9.33\%} & \textbf{19.24\%} / \textbf{19.26\%} & \textbf{28.92\%} / \textbf{28.77\%} & \textbf{38.58\%} / \textbf{38.59\%} & \textbf{48.20\%} / \textbf{48.03\%} & \textbf{57.82\%} / \textbf{57.66\%} & \textbf{67.44\%} / \textbf{66.97\%} & \textbf{77.01\%} / \textbf{76.50\%} & \textbf{86.57\%} / \textbf{86.17\%} \\ \hline
     T$\ell_1 (a=0.5)$ &   \textbf{9.62\%} / 9.29\% & \textbf{19.19\%} / 18.82\% & \textbf{28.81\%} / \textbf{28.37\%} & \textbf{38.51\%} / \textbf{37.98\%} & \textbf{48.16\%} / \textbf{47.64\%} & \textbf{57.83\%} / \textbf{57.76\%} & \textbf{67.46\%} / \textbf{67.27\%} & \textbf{77.03\%} / \textbf{77.01\%} & \textbf{86.70\%} / \textbf{86.55\%}\\ \hline
     MCP$(a=15000)$&
    \textbf{9.65\%} / \textbf{9.52\%} & \textbf{19.31\%} / \textbf{19.09\%} & \textbf{28.89\%} / \textbf{28.73\%} & \textbf{38.40\%} / \textbf{38.03\%} & \textbf{47.88\%} / \textbf{47.53\%} & 57.44\% / \textbf{56.81\%} & 67.05\% / \textbf{66.62\%} & 76.60\% / \textbf{76.05\%} & NA \\ \hline
     MCP$(a=10000)$ &9.51\% / \textbf{9.42\%} & 19.02\% / \textbf{18.92\%} & 28.60\% / \textbf{28.36\%} & 38.22\% / \textbf{37.67\%} & 47.73\% / \textbf{47.15\%} & 57.26\% / \textbf{56.59\%} & 66.95\% / \textbf{65.99\%} & 76.61\% / \textbf{75.71\%} & 86.14\% / \textbf{85.29\%} \\ \hline
     MCP$(a=5000)$& \textbf{9.55\%} / \textbf{9.44\%} & \textbf{19.14\%} / \textbf{18.89\%} & \textbf{28.70\%} / \textbf{28.36\%} & 38.25\% / \textbf{37.66\%} & \textbf{47.89\%} / \textbf{47.26\%} & 57.48\% / \textbf{56.57\%} & 67.02\% / \textbf{66.10\%} & 76.58\% / \textbf{75.69\%} & 86.10\% / 84.97\%  \\ \hline
     SCAD$(a=15000)$& \textbf{9.55\%} / \textbf{9.31\%} & 19.09\% / 18.75\% & \textbf{28.71\%} / \textbf{28.52\%} & \textbf{38.26\%} / \textbf{38.02\%} & 47.84\% / \textbf{47.30\%} & 57.47\% / \textbf{56.49\%} & \textbf{67.13\%} / \textbf{66.27\%} & 76.57\% / \textbf{75.62\%} &NA \\ \hline
     SCAD$(a=10000)$& \textbf{9.66\%} / \textbf{9.82\%} & \textbf{19.37\%}* / \textbf{19.25\%} & \textbf{28.88\%} / \textbf{28.99\%} & \textbf{38.46\%} / \textbf{38.36\%} & \textbf{48.06\%} / \textbf{47.87\%} & \textbf{57.53\%} / \textbf{57.31\%} & \textbf{67.13\%} / \textbf{66.95\%} & 76.61\% / \textbf{76.55\%} & NA \\ \hline
     SCAD$(a=5000)$&
    \textbf{9.55\%} / \textbf{9.31\%} & 19.09\% / 18.75\% & \textbf{28.71\%} / \textbf{28.52\%} & \textbf{38.26\%} / \textbf{38.02\%} & 47.84\% / \textbf{47.30\%} & 57.47\% / \textbf{56.49\%} & \textbf{67.13\%} / \textbf{66.27\%} & 76.57\% / \textbf{75.62\%} & NA  \\ \hline
\end{tabular}
}\end{minipage}}
\end{table*}

\begin{figure*}
\centering
    \includegraphics[width=\textwidth]{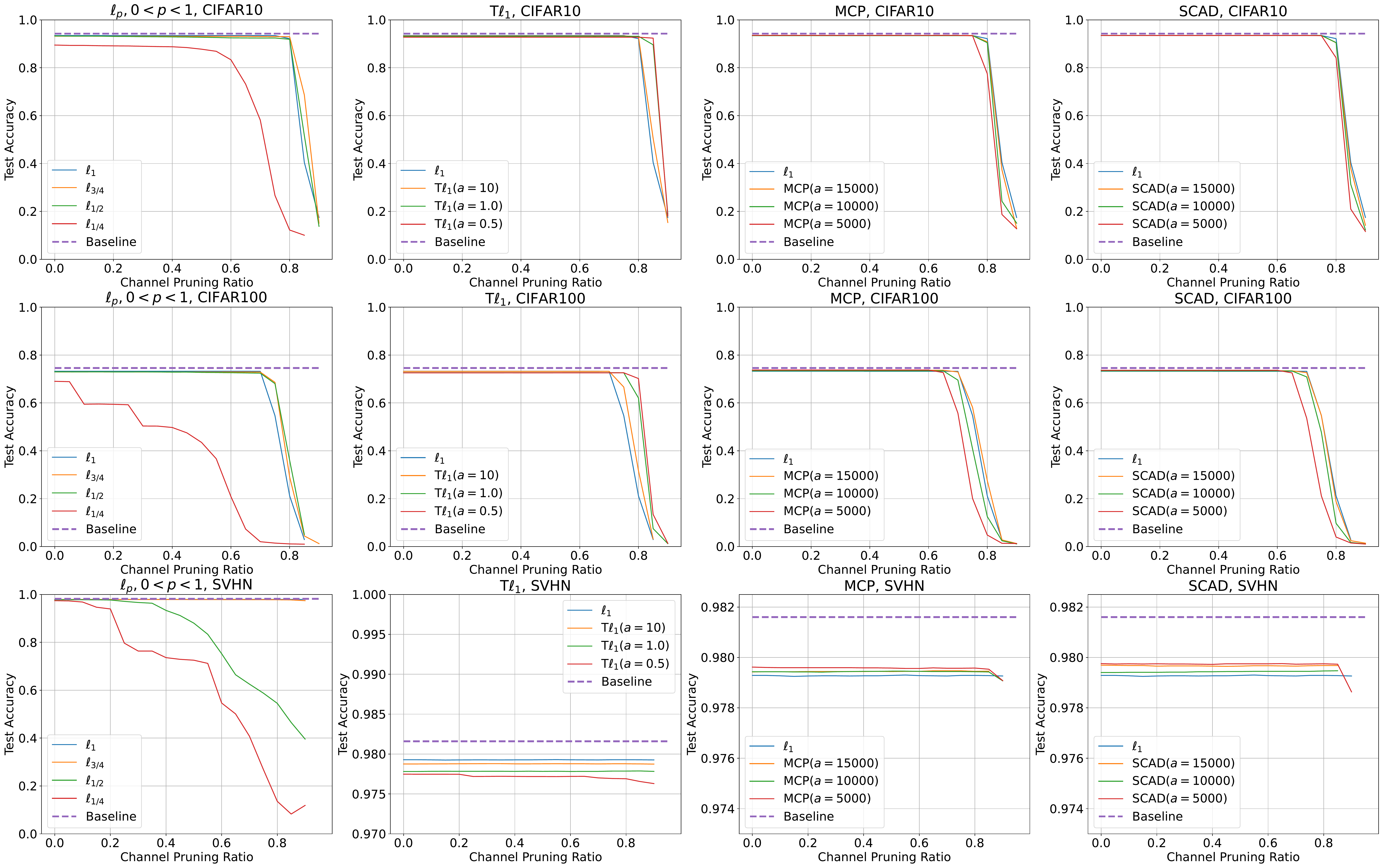}
    \caption{Effect of channel pruning on the mean test accuracy of five runs of DenseNet-40 on CIFAR 10/100 and SVHN. Baseline refers to the mean test accuracy of the unregularized model that is not pruned. Baseline accuracies are 94.25\% for CIFAR 10, 74.58\% for CIFAR 100, and 98.16\% for SVHN.}
    \label{fig:densenet}
\end{figure*}
\begin{table*}[p]
  \rotatebox{90}{\begin{minipage}{\textheight}
\caption{Effect of channel pruning on the mean pruned parameter / FLOPs percentages (\%) on ResNet-164 trained on (a) CIFAR 10, (b) CIFAR 100, and (c) SVHN. The mean is computed from five runs for each regularizer. For each channel pruning ratio, \textbf{bold} indicates outperforming $\ell_1$; * indicates best value; and NA indicates at least one of the five models is over-pruned. }
\label{tab:res}
\centering
\scriptsize
\scalebox{0.975}{
\begin{tabular}{|c|c|c|c|c|c|c|c|}\hline
    (a) &\multicolumn{7}{c|}{CIFAR 10}  \\
     \hline
     \makecell{Channel Pruning \\ Ratio}& 0.10 & 0.20 & 0.30 & 0.40 & 0.50 & 0.60 & 0.70  \\ \hline
     $\ell_1$& 8.57\% / 8.37\% & 16.67\% / 16.62\% & 24.44\% / 24.29\% & 31.39\% / 31.57\% & 38.50\% / 38.31\% & 46.43\% / 45.54\% & NA  \\ \hline
     $\ell_{3/4}$& \textbf{10.87\%} / \textbf{10.07\%} & \textbf{21.51\%} / \textbf{19.54\%} & \textbf{31.23\%} / \textbf{28.32\%} & \textbf{40.07\%} / \textbf{36.66\%} & \textbf{47.86\%} / \textbf{43.96\%} & \textbf{55.15\%} / \textbf{50.71\%} & \textbf{62.88\%} / \textbf{58.36\%} \\ \hline 
     $\ell_{1/2}$& \textbf{12.13\%} / \textbf{10.96\%} & \textbf{22.88\%} / \textbf{21.41\%} & \textbf{33.23\%} / \textbf{31.09\%} & \textbf{42.54\%} / \textbf{39.81\%} & \textbf{50.88\%} / \textbf{48.02\%} & \textbf{58.17\%} / \textbf{55.12\%} & \textbf{64.88\%} / \textbf{61.46\%} \\ \hline
     $\ell_{1/4}$&  \textbf{14.26\%}* / \textbf{13.00\%}* & \textbf{26.44\%}* / \textbf{24.50\%}* & \textbf{37.64\%}* / \textbf{34.93\%}* & \textbf{47.82\%}* / \textbf{44.89\%}* & \textbf{57.10\%}* / \textbf{54.47\%}* & \textbf{65.58\%}* / \textbf{64.27\%}* & NA \\ \hline
     T$\ell_1 (a=10.0)$&\textbf{8.99\%} / \textbf{8.56\%} & \textbf{17.92\%} / \textbf{16.88\%} & \textbf{25.80\%} / 24.18\% & \textbf{33.26\%} / 31.14\% & \textbf{40.50\%} / 38.27\% & \textbf{47.52\%} / 44.88\% & NA  \\ \hline
     T$\ell_1 (a=1.0)$ & \textbf{11.99\%} / \textbf{10.74\%} & \textbf{22.86\%} / \textbf{20.72\%} & \textbf{33.08\%} / \textbf{29.61\%} & \textbf{42.64\%} / \textbf{38.44\%} & \textbf{51.03\%} / \textbf{46.02\%} & \textbf{58.52\%} / \textbf{53.03\%} & NA \\ \hline
     T$\ell_1 (a=0.5)$ &   \textbf{12.51\%} / \textbf{11.29\%} & \textbf{23.95\%} / \textbf{21.61\%} & \textbf{34.43\%} / \textbf{31.01\%} & \textbf{44.10\%} / \textbf{39.82\%} & \textbf{52.93\%} / \textbf{47.65\%} & \textbf{60.81\%} / \textbf{54.86\%} & \textbf{67.15\%} / \textbf{61.20\%}\\ \hline
     MCP$(a=15000)$&
    8.07\% / 7.90\% & 16.00\% / 15.54\% & 23.46\% / 22.69\% & 30.50\% / 29.08\% & 36.84\% / 35.05\% & \textbf{47.73\%} / \textbf{45.63\%} & NA\\ \hline
     MCP$(a=10000)$ &7.10\% / 7.61\% & 13.90\% / 14.43\% & 20.58\% / 21.06\% & 26.87\% / 27.32\% & 32.74\% / 33.29\% & NA  & NA \\ \hline
     MCP$(a=5000)$&4.19\% / 5.55\% & 8.64\% / 10.97\% & 12.85\% / 16.08\% & 17.06\% / 21.00\% & 23.89\% / 29.34\% & NA   & NA \\ \hline
     SCAD$(a=15000)$& 7.71\% / 7.65\% & 15.53\% / 15.44\% & 22.58\% / 22.83\% & 29.51\% / 29.44\% & 36.44\% / 35.82\% & \textbf{47.47\%} / \textbf{46.15\%} & NA \\ \hline
     SCAD$(a=10000)$& 7.19\% / 7.33\% & 13.99\% / 14.30\% & 20.51\% / 20.88\% & 26.71\% / 27.26\% & 32.79\% / 33.27\% & NA & NA  \\ \hline
     SCAD$(a=5000)$&
   4.62\% / 5.68\% & 8.98\% / 11.02\% & 13.24\% / 16.23\% & 17.45\% / 21.35\% & 23.94\% / 29.47\% & NA  &NA\\ \hline
   \hline
    (b) &\multicolumn{7}{c|}{CIFAR 100}  \\
     \hline
     \makecell{Channel Pruning \\ Ratio}& 0.10 & 0.20 & 0.30 & 0.40 & 0.50 &0.60 & 0.70\\ \hline
     $\ell_1$& 4.01\% / 7.42\% & 7.98\% / 14.48\% & 11.88\% / 20.94\% & 15.72\% / 26.88\% & NA & NA & NA\\ \hline
     $\ell_{3/4}$& \textbf{4.95\%} / \textbf{7.55\%} & \textbf{9.84\%} / \textbf{14.90\%} & \textbf{14.70\%} / \textbf{22.08\%} & \textbf{19.15\%} / \textbf{28.15\%} & \textbf{24.58\%} / \textbf{35.33\%} & NA & NA \\ \hline 
     $\ell_{1/2}$& \textbf{5.72\%} / \textbf{8.53\%} & \textbf{11.35\%} / \textbf{16.51\%} & \textbf{16.66\%} / \textbf{23.82\%} & \textbf{21.86\%} / \textbf{30.73\%} & \textbf{26.64\%} / \textbf{36.87\%} & NA & NA\\ \hline
     $\ell_{1/4}$&  \textbf{11.13\%}* / \textbf{11.46\%}* & \textbf{20.98\%}* / \textbf{21.85\%}* & \textbf{30.00\%}* / \textbf{31.48\%}* & \textbf{37.85\%}* / \textbf{41.10\%}* & NA & NA &NA\\ \hline
     T$\ell_1 (a=10.0)$&\textbf{4.08\%} / 7.07\% & \textbf{8.29\%} / 13.87\% & \textbf{12.36\%} / 20.03\% & \textbf{16.36\%} / 25.92\% & NA  & NA & NA\\ \hline
     T$\ell_1 (a=1.0)$ & \textbf{6.08\%} / \textbf{8.09\%} & \textbf{11.96\%} / \textbf{15.67\%} & \textbf{17.38\%} / \textbf{22.93\%} & \textbf{22.99\%} / \textbf{29.81\%} & \textbf{28.27\%} / \textbf{36.36\%} & NA& NA\\ \hline
     T$\ell_1 (a=0.5)$ &  \textbf{6.37\%} / \textbf{9.25\%} & \textbf{12.68\%} / \textbf{17.30\%} & \textbf{18.82\%} / \textbf{25.19\%} & \textbf{24.87\%} / \textbf{31.89\%} & \textbf{30.54\%}* / \textbf{38.63\%}* & NA&NA\\ \hline
     MCP$(a=15000)$&
    3.64\% / 6.64\% & 7.25\% / 12.89\% & 10.92\% / 18.99\% & 15.22\% / 25.05\% & NA& NA &NA\\ \hline
     MCP$(a=10000)$ &3.51\% / 6.65\% & 7.01\% / 12.53\% & 10.42\% / 18.48\% & 14.45\% / 24.81\% & NA & NA & NA\\ \hline
     MCP$(a=5000)$&3.32\% / 6.37\% & 6.52\% / 12.13\% & 9.67\% / 17.58\% & NA &NA &NA &NA\\ \hline
     SCAD$(a=15000)$& 3.62\% / 6.56\% & 7.20\% / 13.01\% & 10.88\% / 19.36\% & 15.16\% / 25.79\% & NA& NA& NA\\ \hline
     SCAD$(a=10000)$& 3.53\% / 6.36\% & 6.99\% / 12.61\% & 10.36\% / 18.46\% & 14.54\% / 25.33\% & NA  &NA & NA \\ \hline
     SCAD$(a=5000)$&
   3.31\% / 5.92\% & 6.59\% / 11.83\% & 9.77\% / 17.33\% & 14.15\% / 25.57\% &NA &NA &NA\\ \hline
   \hline
     (c) &\multicolumn{7}{c|}{SVHN}  \\
     \hline
     \makecell{Channel Pruning \\ Ratio}& 0.10 & 0.20 & 0.30 & 0.40 & 0.50 & 0.60  &0.70 \\ \hline
     $\ell_1$& 12.32\% / 17.02\%* & 22.70\% / 29.19\% & 32.63\% / 41.26\% & 41.88\% / 52.39\% & 50.14\% / 62.14\% & NA  &NA \\ \hline
     $\ell_{3/4}$& \textbf{13.09\%} / 15.50\% & \textbf{25.49\%} / \textbf{29.87\%} & \textbf{36.84\%} / \textbf{42.16\%} & \textbf{47.02\%} / \textbf{53.46\%} & \textbf{55.77\%} / \textbf{63.17\%} &  NA&NA\\ \hline 
     $\ell_{1/2}$& \textbf{13.80\%} / 15.21\% & \textbf{26.62\%} / \textbf{29.57\%} & \textbf{38.20\%} / \textbf{42.21\%} & \textbf{48.80\%} / \textbf{53.60\%} & \textbf{58.45\%} / \textbf{63.91\%}* & \textbf{66.65\%} / \textbf{72.62\%} &NA\\ \hline
     $\ell_{1/4}$&  \textbf{15.16\%}* / 15.61\% & \textbf{29.05\%}* / \textbf{29.73\%} & \textbf{41.52\%}* / \textbf{42.43\%} & \textbf{52.47\%}* / \textbf{53.73\%}* & \textbf{62.39\%}* / \textbf{63.68\%} & \textbf{71.50\%}* / \textbf{72.79\%}* & NA \\ \hline
     T$\ell_1 (a=10.0)$&12.13\% / 16.66\% & \textbf{23.13\%} / \textbf{30.10\%}* & \textbf{33.11\%} / \textbf{41.50\%} & \textbf{42.70\%} / \textbf{52.97\%} & \textbf{50.87\%} / \textbf{62.07\%} & \textbf{58.16\%} / \textbf{69.81\%}  & NA\\ \hline
     T$\ell_1 (a=1.0)$ & \textbf{13.45\%} / 15.39\% & \textbf{25.82\%} / \textbf{29.90\%} & \textbf{37.29\%} / \textbf{42.59\%} & \textbf{47.70\%} / \textbf{53.87\%} & \textbf{56.79\%} / \textbf{63.43\%} &  NA &NA\\ \hline
     T$\ell_1 (a=0.5)$ & \textbf{14.35\%} / 15.83\% & \textbf{26.94\%} / \textbf{29.53\%} & \textbf{38.69\%} / \textbf{42.68\%}* & \textbf{48.83\%} / \textbf{53.70\%} & \textbf{58.31\%} / \textbf{63.81\%} & \textbf{66.44\%} / \textbf{72.28\%}&NA\\ \hline
     MCP$(a=15000)$&
    12.07\% / 15.25\% & \textbf{23.19\%} / 28.99\% & \textbf{32.89\%} / 40.96\% & 41.67\% / 51.50\% & 49.89\% / 60.89\% & \textbf{57.23\%} / \textbf{68.84\%}&NA\\ \hline
     MCP$(a=10000)$ &11.39\% / 15.19\% & 22.09\% / 28.56\% & 32.33\% / 40.67\% & 41.32\% / 51.23\% & 49.08\% / 60.14\% & NA& NA\\ \hline
     MCP$(a=5000)$&9.90\% / 13.98\% & 19.13\% / 26.99\% & 27.85\% / 38.51\% & 35.80\% / 48.73\% & 43.23\% / 57.77\% & NA& NA\\ \hline
     SCAD$(a=15000)$& 11.45\% / 15.70\% & 22.01\% / 28.82\% & 32.14\% / 40.65\% & 41.05\% / 51.61\% & 49.47\% / 61.02\% & \textbf{56.76\%} / \textbf{68.83\%} & NA\\ \hline
     SCAD$(a=10000)$& 12.30\% / 16.86\% & 22.63\% / 29.36\% & 32.39\% / 40.89\% & 41.23\% / 51.75\% & NA & NA & NA \\ \hline
     SCAD$(a=5000)$&
   10.42\% / 15.04\% & 19.82\% / 27.80\% & 28.52\% / 38.81\% & 36.76\% / 49.44\% & NA & NA & NA\\ \hline
\end{tabular}}
\end{minipage}}
\end{table*}
\begin{figure*}
\centering
    \includegraphics[width=\textwidth]{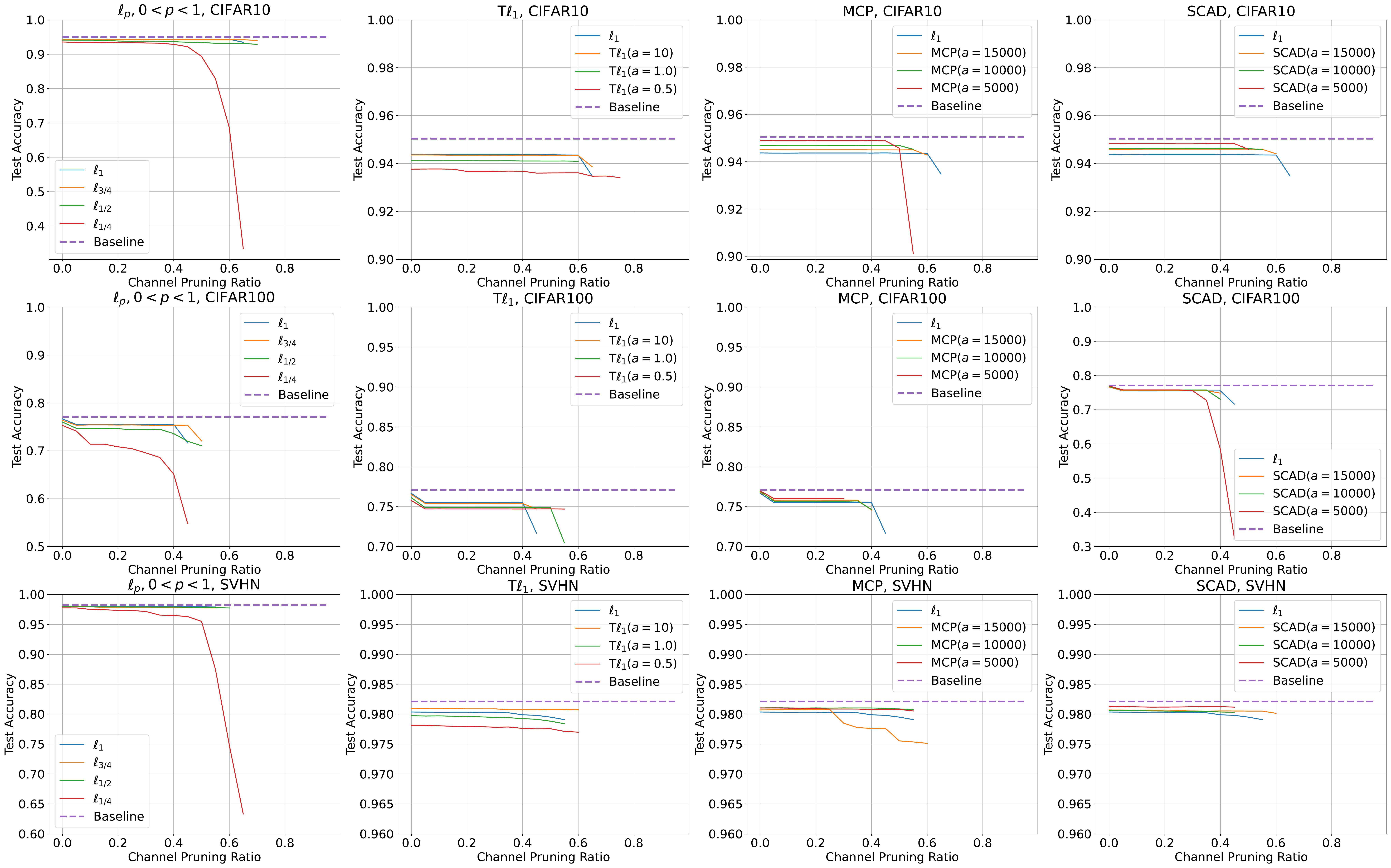}
    \caption{Effect of channel pruning on the mean test accuracy of five runs of ResNet-164 on CIFAR 10/100 and SVHN. Baseline refers to the mean test accuracy of the unregularized model that is not pruned. Baseline accuracies are 95.04\% for CIFAR 10, 77.10\% for CIFAR 100, and 98.21\% for SVHN.}
    \label{fig:resnet}
\end{figure*}

On CIFAR 10, according to Table \ref{tab:vgg_result}a, most of the nonconvex regularizers prune more parameters than $\ell_1$ up to channel pruning ratio 0.50. Although more parameters are pruned, MCP and SCAD require more FLOPs in general compared to $\ell_1$. On the other hand, $\ell_p$ and T$\ell_1$ outperform $\ell_1$ with respect to percentages of parameters/FLOPs pruned for channel pruning ratio at least 0.60. Additionally, the models trained with $\ell_{1/2}$ and $\ell_{3/4}$ can have at least 80\% of its channels pruned and still be used for inference even though their test accuracies are low. However, their test accuracies can be improved if the models were retrained. According to Figure \ref{fig:vgg_result}, $\ell_{3/4}$, T$\ell_1$, MCP, and SCAD are more robust than $\ell_1$ to channel pruning since their accuracies drop at higher channel pruning ratios. Although both $\ell_{1/2}$ and $\ell_{1/4}$ compress the model significantly compared to other regularizers, they are very sensitive to channel pruning.

On CIFAR 100,  according to Table \ref{tab:vgg_result}b, $\ell_p$ and T$\ell_1 (a=0.5, 1.0)$ require less parameters and FLOPs compared to $\ell_1$ when the channel pruning ratios are at least 0.40. MCP and SCAD have comparable number of parameters and FLOPs pruned as $\ell_1$. Figure \ref{fig:vgg_result} shows that T$\ell_1$ is robust against channel pruning, especially when $a=0.5$. At channel pruning ratio 0.6, the accuracy for T$\ell_1 (a=0.5)$ does not drop as much compared to other values of $a$ and also other nonconvex regularizers. For the other regularizers, $\ell_1$ is outperformed by $\ell_{3/4}$, $\ell_{1/2}$, MCP, and SCAD $(a=10000, 15000)$. Like for CIFAR 10, models trained with either $\ell_{1/2}$ or $\ell_{1/4}$ are still sensitive to channel pruning.

Lastly, for SVHN, according to Table \ref{tab:vgg_result}c, MCP and SCAD generally outperform $\ell_1$ in parameter pruning percentages for channel pruning ratios up to 0.60, but they do not save more on FLOPs. However, FLOPs are reduced more by $\ell_p$ and T$\ell_1$ in general across all channel pruning ratios. By Figure \ref{fig:vgg_result}, $\ell_{3/4}, \ell_{1/2}$, and T$\ell_1$ have higher test accuracies than $\ell_1$ when the channel pruning ratio is at 0.85. 

In general, nonconvex regularizers save more on parameters, FLOPs, or both. It is important to note that T$\ell_1$, especially $a=0.5$, helps preserve model accuracy against channel pruning, and $\ell_{1/4}$ is very sensitive to channel pruning.

\textbf{DenseNet-40.} DenseNet-40 has about 1 million parameters and $5.33 \times 10^8$ FLOPs. Table \ref{tab:DenseNet_result} shows the relationships between channel pruning ratios and mean percentages of parameters/FLOPs pruned. Figure \ref{fig:densenet} shows the effect of channel pruning on mean test accuracies. 

On CIFAR 10, by Table \ref{tab:DenseNet_result}a,  $\ell_p$ and T$\ell_1$ compress the model more in terms of number of parameters and FLOPs than $\ell_1$ after channel pruning across the various levels of channel pruning ratios. In general, MCP and SCAD require slightly more FLOPS than $\ell_1$, but they require similar number of parameters as $\ell_1$. According to Figure \ref{fig:densenet}, $\ell_p (p=1/2,3/4)$ and T$\ell_1$ are more robust to channel pruning than $\ell_1$ since their accuracies drop at higher channel pruning ratios, while MCP and SCAD are worse.

For CIFAR 100, Table \ref{tab:DenseNet_result}b demonstrates that $\ell_p$ and T$\ell_1$ generally reduce more parameters and FLOPs required than $\ell_1$ after channel pruning. At channel pruning ratios 0.60 and above, MCP and SCAD reduce only more FLOPs than $\ell_1$. In addition, models with MCP and SCAD regularization remain usable for inference after 90\% of their channels are pruned, unlike models with $\ell_1$ regularization. However, their test accuracies are unacceptable so that the models will need to be retrained to recover its original accuracies. According to Figure \ref{fig:densenet},  $\ell_p (p=1/2,3/4)$, T$\ell_1$, and MCP $(a=15000)$ are more robust to channel pruning than $\ell_1$ because their test accuracies drop at higher channel pruning ratios than $\ell_1$'s.

For SVHN, Table \ref{tab:DenseNet_result}c shows that $\ell_p$ and T$\ell_1$ have larger parameter/FLOPs pruning percentages than $\ell_1$ across different levels of the channel pruning ratios. In general, MCP also saves more on parameters and FLOPs for channel pruning ratio up to 0.50. After 0.50, MCP saves more on only FLOPs. SCAD also generally saves more on FLOPs than $\ell_1$. According to Figure \ref{fig:densenet}, the test accuracy remains nearly constant for channel pruning ratio up to 0.90 for all regularizers except for $\ell_{1/4}$ and $\ell_{1/2}$. We also observe that across different channel pruning ratios, T$\ell_1$ has slightly worse test accuracy than $\ell_1$ while MCP and SCAD mostly have better test accuracies than $\ell_1$. 

In summary, we observe that $\ell_p$ and T$\ell_1$ reduce more parameters and FLOPs required than $\ell_1$ after channel pruning, while MCP and SCAD save more on only FLOPs specifically for CIFAR 100 and SVHN. Like for VGG-19, T$\ell_1 (a=0.5)$ is the most robust against channel pruning, whereas $\ell_{1/4}$ is the most sensitive to it.

\textbf{ResNet-164.} ResNet-164 has about 1.70 million parameters and requires $5.00 \times 10^8$ FLOPs. Table \ref{tab:res} records the mean percentages of parameters/FLOPs pruned for different channel pruning ratios. Figure \ref{fig:resnet} shows the effect of channel pruning on the test accuracies of the regularized models.

On CIFAR 10, Table \ref{tab:res}a shows a quite noticeable difference in the numbers of parameters and FLOPs pruned between $\ell_1$ and $\ell_p$ or T$\ell_1 (a=0.5, 1.0)$. For example, $\ell_{1/2}$ saves at least 10\% more weight parameters and at least 8\% more FLOPs than $\ell_1$ at channel pruning ratio 0.40 and above. On the other hand, SCAD and MCP are outperformed by $\ell_1$ in percentages of parameters/FLOPs pruned. According to Figure \ref{fig:resnet}, most of the regularizers do not suffer a significant drop in test accuracy when large number of channels are pruned. 

On CIFAR 100, according to Table \ref{tab:res}b, $\ell_p (p=1/4, 1/2)$ and T$\ell_1(a=0.5, 1.0)$ prune at least 3\% more parameters and at least 1\% more FLOPs than $\ell_1$. However, MCP and SCAD are outperformed by $\ell_1$ again for percentages of parameters and FLOPs pruned. In Figure \ref{fig:resnet}, we observe that most of the regularizers are robust against channel pruning since the test accuracies do not drop severely at higher channel pruning ratios.

On SVHN, Table \ref{tab:res}c reports that $\ell_p$ and T$\ell_1$ save more parameters and FLOPs than $\ell_1$ for channel pruning ratio at least 0.20, while that MCP and SCAD do not. Like for CIFAR 10 and CIFAR 100, most regularizers yield models whose test accuracies are robust against channel pruning according to Figure \ref{fig:resnet}. 

In general, the test accuracies of ResNet-164 models with any regularizers, except for $\ell_{1/4}$, are stable against channel pruning. In addition, $\ell_{p}$ and T$\ell_1 (a=0.5,1.0)$ prune more parameters and FLOPs than $\ell_1$. Overall, MCP and SCAD do not perform well on ResNet-164.

\subsection{Retraining After Pruning}
\begin{table*}[h!!]
    \caption{Results from five retrained VGG-19 on CIFAR 10/100 after pruning. Baseline refers to the VGG-19 model trained without regularization on the scaling factors.}
        \label{tab:vgg_retrain}
\centering
\begin{subtable}[h]{\textwidth}
\centering
\fontsize{6}{6}\selectfont
    \begin{tabular}{l|c|c|c|c}
     & Number of Parameters/FLOPs & Percentage of Parameters/FLOPs Pruned (\%) & \makecell{Mean Test Accuracy\\
     before Retraining (\%)}& \makecell{Mean Test Accuracy\\
     after Retraining (\%)}\\
     \hline
     Baseline & 20.04M/$7.97\times 10^8$ & 0.00/0.00 & 93.83 & N/A\\ \hline \hline 
     $\ell_1$ (0\% Pruned)  & 20.04M/$7.97\times 10^8$ & 0.00/0.00 & 93.63 & N/A \\
     $\ell_1$ (70\% Pruned) & 2.24M/$3.83\times 10^8$& 88.81/51.93 & 28.28 &93.91 \\
     \hline \hline
     $\ell_{3/4}$ (0\% Pruned)  & 20.04M/$7.97\times 10^8$ & 0.00/0.00 & 93.53& N/A\\
     $\ell_{3/4}$ (70\% Pruned) & 2.07M/$3.59\times 10^8$ & 89.69/54.96 & 88.87 &93.90\\
     $\ell_{3/4}$ (75\% Pruned) & 1.79M/$3.43\times 10^8$ & 91.06/57.00 & 16.18 &93.79 \\
     \hline \hline
     $\ell_{1/2}$ (0\% Pruned)  & 20.04M/$7.97\times 10^8$ & 0.00/0.00 & 93.57& N/A \\
    $\ell_{1/2}$ (70\% Pruned) & 2.00M/$3.50\times 10^8$ & 90.01/56.12 & 40.07&93.77\\
     $\ell_{1/2}$ (75\% Pruned) & 1.66M/$3.25\times 10^8$ & 91.70/59.20 &13.65 &93.82 \\
     \hline \hline
      $\ell_{1/4}$ (0\% Pruned) & 20.04M/$7.97\times 10^8$ & 0.00/0.00 & 86.97& N/A \\
         $\ell_{1/4}$ (70\% Pruned) & 1.58M/$1.44\times 10^8$ & 92.14/81.89  & 47.59&92.15\\
     $\ell_{1/4}$ (90\% Pruned) & 0.19M/$0.13 \times 10^8$ & 99.05/98.32 & 10.00 &81.57 \\
          \hline \hline
     T$\ell_{1} (a=10.0)$ (0\% Pruned)  & 20.04M/$7.97\times 10^8$ & 0.00/0.00 & 93.64& N/A\\
    T$\ell_{1} (a=10.0)$ (70\% Pruned) & 2.19M/$3.77\times 10^8$ & 89.06/52.75 & 47.70 & 93.86\\
     T$\ell_{1} (a=10.0)$ (75\% Pruned) & 1.84M/$3.49\times 10^8$ & 90.82/56.19 &10.00 &93.72 \\    
     \hline \hline
     T$\ell_{1} (a=1.0)$ (0\% Pruned)  & 20.04M/$7.97\times 10^8$ & 0.00/0.00 & 93.55& N/A\\
    T$\ell_{1} (a=1.0)$ (70\% Pruned) & 1.93M/$3.39\times 10^8$ & 90.35/57.43 & 93.54 &93.86\\
     T$\ell_{1} (a=1.0)$ (75\% Pruned) & 1.66M/$3.24\times 10^8$ & 91.71/59.29 &86.83 &93.82 \\      \hline \hline
     T$\ell_{1} (a=0.5)$ (0\% Pruned)  & 20.04M/$7.97\times 10^8$ & 0.00/0.00 & 93.15& N/A\\
    T$\ell_{1} (a=0.5)$ (70\% Pruned) & 1.83M/$3.20\times 10^8$ & 90.88/59.84 & 93.14&93.75\\
     T$\ell_{1} (a=0.5)$ (75\% Pruned) & 1.53M/$3.05\times 10^8$ & 92.38/61.74 &92.38 &93.77 \\
          \hline \hline
                         MCP $(a=15000)$ (0\% Pruned)  & 20.04M/$7.97\times 10^8$ & 0.00/0.00 & 93.65& N/A\\
    MCP $(a=15000)$ (70\% Pruned) & 2.29M/$3.93\times 10^8$ & 88.58/50.69 & 47.18&93.97\\
     MCP $(a=15000)$ (75\% Pruned) & 1.89M/$3.58\times 10^8$ & 90.58/55.04 &10.00 &93.68 \\
          \hline \hline
     MCP $(a=10000)$ (0\% Pruned)  & 20.04M/$7.97\times 10^8$ & 0.00/0.00 & 93.69& N/A\\
    MCP $(a=10000)$ (70\% Pruned) & 2.28M/$3.95\times 10^8$ & 88.63/50.49 & 40.24&94.12\\
     MCP $(a=10000)$ (75\% Pruned) & 1.89M/$3.62\times 10^8$ & 90.56/54.54 &10.00 &93.73 \\
          \hline \hline

               SCAD $(a=15000)$ (0\% Pruned)  & 20.04M/$7.97\times 10^8$ & 0.00/0.00 & 93.64& N/A\\
    SCAD $(a=15000)$ (70\% Pruned) & 2.26M/$3.93\times 10^8$ & 88.71/50.70 & 52.72&93.94\\
    SCAD $(a=15000)$ (75\% Pruned) & 1.87M/$3.59\times 10^8$ & 90.65/54.97 & 10.00&93.91\\ \hline \hline
     SCAD $(a=10000)$ (0\% Pruned)  & 20.04M/$7.97\times 10^8$ & 0.00/0.00 & 93.60& N/A\\
    SCAD $(a=10000)$ (70\% Pruned) & 2.29M/$3.95\times 10^8$ & 88.57/50.43 & 55.25&93.88\\

    \end{tabular}
           \caption{CIFAR 10}
           \label{tab:vgg_retrain_cifar10}
\end{subtable} \\
\centering
\begin{subtable}[h]{\textwidth}
             \fontsize{6}{6}\selectfont
             \centering
    \begin{tabular}{l|c|c|c|c}
     & Number of Parameters/FLOPs & Percentage of Parameters/FLOPs Pruned (\%) & \makecell{Mean Test Accuracy\\
     before Retraining (\%)}& \makecell{Mean Test Accuracy\\
     after Retraining (\%)}\\
     \hline
     Baseline & 20.08M/$7.97\times 10^8$ & 0.00/0.00 & 72.73 & N/A\\ \hline \hline 
     $\ell_1$ (0\% Pruned) &20.08M/$7.97\times 10^8$ & 0.00/0.00 & 72.57 & N/A\\
          $\ell_1$ (45\% Pruned) & 5.67M/$5.26\times 10^8$& 71.78/34.00 & 51.16 &73.44\\
     $\ell_1$ (55\% Pruned) & 4.31M/$4.89\times 10^8$& 78.53/38.66 & 1.00 &72.98 \\
     \hline \hline
     $\ell_{3/4}$ (0\% Pruned) &20.08M/$7.97\times 10^8$ & 0.00/0.00 & 72.14 & N/A\\     $\ell_{3/4}$ (45\% Pruned) & 5.49M/$5.04\times 10^8$ & 72.68/36.75 &71.76 &73.24\\
     $\ell_{3/4}$ (55\% Pruned) & 4.10M/$4.76\times 10^8$ & 79.59/40.28 &3.40 &73.26\\
     \hline \hline
      $\ell_{1/2}$ (0\% Pruned) &20.08M/$7.97\times 10^8$ & 0.00/0.00 & 72.06& N/A\\
    $\ell_{1/2}$ (45\% Pruned) & 5.38M/$5.03\times 10^8$ & 73.21/36.95 & 71.27&73.34\\
     $\ell_{1/2}$ (60\% Pruned) & 3.40M/$4.48\times 10^8$ & 83.07/43.82 & 1.08&71.59 \\
     \hline \hline
      $\ell_{1/4}$ (0\% Pruned) &20.08M/$7.97\times 10^8$ &0.00/0.00 & 70.95& N/A\\
         $\ell_{1/4}$ (45\% Pruned) & 5.30M/$4.76\times 10^8$ & 73.59/40.26 & 22.70& 72.50\\
     $\ell_{1/4}$ (80\% Pruned) & 0.69M/$1.05\times 10^8$ & 96.54/86.86 &1.00 &46.97 \\
     \hline \hline
     T$\ell_{1} (a=10.0)$ (0\% Pruned)&20.08M/$7.97\times 10^8$ & 0.00/0.00 & 72.36& N/A\\
         T$\ell_{1} (a=10.0)$ (45\% Pruned) & 5.53M/$5.18\times 10^8$ & 72.45/34.95& 69.35& 73.39\\
    T$\ell_{1} (a=10.0)$ (55\% Pruned) & 4.21M/$4.85\times 10^8$ & 79.05/39.19& 1.46& 73.17\\
     \hline \hline
     T$\ell_{1} (a=1.0)$ (0\% Pruned)&20.08M/$7.97\times 10^8$ & 0.00/0.00 & 72.07& N/A\\
    T$\ell_{1} (a=1.0)$ (45\% Pruned) & 5.39M/$4.87\times 10^8$ & 73.16/38.89& 72.07& 73.03\\
     T$\ell_{1} (a=1.0)$ (60\% Pruned) & 3.43M/$4.47\times 10^8$ & 82.90/43.94 & 1.84& 73.06 \\      \hline \hline
     T$\ell_{1} (a=0.5)$ (0\% Pruned)&20.08M/$7.97\times 10^8$ & 0.00/0.00 & 71.63& N/A\\
    T$\ell_{1} (a=0.5)$ (45\% Pruned) & 5.29M/$4.74\times 10^8$ & 73.66/40.48 &71.63 &72.69\\
     T$\ell_{1} (a=0.5)$ (60\% Pruned) & 3.19M/$4.21\times 10^8$ & 84.09/47.15 &66.50 & 72.81 \\
               \hline \hline
                         MCP $(a=15000)$ (0\% Pruned)  & 20.08M/$7.97\times 10^8$ & 0.00/0.00 & 72.26& N/A\\
                              MCP $(a=15000)$ (45\% Pruned) & 5.66M/$5.27\times 10^8$ & 71.82/33.87 &66.14 &73.68\\
     MCP $(a=15000)$ (55\% Pruned) & 4.30M/$4.92\times 10^8$ & 78.58/38.21 &1.00 &72.94 \\
          \hline \hline

               SCAD $(a=15000)$ (0\% Pruned)  & 20.08M/$7.97\times 10^8$ & 0.00/0.00 & 72.50& N/A\\
                   SCAD $(a=15000)$ (45\% Pruned) & 5.64M/$5.26\times 10^8$ & 71.89/33.99 & 65.72&73.61\\ 
    SCAD $(a=15000)$ (55\% Pruned) & 4.32M/$4.90\times 10^8$ & 78.48/38.49 & 1.00&72.67\\ \hline \hline
     SCAD $(a=10000)$ (0\% Pruned)  & 20.08M/$7.97\times 10^8$ & 0.00/0.00 & 72.33& N/A\\
         SCAD $(a=10000)$ (45\% Pruned) & 5.72M/$5.32\times 10^8$ & 71.50/33.21 & 64.98&73.52\\
    SCAD $(a=10000)$ (55\% Pruned) & 4.37M/$4.94\times 10^8$ & 78.22/37.99 & 1.00&71.98\\
    \end{tabular}
            \caption{CIFAR 100}
            \label{tab:vgg_retrain_cifar100}
    \end{subtable}
\end{table*}
\begin{table*}[h!!!]   
    \caption{Results from five retrained DenseNet-40 on CIFAR 10/100 after pruning. Baseline refers to the DenseNet-40 model trained without regularization on the scaling factors.}
    \label{tab:dense_retrain}
    \fontsize{6}{6}\selectfont
\centering
\begin{subtable}[h]{\textwidth}
\fontsize{5.5}{5.5}\selectfont
       \centering
    \begin{tabular}{l|c|c|c|c}
     &  Number of Parameters/FLOPs & Percentage of Parameters/FLOPs Pruned (\%) & \makecell{Mean Test Accuracy\\
     before Retraining (\%)}&\makecell{Mean Test Accuracy\\
     after Retraining (\%)}\\
     \hline
     Baseline & 1.02M/$5.33\times 10^8$ & 0.00/0.00 & 94.25& N/A\\ \hline \hline 
     $\ell_1$ (0 \% Pruned) & 1.02M/$5.33\times 10^8$ &0.00/0.00 & 93.46& N/A\\
          $\ell_1$ (82.5\% Pruned) & 0.24M/$1.54\times 10^8$ & 76.21/71.20  & 78.27 & 93.46\\
     $\ell_1$ (90\% Pruned) & 0.17M/$1.08\times 10^8$ & 83.76/79.75  & 17.47 & 91.42\\
     \hline \hline
     $\ell_{3/4}$ (0\% Pruned)& 1.02M/$5.33\times 10^8$ & 0.00/0.00 & 93.19& N/A\\
          $\ell_{3/4}$ (82.5\% Pruned) & 0.24M/$1.53\times 10^8$  & 76.57/71.34& 90.17&93.33\\
     $\ell_{3/4}$ (90\% Pruned) & 0.16M/$1.06\times 10^8$  & 84.02/80.07& 15.06&91.54\\
     \hline \hline
       $\ell_{1/2}$ (0\% Pruned)&1.02M/$5.33\times 10^8$ & 0.00/0.00 & 93.28& N/A\\
         $\ell_{1/2}$ (82.5\% Pruned) & 0.25M/$1.51\times 10^8$ &76.84/71.76  & 83.17&93.43\\
    $\ell_{1/2}$ (90\% Pruned) & 0.16M/$1.06\times 10^8$ &84.36/80.13  & 13.76&91.31\\
     \hline \hline 
          $\ell_{1/4}$ (0\% Pruned)& 1.02M/$5.33\times 10^8$ & 0.00/0.00& 89.48& N/A\\
     $\ell_{1/4}$ (82.5\% Pruned) & 0.21M/$1.14\times 10^8$ &79.81/78.63  & 11.29& 91.68\\
         $\ell_{1/4}$ (85\% Pruned) & 0.18M/$0.98\times 10^8$ &82.57/81.64  & 10.05& 91.44\\
     \hline \hline
           T$\ell_{1} (a=10.0)$ (0\% Pruned)& 1.02M/$5.33\times 10^8$ & 0.00/0.00 & 93.30& N/A\\
         T$\ell_{1} (a=10.0)$ (82.5\% Pruned) &0.24M/$1.54\times 10^8$ &76.33/71.10 &83.24 & 93.38\\ 
    T$\ell_{1} (a=10.0)$ (90\% Pruned) &0.16M/$1.08\times 10^8$  &83.89/79.72  & 15.35& 91.37\\     \hline \hline
      T$\ell_{1} (a=1.0)$ (0\% Pruned)& 1.02M/$5.33\times 10^8$ & 0.00/0.00 & 93.16& N/A\\
         T$\ell_{1} (a=1.0)$ (82.5\% Pruned) &0.24M/$1.53\times 10^8$ &76.80/71.35 &93.17  & 93.26\\ 
    T$\ell_{1} (a=1.0)$ (90\% Pruned) &0.16M/$1.06\times 10^8$  &84.23/80.19  & 18.91& 91.70\\     \hline \hline
      T$\ell_{1} (a=0.5)$ (0\% Pruned)& 1.02M/$5.33\times 10^8$ & 0.00/0.00 & 92.78& N/A\\
        T$\ell_{1} (a=0.5)$ (82.5\% Pruned) &0.23M/$1.50\times 10^8$  & 77.21/71.83 & 92.74&93.05\\
    T$\ell_{1} (a=0.5)$ (90\% Pruned) &0.16M/$1.03\times 10^8$  & 84.45/80.70 & 18.12&91.69\\\hline \hline
          MCP$(a=15000)$ (0\% Pruned)& 1.02M/$5.33\times 10^8$ & 0.00/0.00 & 93.48& N/A\\
        MCP$(a=15000)$(82.5\% Pruned) &0.24M/$1.55\times 10^8$  & 76.23/71.00 & 92.74&93.44\\
    MCP$(a=15000)$ (90\% Pruned) &0.17M/$1.10\times 10^8$  & 83.72/79.37 & 12.92&91.31\\\hline \hline
          MCP$(a=10000)$ (0\% Pruned)& 1.02M/$5.33\times 10^8$ &0.00/0.00 & 93.41& N/A\\
        MCP$(a=10000)$ (82.5\% Pruned) &0.24M/$1.53\times 10^8$  & 76.37/71.23 & 67.36&93.53\\
    MCP$(a=10000)$ (90\% Pruned) &0.16M/$1.10\times 10^8$  & 83.85/79.39& 15.08&91.24\\\hline \hline
              SCAD$(a=15000)$ (0\% Pruned)& 1.02M/$5.33\times 10^8$ & 0.00/0.00 & 93.48& N/A\\
        SCAD$(a=15000)$ (82.5\% Pruned) &0.24M/$1.54\times 10^8$  & 76.28/71.02 & 71.33&93.42\\
    SCAD$(a=15000)$ (90\% Pruned) &0.17M/$1.10\times 10^8$  & 83.80/79.42 & 14.21&91.26\\\hline \hline
    SCAD$(a=10000)$ (0\% Pruned)& 1.02M/$5.33\times 10^8$ & 0.00/0.00 & 93.52& N/A\\
        SCAD$(a=10000)$ (82.5\% Pruned) &0.24M/$1.55\times 10^8$  & 76.25/70.93 & 71.49&93.49\\
    SCAD$(a=10000)$ (90\% Pruned) &0.17M/$1.10\times 10^8$  & 83.75/79.27 & 12.27&91.18\\
    \end{tabular}
           \caption{CIFAR 10}
           \label{tab:dense_cifar10}
\end{subtable} \\
\centering
\begin{subtable}[h]{\textwidth}
             \fontsize{5.5}{5.5}\selectfont
             \centering
    \begin{tabular}{l|c|c|c|c}
     & Number of Parameters/FLOPs & Percentage of Parameters/FLOPs Pruned (\%) &  \makecell{Mean Test Accuracy\\
     before Retraining (\%)}& \makecell{Mean Test Accuracy\\
     after Retraining (\%)}\\
     \hline
     Baseline & 1.06M/$5.33\times 10^8$ & 0.00/0.00 & 74.58 & N/A\\ \hline \hline 
               $\ell_1$ (0\% Pruned) & 1.06M/$5.33\times 10^8$ & 0.00/0.00 & 73.24 & N/A\\
          $\ell_1$ (75\% Pruned) &0.35M/$2.14\times 10^8$& 68.73/59.89 & 54.68 & 73.73 \\
     $\ell_1$ (85\% Pruned) &0.23M/$1.46\times 10^8$& 78.08/72.60 & 2.94 & 72.40\\
     \hline \hline
      $\ell_{3/4}$ (0\% Pruned)& 1.06M/$5.33\times 10^8$ & 0.00/0.00 & 72.97 & N/A\\
               $\ell_{3/4}$ (75\% Pruned) & 0.33M/$2.11\times 10^8$& 68.93/60.40 & 68.60 & 73.75\\
     $\ell_{3/4}$ (90\% Pruned) & 0.18M/$1.07\times 10^8$ &83.34/79.89 & 1.23 & 69.33 \\
     \hline \hline
     $\ell_{1/2}$ (0\% Pruned) & 1.06M/$5.33\times 10^8$ & 0.00/0.00 & 72.98& N/A\\
         $\ell_{1/2}$ (75\% Pruned) &0.33M/$2.06\times 10^8$  & 69.03/61.41&68.05 & 73.39  \\
    $\ell_{1/2}$ (85\% Pruned) &0.23M/$1.42\times 10^8$  & 78.42/73.43 & 5.05 & 72.52\\
     \hline \hline
      $\ell_{1/4}$ (0\% Pruned) & 1.06M/$5.33\times 10^8$ & 0.00/0.00 & 69.02& N/A\\
              $\ell_{1/4}$ (75\% Pruned) &0.31M/$1.62\times 10^8$  & 70.81/69.59  & 1.45 & 71.62\\

         $\ell_{1/4}$ (85\% Pruned) &0.19M/$0.88\times 10^8$  & 82.28/83.54  & 1.00& 67.76\\
     \hline \hline
            T$\ell_{1} (a=10.0)$ (0\% Pruned)&1.06M/$5.33\times 10^8$ & 0.00/0.00 & 73.18& N/A\\
          T$\ell_{1} (a=10.0)$ (75\% Pruned) & 0.33M/$2.12\times 10^8$ & 68.84/60.18 & 66.62 & 73.78\\ 
    T$\ell_{1} (a=10.0)$ (85\% Pruned) & 0.23M/$1.47\times 10^8$ & 78.21/72.37 & 3.17 & 72.69 \\      \hline \hline
       T$\ell_{1} (a=1.0)$ (0\% Pruned)& 1.06M/$5.33\times 10^8$ & 0.00/0.00 & 72.63& N/A\\
          T$\ell_{1} (a=1.0)$ (75\% Pruned) & 0.33M/$2.12\times 10^8$ & 69.16/60.24 & 72.60 & 73.42\\ 
    T$\ell_{1} (a=1.0)$ (90\% Pruned) & 0.18M/$1.07\times 10^8$ & 83.48/80.01 & 1.24 & 69.98 \\      \hline \hline
    T$\ell_{1} (a=0.5)$ (0\% Pruned) & 1.06M/$5.33\times 10^8$ & 0.00/0.00 & 72.57& N/A\\   

        T$\ell_{1} (a=0.5)$ (75\% Pruned) & 0.33M/$2.10\times 10^8$ & 69.33 /60.56 & 72.59 & 73.23\\  

    T$\ell_{1} (a=0.5)$ (90\% Pruned) & 0.17M/$1.06\times 10^8$ & 83.61/80.16 & 1.37&70.16\\ \hline \hline
            MCP$(a=15000)$ (0\% Pruned)& 1.06M/$5.33\times 10^8$ & 0.00/0.00 & 73.64& N/A\\
        MCP$(a=15000)$(75\% Pruned) &0.33M/$2.10\times 10^8$  & 68.80/60.61 & 58.12&73.73\\
    MCP$(a=15000)$ (90\% Pruned) &0.18M/$1.08\times 10^8$  & 83.35/79.73 & 1.27&69.94\\\hline \hline
          MCP$(a=10000)$ (0\% Pruned)& 1.06M/$5.33\times 10^8$ &0.00/0.00 & 73.40& N/A\\
        MCP$(a=10000)$ (75\% Pruned) &0.33M/$2.06\times 10^8$  & 68.73/61.36 & 40.76&73.95\\
    MCP$(a=10000)$ (90\% Pruned) &0.18M/$1.08\times 10^8$  & 83.19/79.68& 1.10&69.10\\\hline \hline
              SCAD$(a=15000)$ (0\% Pruned)& 1.06M/$5.33\times 10^8$ & 0.00/0.00 & 73.41& N/A\\
        SCAD$(a=15000)$ (75\% Pruned) &0.33M/$2.09\times 10^8$  & 68.79/60.83 & 54.71&73.97\\
    SCAD$(a=15000)$ (90\% Pruned) &0.18M/$1.08\times 10^8$  & 83.33/79.72 & 1.42&69.87\\\hline \hline
    SCAD$(a=10000)$ (0\% Pruned)& 1.06M/$5.33\times 10^8$ & 0.00/0.00 & 73.37& N/A\\
        SCAD$(a=10000)$ (75\% Pruned) &0.33M/$2.04\times 10^8$  & 68.80/61.66 & 47.70&73.75\\
    SCAD$(a=10000)$ (90\% Pruned) &0.18M/$1.09\times 10^8$  & 83.36/79.61 & 1.08&69.73\\
    \end{tabular}
            \caption{CIFAR 100}
            \label{tab:dense_cifar100}
    \end{subtable}

\end{table*}

Because the test accuracy drops after channel pruning for VGG-19 and DenseNet-40 trained on CIFAR 10/100, we retrain the models without regularization on the scaling factors and examine whether or not the original test accuracy is recovered. For  brevity, we analyze $\ell_1$ and the nonconvex regularizers whose possible channel pruning percentages are at least the same as $\ell_1$'s. 

\textbf{VGG-19.} The results for VGG-19 on CIFAR 10/100 are presented in Table \ref{tab:vgg_retrain}. Generally, we observe that the test accuracy after retraining is better than the original test accuracy before channel pruning and retraining. For CIFAR 10, after the models are retrained with 70\% of their channels pruned, only $\ell_1$ $\ell_{3/4}$, T$\ell_1(a=1.0, 10.0)$, MCP ($a=10000, 15000)$, and SCAD $(a=10000,15000)$ exceed the baseline test accuracy of 93.83\%. Among the nonconvex regularizers, $\ell_{3/4}$ and T$\ell_1 (a=1.0, 10.0)$ yield more compressed models in terms of both parameters and FLOPS but have slightly lower test accuracies than $\ell_1$. On the other hand, MCP $(a=15000, 10000)$ and SCAD $(a=15000)$ are slightly less compressed than $\ell_1$ but have better test accuracies. When 75\% of the channels are pruned, their retrained test accuracies decrease slightly due to compressing the models further. Among the nonconvex regularizers, the test accuracy for SCAD $(a=15000)$ is better than the baseline. Moreover, SCAD $(a=15000)$ with 75\% of its channels pruned requires less parameters and FLOPS than $\ell_1$ with 70\% of its channels pruned. For $\ell_{1/4}$, when 90\% of the channels are pruned, at least 98\% of parameters and FLOPs are pruned, but the test accuracy after retraining is 81.57\%. For CIFAR 100, with 45\% of the channels pruned, all of the regularizers except for $\ell_{1/4}$ and T$\ell_{1} (a=0.5)$  attain better test accuracies than the baseline accuracy of $72.73\%$. Similar to CIFAR 10, $\ell_{3/4}, \ell_{1/2}$, and T$\ell_{1} (a=1.0, 10.0)$ have slightly lower test accuracies than $\ell_1$ but have better compression. MCP and SCAD have better test accuracies than $\ell_1$ with similar parameter and FLOP compression. When more channels are pruned, most of the regularizers suffer a slight decrease in retrained test accuracies. Only $\ell_{3/4}$ with 55\% channels pruned and T$\ell_1(a=0.5, 1.0)$ with 60\% channels pruned experience a modest improvement in test accuracy, but their test accuracies exceed the baseline test accuracy and $\ell_1$'s test accuracy with 55\% channels pruned. 

\begin{table*}
\caption{Counts of scaling factors that are averaged across five runs per model and regularizer.}
\label{tab:scaling_factor}
\begin{subtable}[h]{\textwidth}
    \centering
    \begin{tabular}{l|c|c||c|c||c|c|}
\cline{2-7} 
         & \multicolumn{2}{c||}{VGG-19} & \multicolumn{2}{c||}{DenseNet-40} &  \multicolumn{2}{c|}{ResNet-164}   \\ \cline{2-7} 
         & $|\gamma|\leq 10^{-6}$ & $|\gamma|>10^{-6}$ & $|\gamma|\leq10^{-6}$ & $|\gamma|>10^{-6}$ & $|\gamma|\leq 10^{-6}$ & $|\gamma|>10^{-6}$ \\ \hline
          $\ell_1$ & 3483 & 2021 & 7309.2 & 2050.8 & 7321.4 & 4790.6 \\ \hline
        $\ell_{3/4}$ & 3244.8 & 2259.2 & 6261.4 & 3098.6 & 7944.2 & 4167.8 \\ \hline
        $\ell_{1/2}$ & 263.4 & 5240.6 & 490.6 & 8869.4 & 898 & 11214\\ \hline
        $\ell_{1/4}$ & 3 & 5501 & 4 & 9356 & 11.6 & 12100.4 \\ \hline
         T$\ell_1 (a=10.0)$ & 3559.6 & 1944.4 & 7372.8 & 1987.2 & 7466.6 & 4645.4 \\ \hline
         T$\ell_1 (a=1.0)$ & 4021.2 & 1482.8 & 7731.4 & 1628.6 & 8757.2 & 3354.8 \\ \hline
        T$\ell_1 (a=0.5)$ & 4216 & 1288 & 7839 & 1521 & 9192 & 2920 \\ \hline
        MCP $(a=15000)$ & 3472.4 & 2031.6 & 7180.6 & 2179.4 & 6805.8 & 5306.2  \\ \hline
        MCP $(a=10000)$ & 3485 & 2019 & 7123.6 & 2236.4 & 6438.4 & 5673.6 \\ \hline
        MCP $(a=5000)$ & 3440.4 & 2063.6 &  6880.2 & 2479.8 & 5542.6 & 6569.4 \\ \hline
        SCAD $(a=15000)$ & 3492.6 & 2011.4 & 7204.4 & 2155.6 & 6818.4 & 5293.6 \\ \hline
        SCAD $(a=10000)$ & 3460.2 & 2043.8 & 7121.4 & 2238.6 & 6484.6 & 5627.4  \\ \hline
        SCAD $(a=5000)$ & 3518.2 & 1985.8 & 6947.8 & 2412.2 & 5514.6 & 6597.4\\ \hline
    \end{tabular}                \caption{CIFAR 10}
            \label{tab:cifar10_scale}
\end{subtable}
\begin{subtable}[h]{\textwidth}
    \centering
    \begin{tabular}{l|c|c||c|c||c|c|}
\cline{2-7} 
         & \multicolumn{2}{c||}{VGG-19} & \multicolumn{2}{c||}{DenseNet-40} &  \multicolumn{2}{c|}{ResNet-164}   \\ \cline{2-7} 
         & $|\gamma|\leq 10^{-6}$ & $|\gamma|>10^{-6}$ & $|\gamma|\leq10^{-6}$ & $|\gamma|>10^{-6}$ & $|\gamma|\leq 10^{-6}$ & $|\gamma|>10^{-6}$ \\ \hline 
        $\ell_1$ & 1417.2 & 4086.8 & 6382 & 2978 & 5030.4 & 7081.6 \\ \hline
        $\ell_{3/4}$ & 1895.8 & 3608.2 & 2208.6 & 7151.4 & 5584.6 & 6527.4 \\ \hline
        $\ell_{1/2}$ & 151.6 & 5352.4 & 94.4 & 9265.6 & 430 & 11682 \\ \hline
        $\ell_{1/4}$ & 1.6 & 5502.4 & 6 & 9354 & 6.6 & 12105.4 \\ \hline
        T$\ell_1 (a=10.0)$ & 1629.6 & 3874.4 & 6555.4 & 2804.6 & 5192.8 & 6919.2 \\ \hline
        T$\ell_1 (a=1.0)$ & 2555.6 & 2948.4 & 6919.8 & 2440.2 & 6250 & 5862 \\ \hline
        T$\ell_1 (a=0.5)$ & 2802 & 2702 & 6889.6 & 2470.4 & 6739 & 5373 \\ \hline
        MCP $(a=15000)$ & 1495 & 4009 & 6192.2 & 3167.8 & 4521.8 & 7590.2 \\ \hline
        MCP $(a=10000)$ & 1440.4 & 4063.6 & 6055.6 & 3304.4 & 4191.8 & 7920.2 \\ \hline
        MCP $(a = 5000)$ & 1378 & 4126 & 5627.4 & 3732.6 & 3541.8 & 8570.2 \\ \hline
        SCAD $(a=15000)$ & 1514.4 & 3989.6 & 6190.4 & 3169.6 & 4481.6 & 7630.4 \\ \hline
        SCAD $(a=10000)$ & 1481.6 & 4022.4 & 6034.6 & 3325.4 & 4211.6 & 7900.4 \\ \hline
        SCAD $(a = 5000)$ & 1262 & 4242 & 5595.6 & 3764.4 & 3484.6 & 8627.4 \\ \hline
    \end{tabular}
                \caption{CIFAR 100}
            \label{tab:cifar100_scale}
    \end{subtable}
    \begin{subtable}[h]{\textwidth}
    \centering
    \begin{tabular}{l|c|c||c|c||c|c|}
    \cline{2-7} 
         & \multicolumn{2}{c||}{CIFAR 10} & \multicolumn{2}{c||}{CIFAR 100} &  \multicolumn{2}{c|}{SVHN}   \\ \cline{2-7} 
         & $|\gamma| \leq 10^{-6}$ & $|\gamma| >10^{-6}$ & $|\gamma|\leq10^{-6}$ & $|\gamma|>10^{-6}$ & $|\gamma|\leq 10^{-6}$ & $|\gamma|>10^{-6}$ \\ \hline 
        $\ell_1$ & 4447.6 & 1056.4 & 8447.4 & 912.6 & 10058.8 & 2053.2 \\ \hline
        $\ell_{3/4}$ & 3862.2 & 1641.8 & 7079 & 2281 & 10130.4 & 1981.6 \\ \hline
        $\ell_{1/2}$ & 292 & 5212 & 543.4 & 8816.6 & 1070.4 & 11041.6 \\ \hline
        $\ell_{1/4}$ & 3.4 & 5500.6 & 7.2 & 9352.8 & 12.6 & 12099.4 \\ \hline
        T$\ell_1 (a=10.0)$ & 4505.4 & 998.6 & 8497.4 & 862.6 & 10184.8 & 1927.2 \\ \hline
        T$\ell_1 (a=1.0)$ & 4796.8 & 707.2 & 8674 & 686 & 10813.2 & 1298.8 \\ \hline
        T$\ell_1 (a=0.5)$ & 4874 & 630 & 8746.4 & 613.6 & 11002.4 & 1109.6 \\ \hline
        MCP $(a=15000)$ & 4365.6 & 1138.4 & 8419.8 & 940.2 & 9930.4 & 2181.6 \\ \hline
        MCP $(a=10000)$ & 4356.6 & 1147.4 & 8390.4 & 969.6 & 9841 & 2271 \\ \hline
        MCP $(a = 5000)$ & 4242.6 & 1261.4 & 8330 & 1030 & 9333.6 & 2778.4 \\ \hline
        SCAD $(a=15000)$ & 4378.2 & 1125.8 & 8405.6 & 954.4 & 9894.8 & 2217.2 \\ \hline
        SCAD $(a=10000)$ & 4361.4 & 1142.6 & 8407 & 953 & 9858.8 & 2253.2 \\ \hline
        SCAD $(a = 5000)$ & 4244.4 & 1259.6 & 8330.2 & 1029.8 & 9353.8 & 2758.2 \\ \hline
    \end{tabular}
                \caption{SVHN}
            \label{tab:SVHN_scale}
    \end{subtable}
\end{table*}

Overall, for $\ell_p(p=1/2, 3/4)$ and T$\ell_1 (a=0.5,1.0)$, the retrained models, despite being more compressed than their $\ell_1$ counterparts, have slightly lower test accuracies. However, MCP and SCAD have similar compression as $\ell_1$ but with better test accuracies after retraining. 

\textbf{DenseNet-40.} Table \ref{tab:dense_retrain} reports the results for DenseNet-40 on CIFAR 10/100. Overall, the baseline accuracy is better than all of the retrained test accuracies, but the differences are at most 3.07\% for CIFAR 10 and at most 6.82\% for CIFAR 100. For CIFAR 10, when 82.5\% of the channels are pruned, only MCP $(a=10000)$ and SCAD $(a=10000)$ have better test accuracies than $\ell_1$ with similar compression in parameters and FLOPs. For $\ell_p(p=1/2,3/4)$ and T$\ell_1 (a=1.0)$, their retrained test accuracies are only slightly lower by at most 0.20\%, but this is at the cost of better compression. When 90\% of the channels are pruned, the retrained test accuracies decrease slightly more into the range of 91\%-92\%. Only $\ell_{3/4}$ and T$\ell_1 (a=0.5, 1.0)$ have better test accuracies than $\ell_1$ with much better compression. For CIFAR 100, when 75\% of the channels are pruned, $\ell_{3/4}$, T$\ell_1 (a=10.0)$, MCP, and SCAD have at least the same test accuracies as $\ell_1$ with better compression in parameters and FLOPs. However, increasing the channel pruning percentage to 90\% causes their retrained test accuracies to deteriorate. As a result, none of the models is able to exceed the test accuracy of the $\ell_1$-regularized models retrained with 85\% of their channels pruned. For $\ell_{1/2}$ and T$\ell_1(a=10.0)$, when 85\% of the channels are pruned, their test accuracies exceed $\ell_1$. 

In general, pruning channels for DenseNet at the highest percentage possible can be detrimental to the retrained test accuracy. When channels are pruned at intermediate levels, the nonconvex regularizers can have better retrained test accuracies and/or better compression than $\ell_1$.  
\subsection{Scaling Factor Analysis}
\begin{figure*}[h!!!]
\centering
     \begin{subfigure}[b]{0.40\textwidth}
         \centering
         \includegraphics[width=\textwidth]{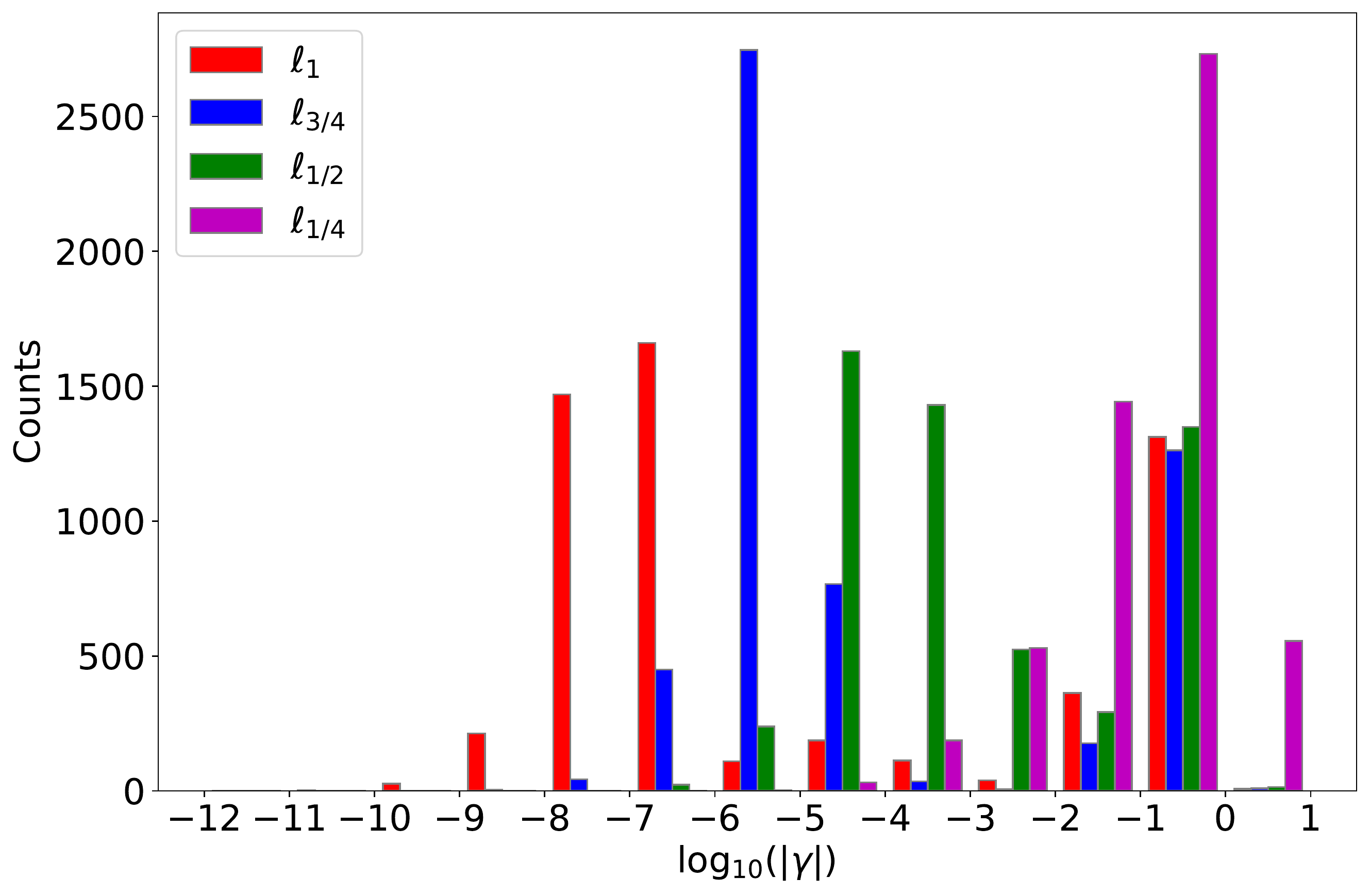}
         \caption{$\ell_p$}
         \label{fig:vgg_lp_scaling_cifar10}
     \end{subfigure}
     \begin{subfigure}[b]{0.40\textwidth}
         \centering
         \includegraphics[width=\textwidth]{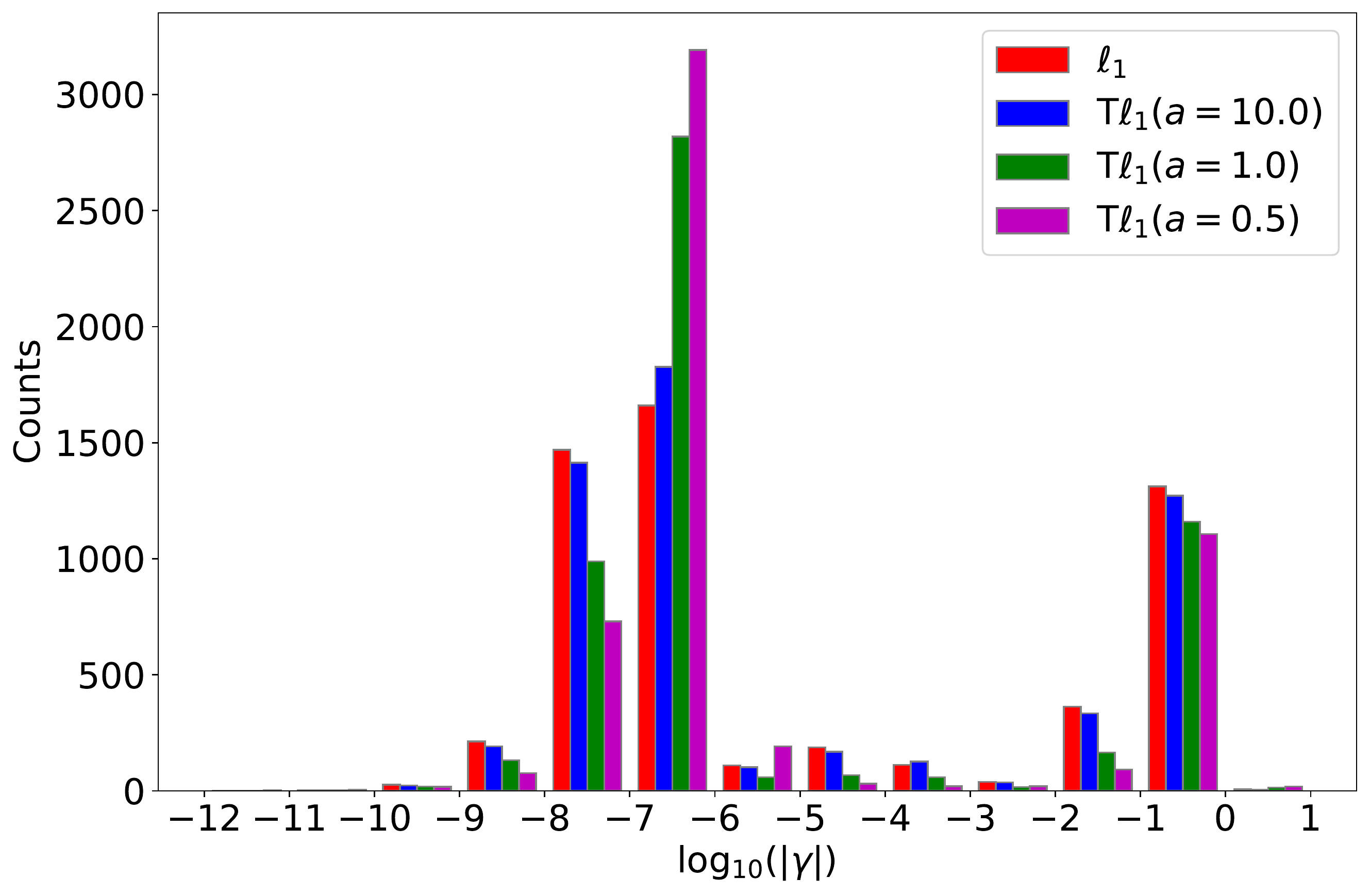}
         \caption{T$\ell_1$}
         \label{fig:vgg_Tl1_scaling_cifar10}
     \end{subfigure}\\
     \begin{subfigure}[b]{0.40\textwidth}
         \centering
         \includegraphics[width=\textwidth]{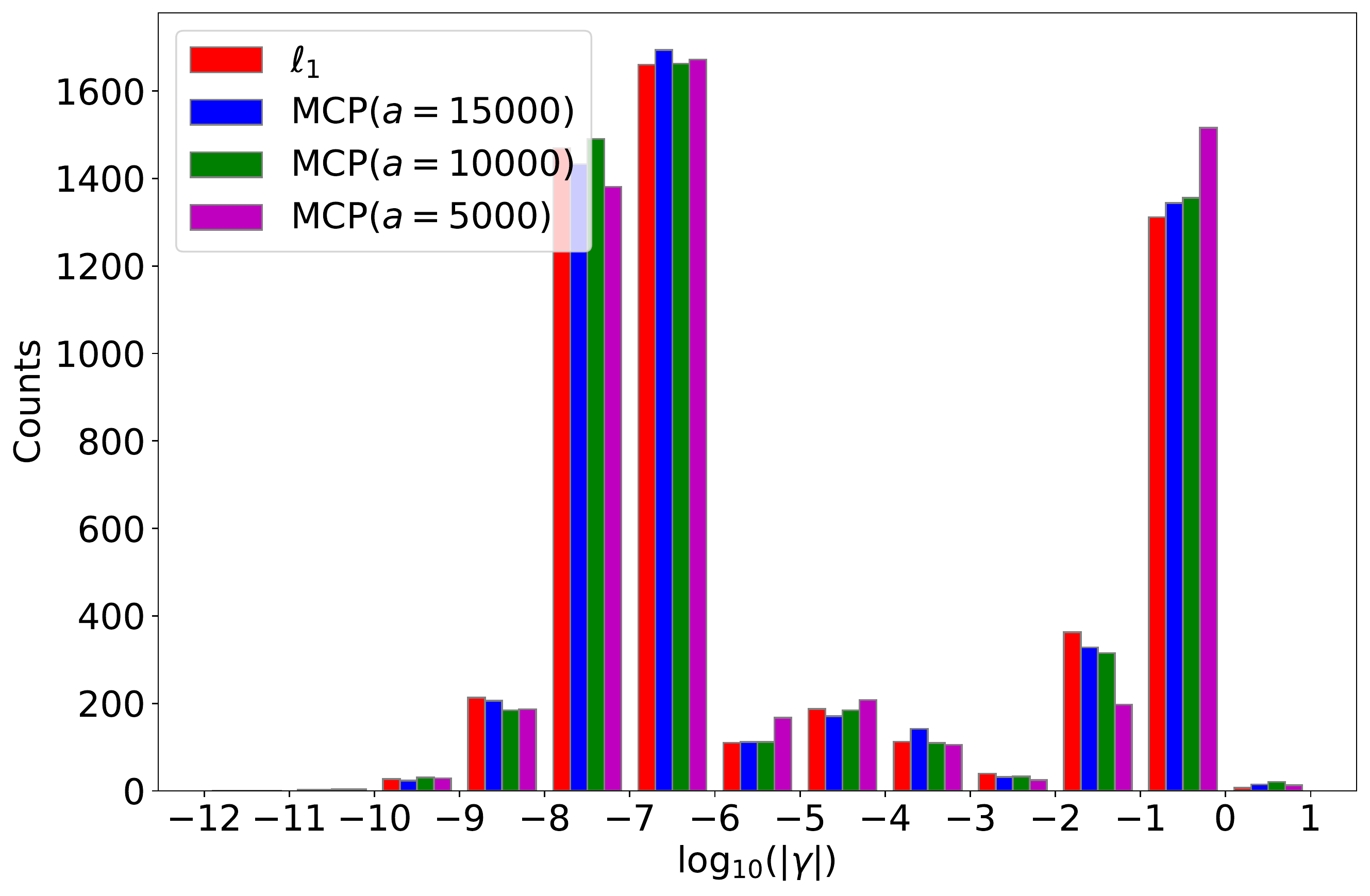}
         \caption{MCP}
         \label{fig:vgg_MCP_scaling_cifar10}
     \end{subfigure}
     \begin{subfigure}[b]{0.40\textwidth}
         \centering
         \includegraphics[width=\textwidth]{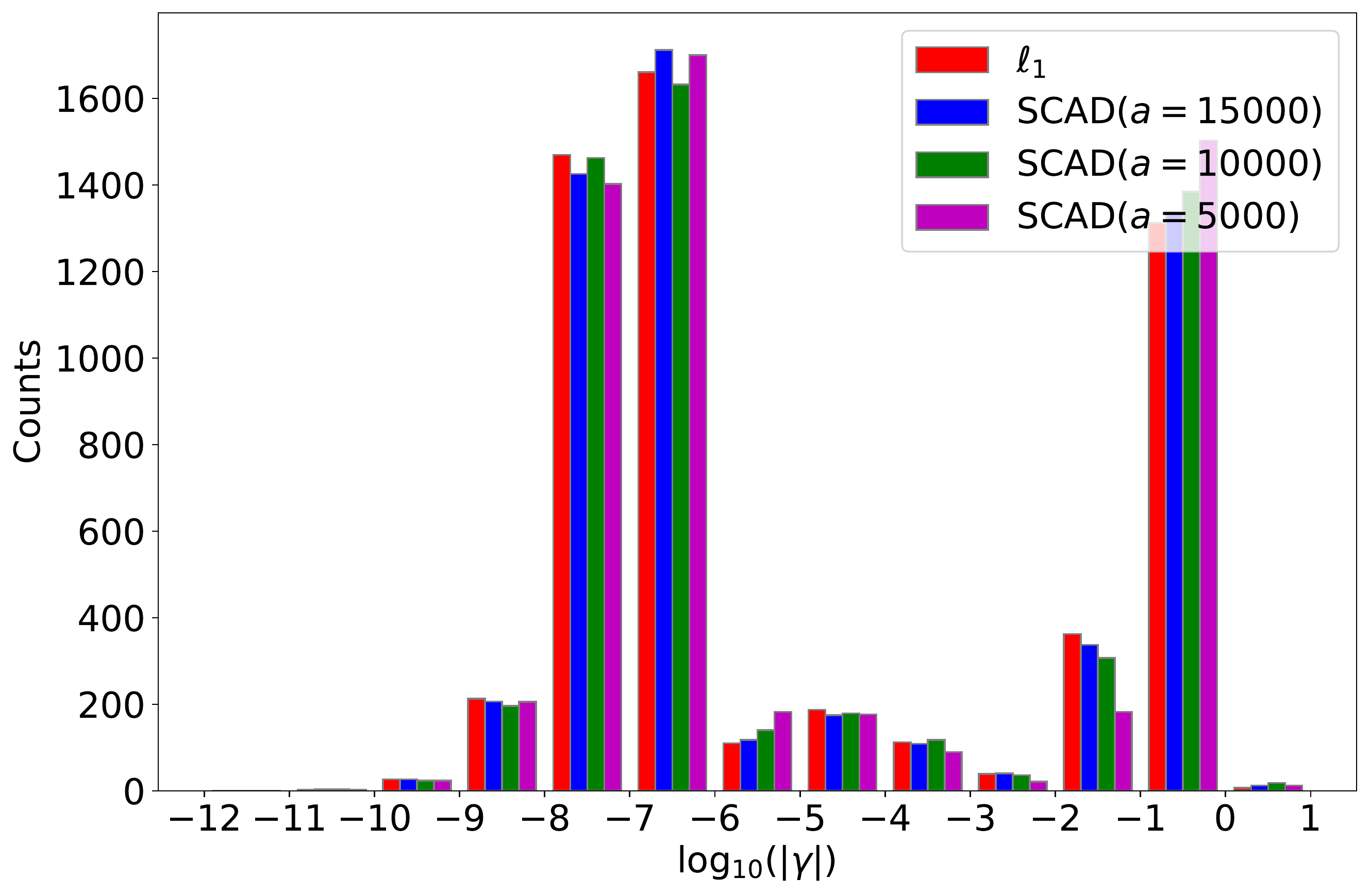}
         \caption{SCAD}
         \label{fig:vgg_SCAD_scaling_cifar10}
     \end{subfigure}
        \caption{Histogram of scaling factors $\gamma$ in VGG-19 trained on CIFAR 10. The $x$-axis is $\log_{10}(|\gamma|)$.}
        \label{fig:vgg_scaling_cifar10}
\end{figure*}
\begin{figure*}[h!!!]
\centering
     \begin{subfigure}[b]{0.40\textwidth}
         \centering
         \includegraphics[width=\textwidth]{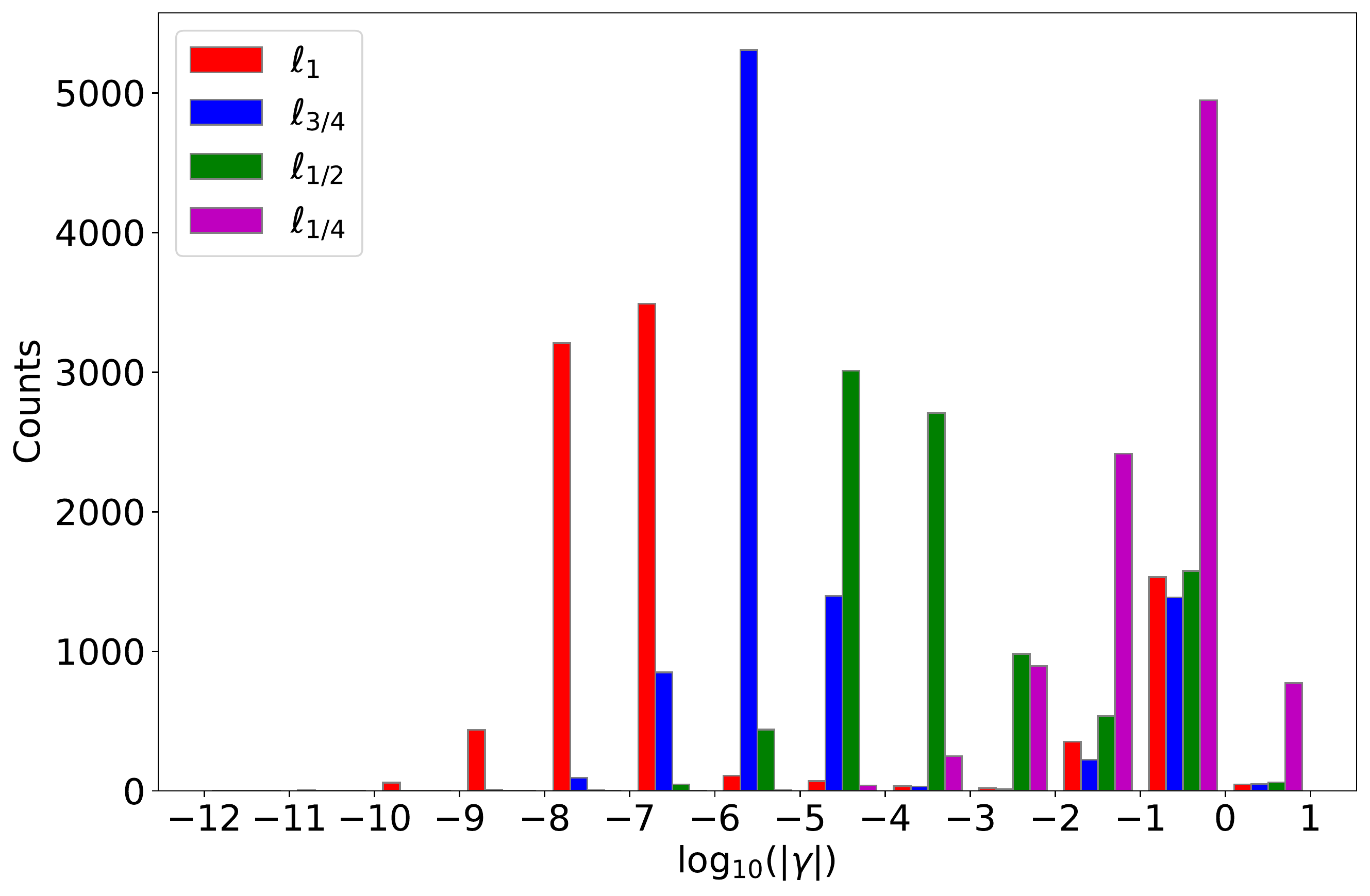}
         \caption{$\ell_p$}
         \label{fig:densenet_Lp_scaling_cifar10}
     \end{subfigure}
     \begin{subfigure}[b]{0.40\textwidth}
         \centering
         \includegraphics[width=\textwidth]{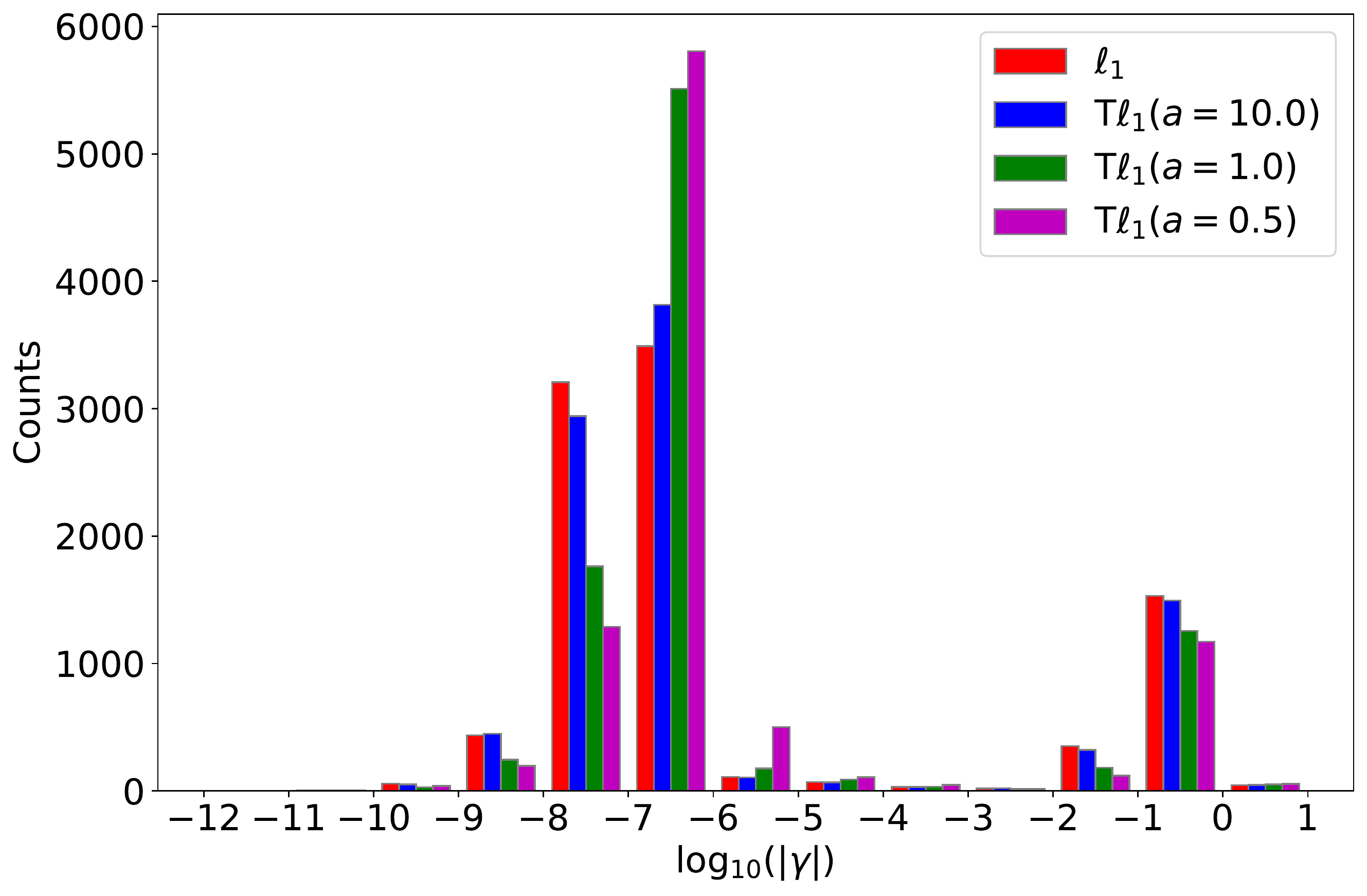}
         \caption{T$\ell_1$}
         \label{fig:densenet_TL1_scaling_cifar10}
     \end{subfigure}\\
     \begin{subfigure}[b]{0.40\textwidth}
         \centering
         \includegraphics[width=\textwidth]{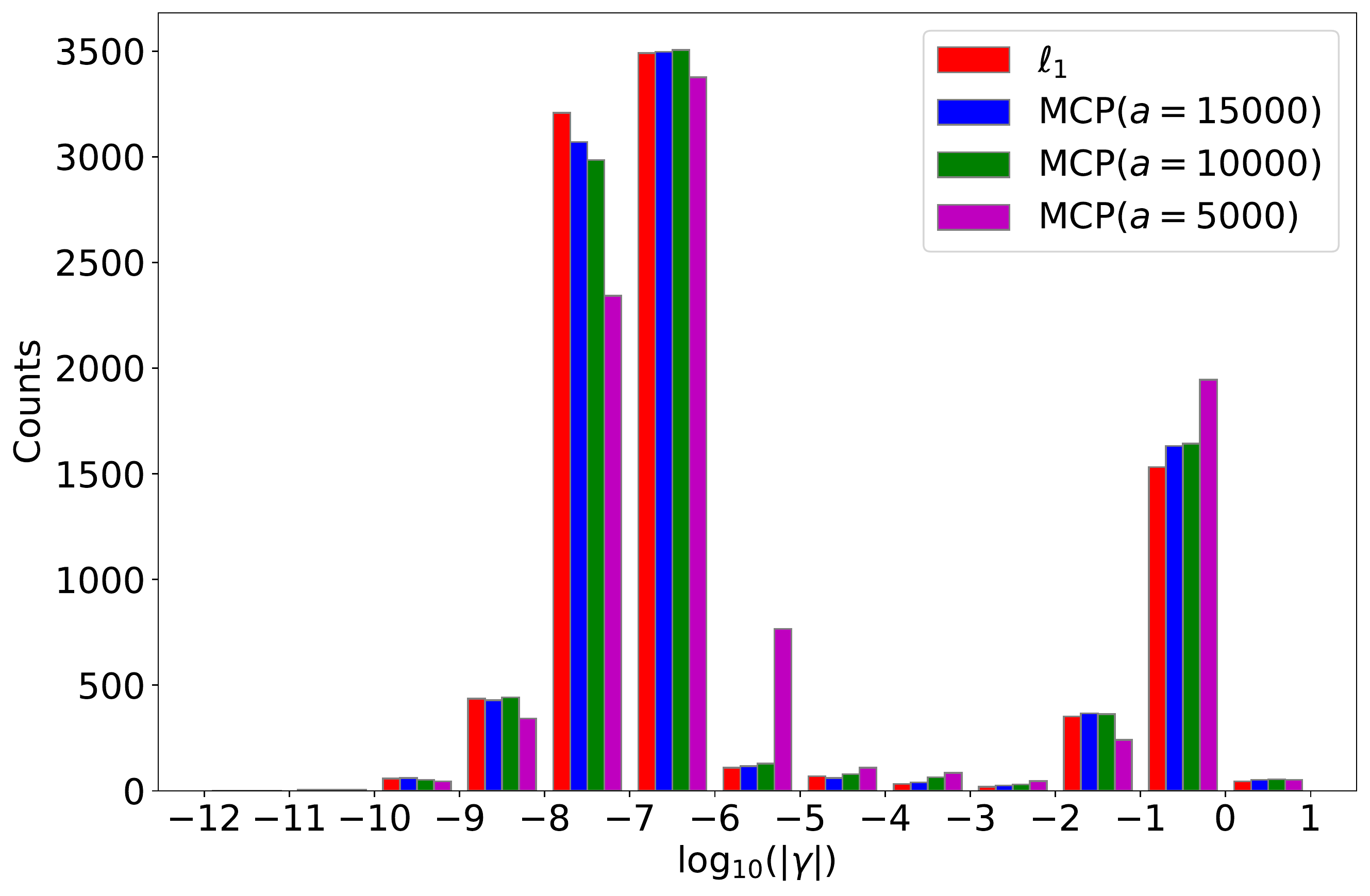}
         \caption{MCP}
         \label{fig:densenet_MCP_scaling_cifar10}
     \end{subfigure}
     \begin{subfigure}[b]{0.40\textwidth}
         \centering
         \includegraphics[width=\textwidth]{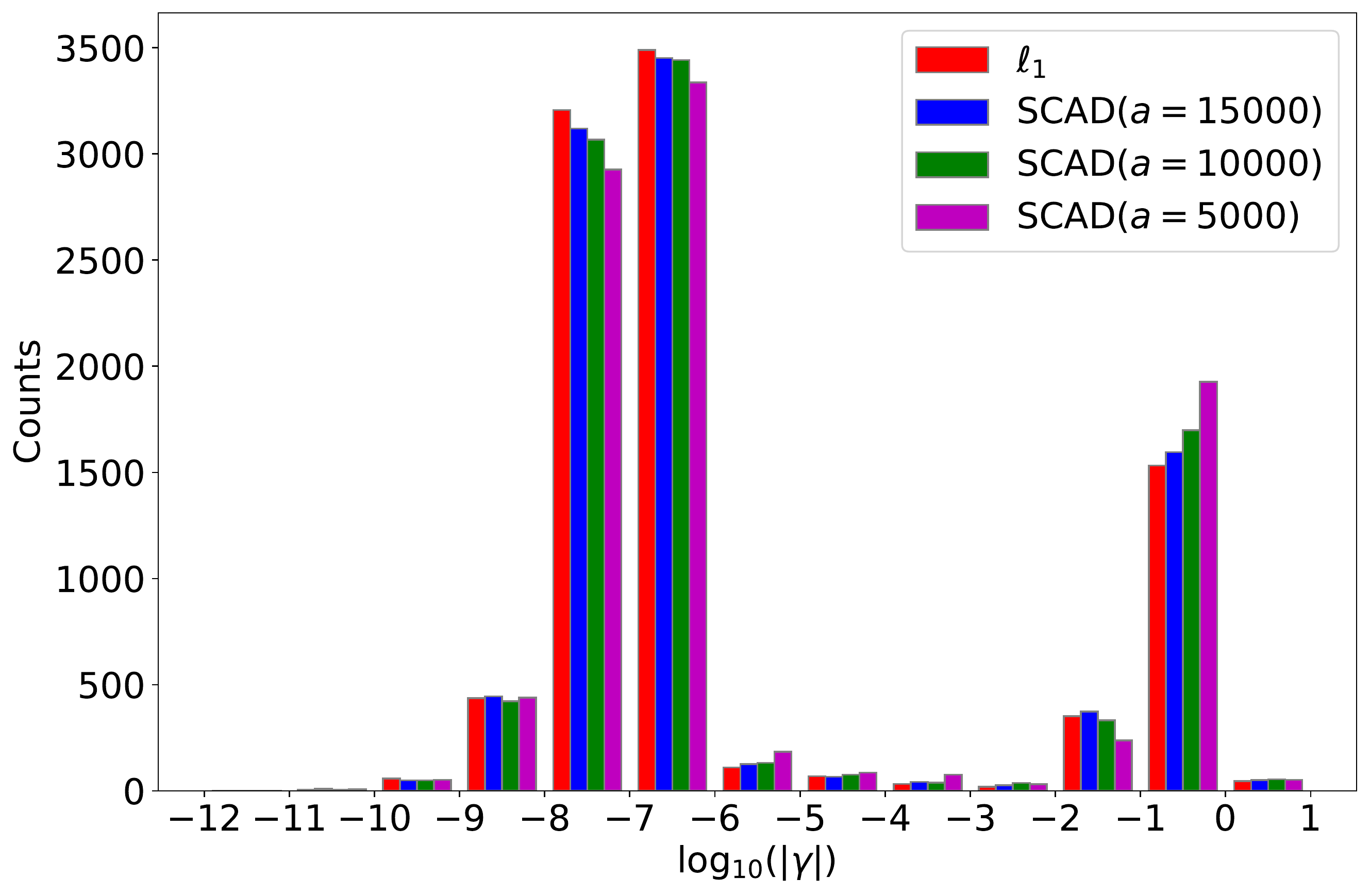}
         \caption{SCAD}
         \label{fig:densenet_SCAD_scaling_cifar10}
     \end{subfigure}
        \caption{Histogram of scaling factors $\gamma$ in DenseNet-40 trained on CIFAR 10. The $x$-axis is $\log_{10}(|\gamma|)$.}
        \label{fig:densenet_scaling_cifar10}
\end{figure*}
\begin{figure*}[h!!!]
\centering
     \begin{subfigure}[b]{0.40\textwidth}
         \centering
         \includegraphics[width=\textwidth]{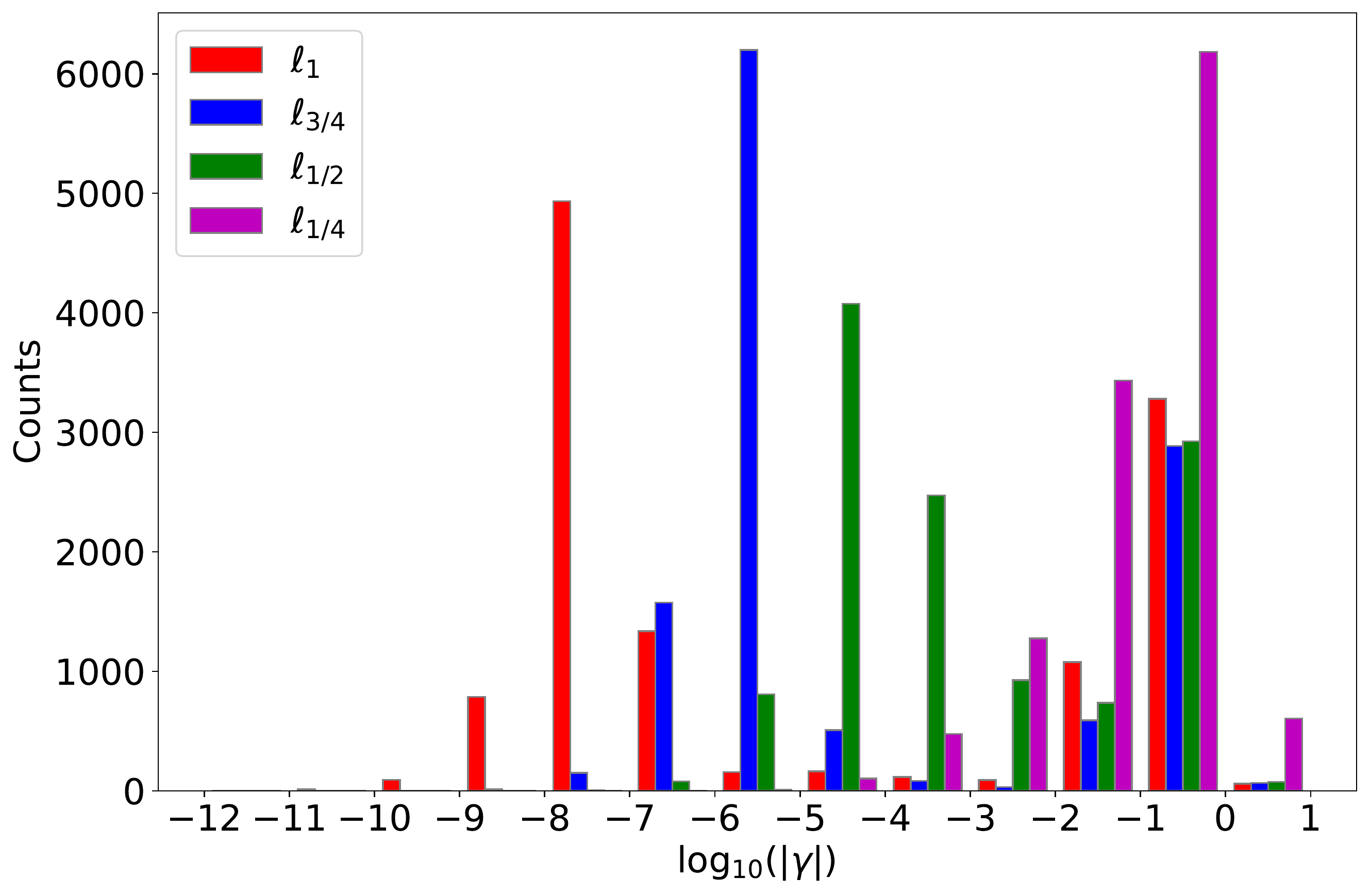}
         \caption{$\ell_p$}
         \label{fig:resnet_Lp_scaling}
     \end{subfigure}
     \begin{subfigure}[b]{0.40\textwidth}
         \centering
         \includegraphics[width=\textwidth]{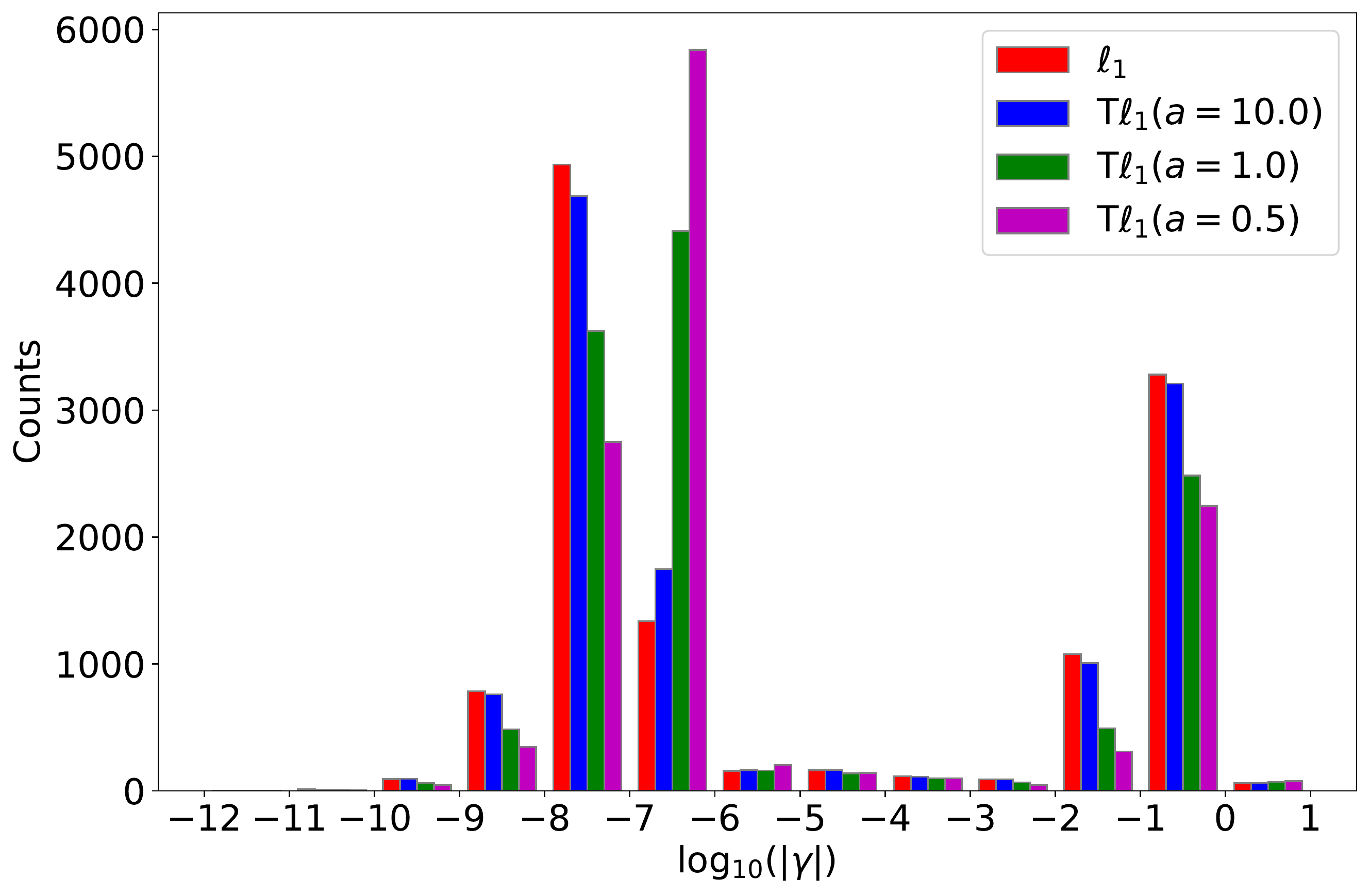}
         \caption{T$\ell_1$}
         \label{fig:resnet_TL1_scaling}
     \end{subfigure}\\
     \begin{subfigure}[b]{0.40\textwidth}
         \centering
         \includegraphics[width=\textwidth]{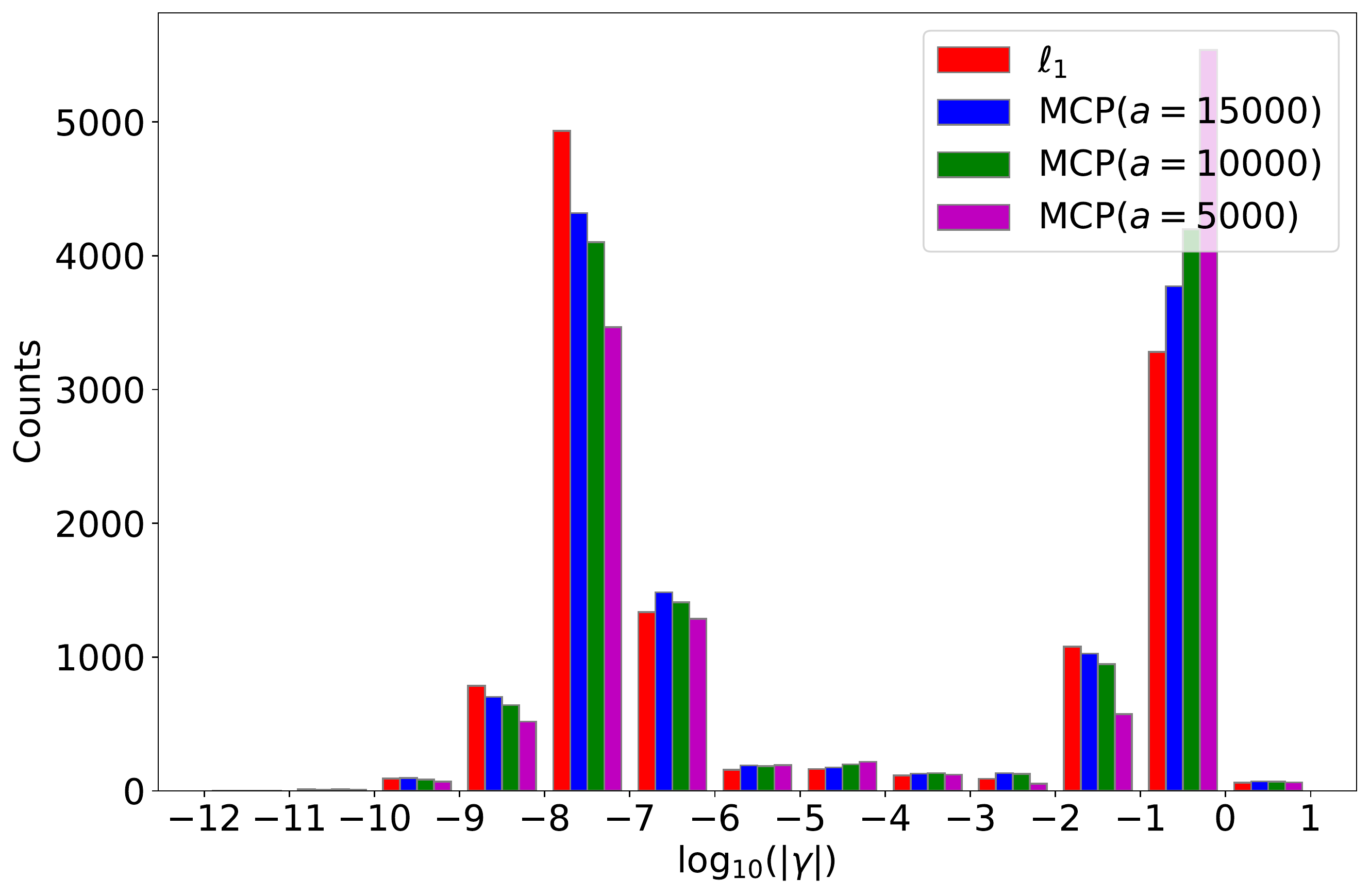}
         \caption{MCP}
         \label{fig:resnet_MCP_scaling}
     \end{subfigure}
     \begin{subfigure}[b]{0.40\textwidth}
         \centering
         \includegraphics[width=\textwidth]{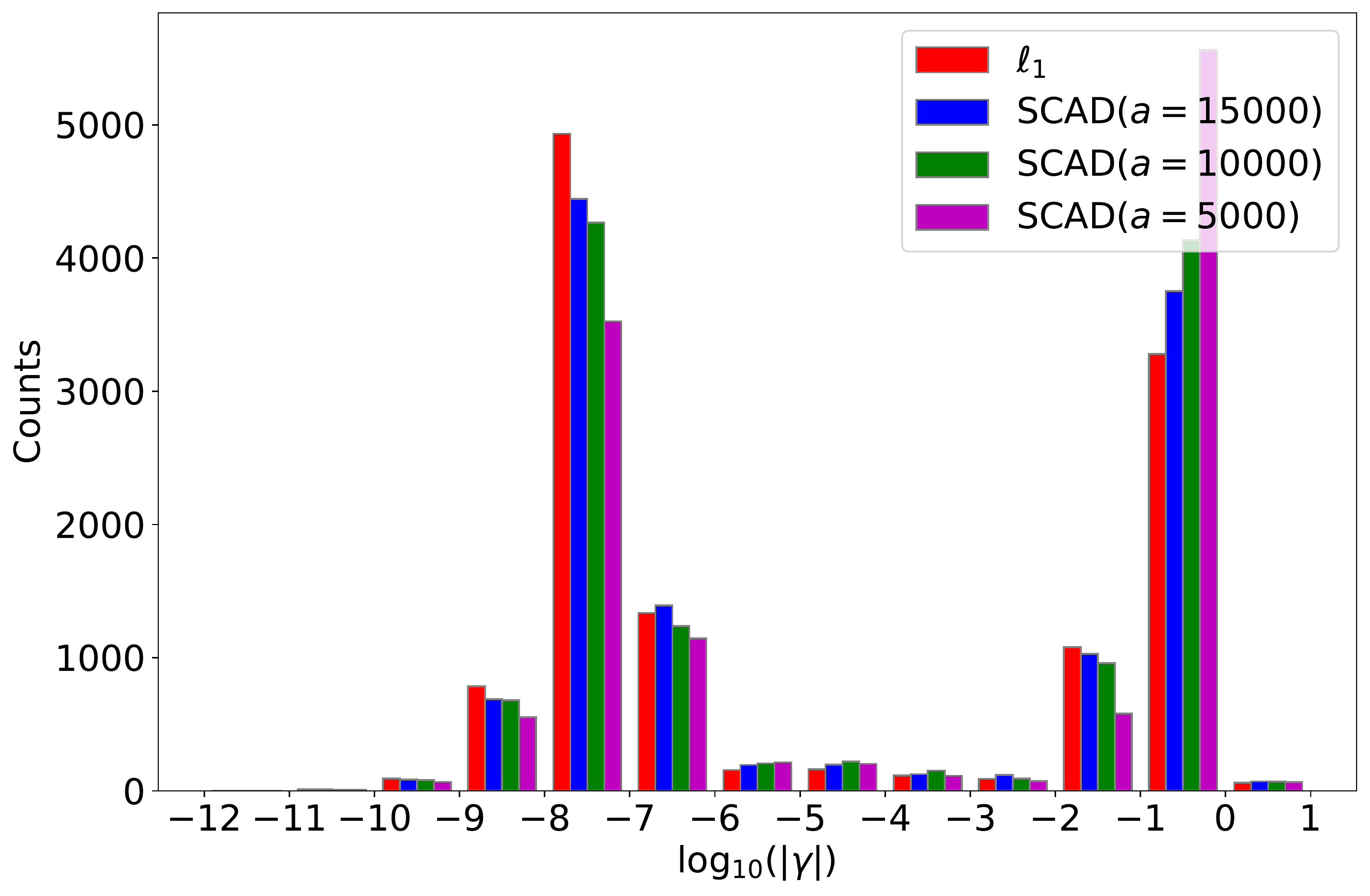}
         \caption{SCAD}
         \label{fig:resnet_SCAD_scaling}
     \end{subfigure}
        \caption{Histogram of scaling factors $\gamma$ in ResNet-164 trained on CIFAR 10. The $x$-axis is $\log_{10}(|\gamma|)$.}
        \label{fig:resnet_scaling_cifar10}
\end{figure*}
\begin{figure*}[h!!!]
\centering
     \begin{subfigure}[b]{0.40\textwidth}
         \centering
         \includegraphics[width=\textwidth]{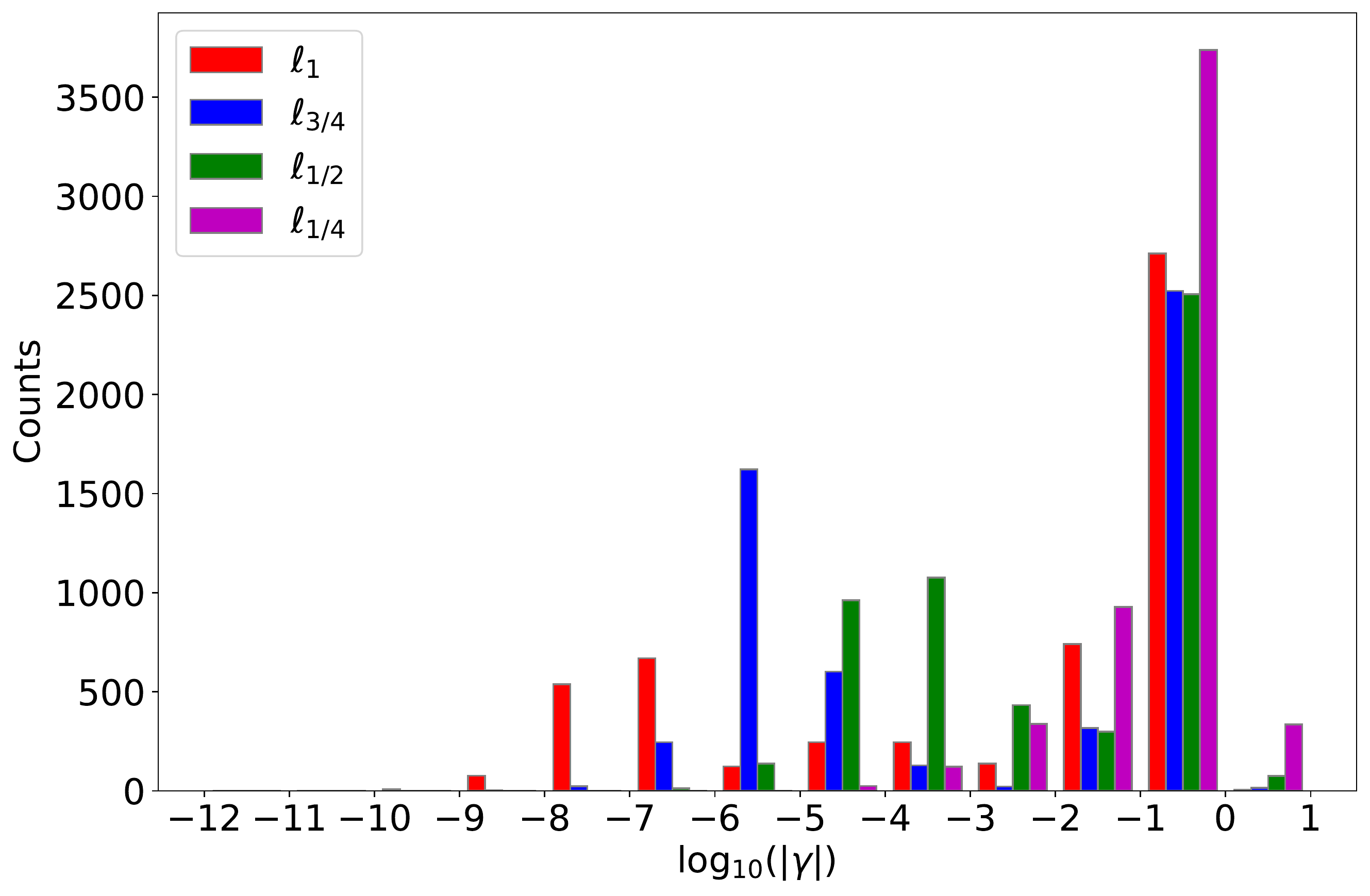}
         \caption{$\ell_p$}
         \label{fig:vgg_lp_scaling_cifar100}
     \end{subfigure}
     \begin{subfigure}[b]{0.40\textwidth}
         \centering
         \includegraphics[width=\textwidth]{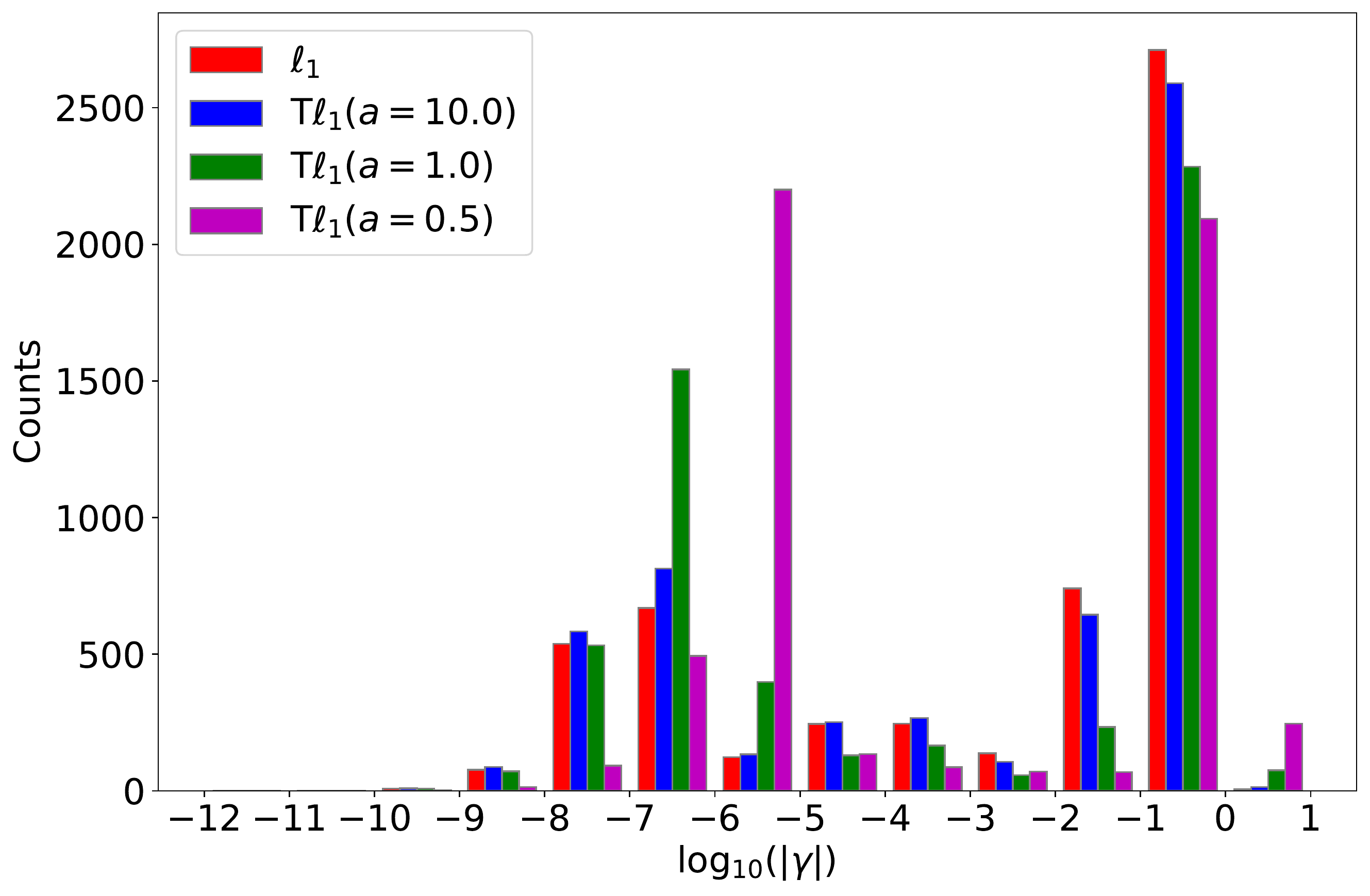}
         \caption{T$\ell_1$}
         \label{fig:vgg_Tl1_scaling_cifar100}
     \end{subfigure}\\
     \begin{subfigure}[b]{0.40\textwidth}
         \centering
         \includegraphics[width=\textwidth]{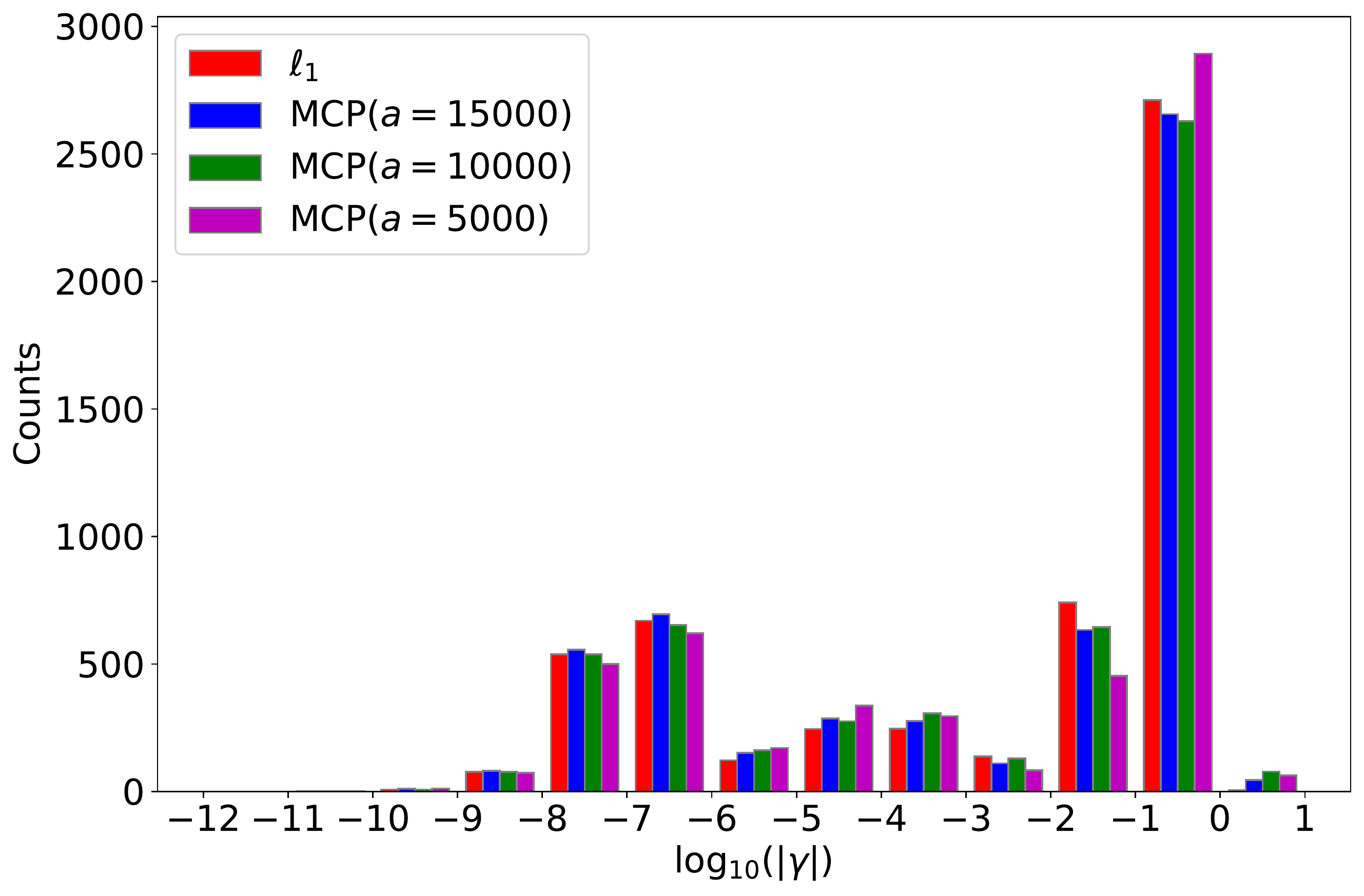}
         \caption{MCP}
         \label{fig:vgg_MCP_scaling_cifar100}
     \end{subfigure}
     \begin{subfigure}[b]{0.40\textwidth}
         \centering
         \includegraphics[width=\textwidth]{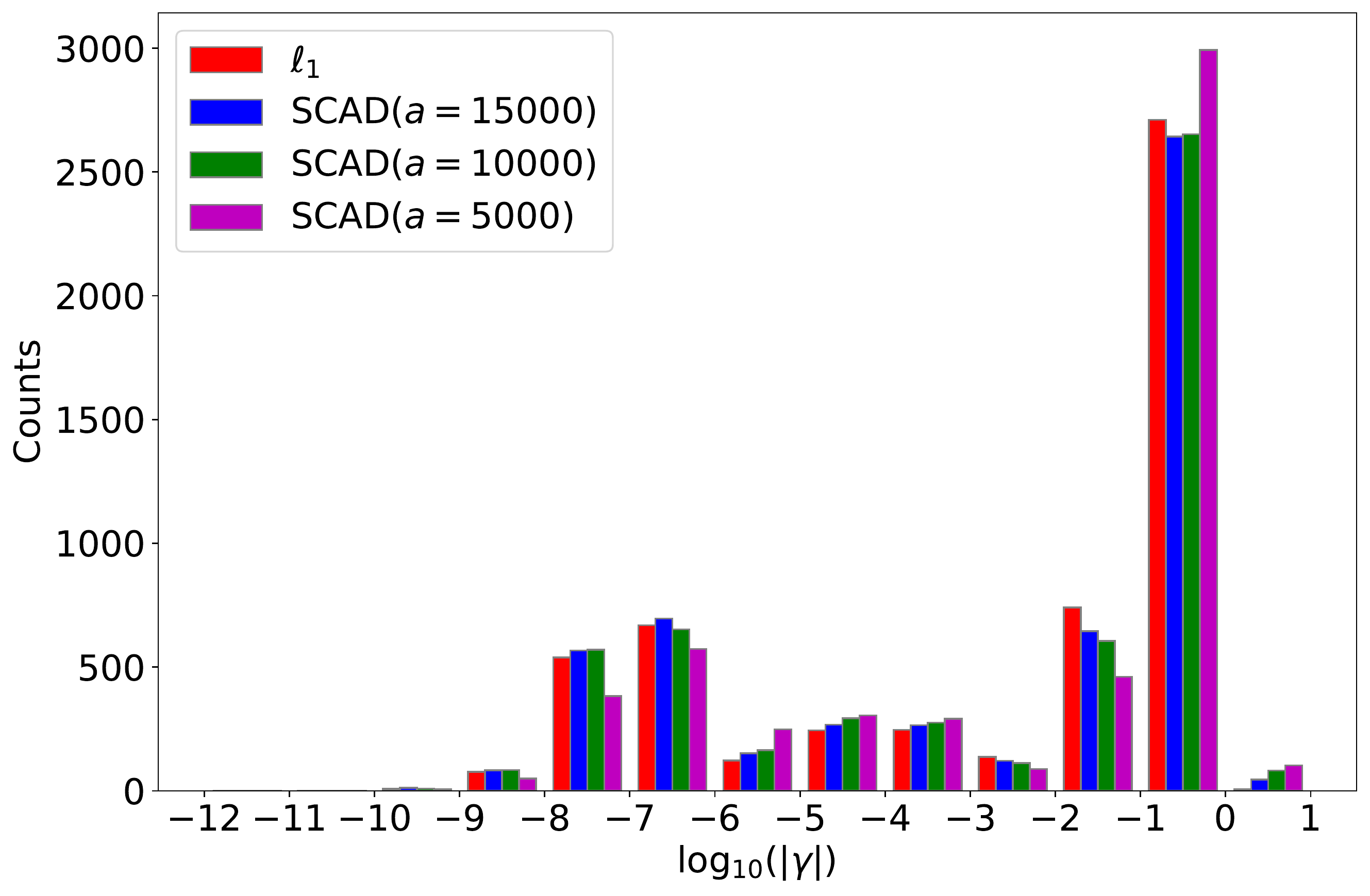}
         \caption{SCAD}
         \label{fig:vgg_SCAD_scaling_cifar100}
     \end{subfigure}
        \caption{Histogram of scaling factors $\gamma$ in VGG-19 trained on CIFAR 100. The $x$-axis is $\log_{10}(|\gamma|)$.}
        \label{fig:vgg_scaling_cifar100}
\end{figure*}
\begin{figure*}[h!!!]
\centering
     \begin{subfigure}[b]{0.40\textwidth}
         \centering
         \includegraphics[width=\textwidth]{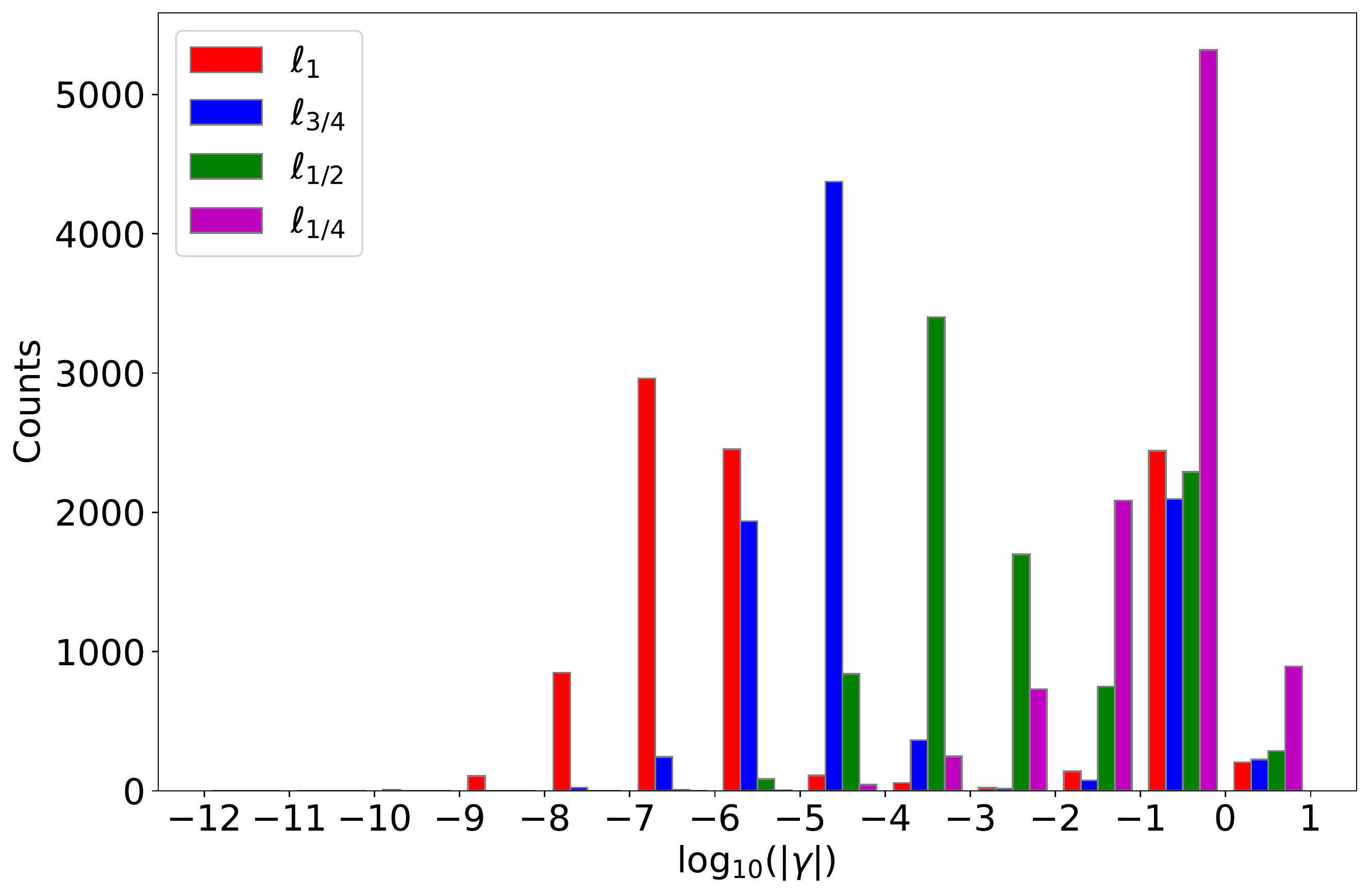}
         \caption{$\ell_p$}
         \label{fig:densenet_lp_scaling_cifar100}
     \end{subfigure}
     \begin{subfigure}[b]{0.40\textwidth}
         \centering
         \includegraphics[width=\textwidth]{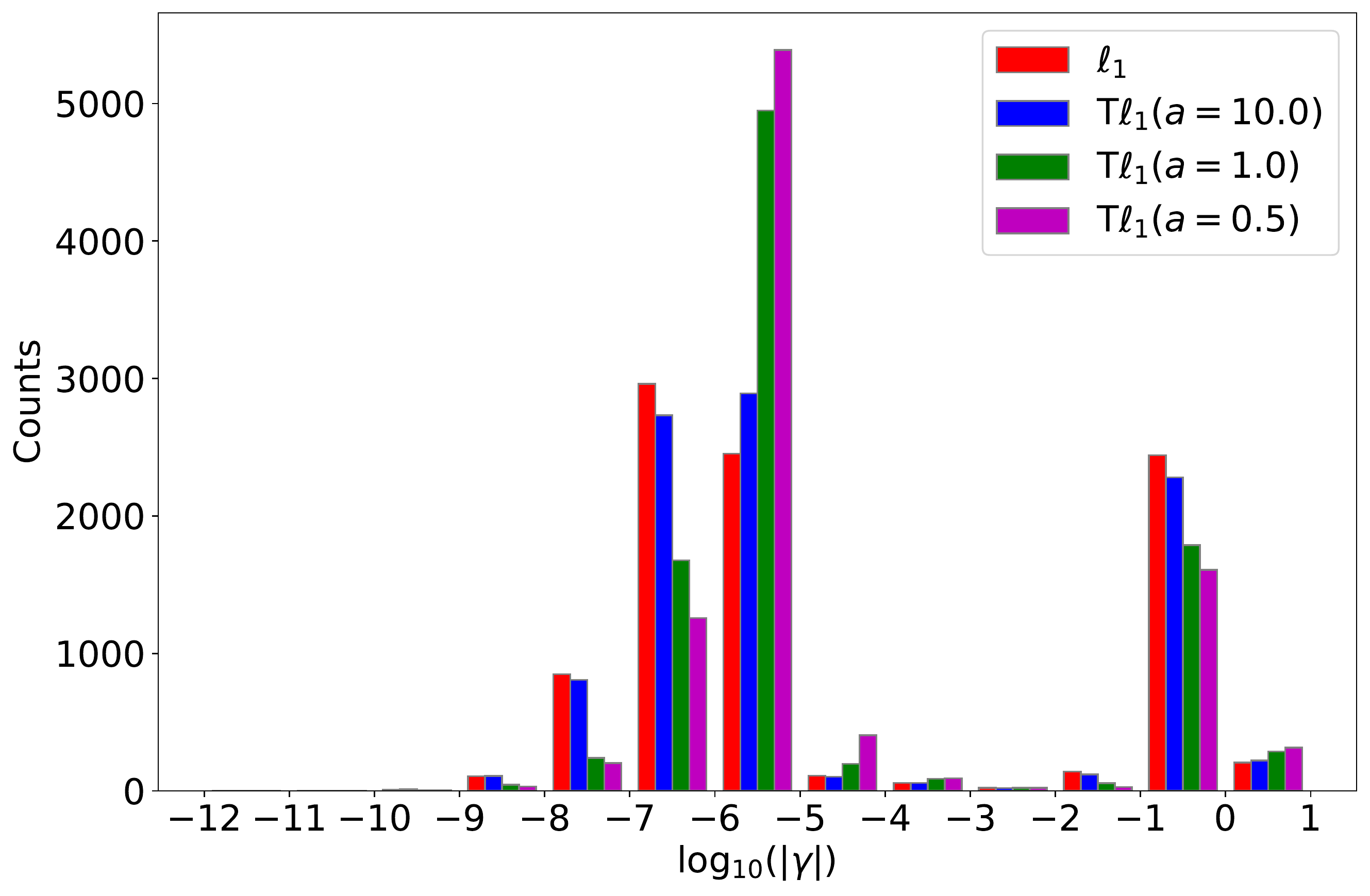}
         \caption{T$\ell_1$}
         \label{fig:densenet_Tl1_scaling_cifar100}
     \end{subfigure}\\
     \begin{subfigure}[b]{0.40\textwidth}
         \centering
         \includegraphics[width=\textwidth]{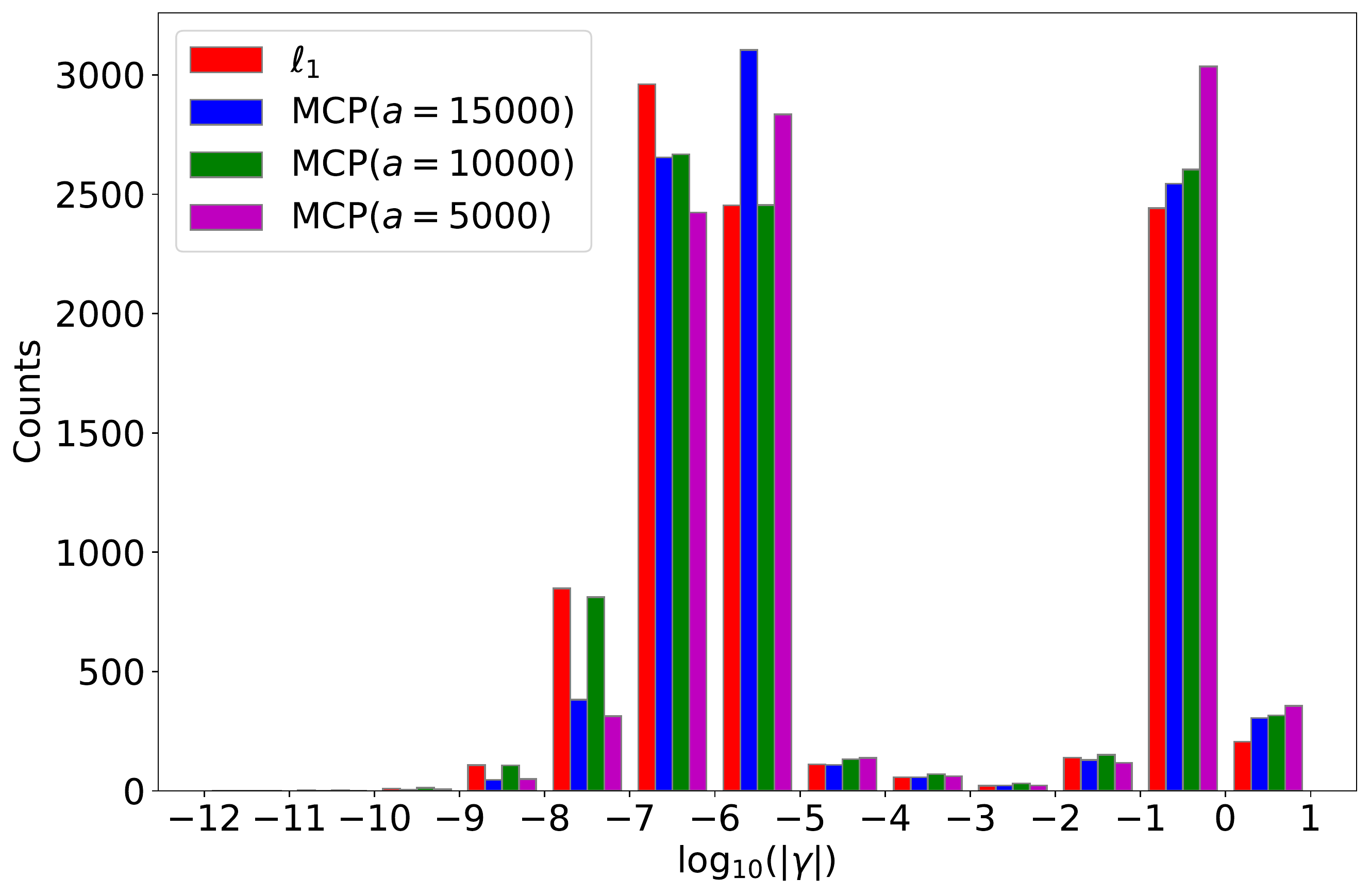}
         \caption{MCP}
         \label{fig:densenet_MCP_scaling_cifar100}
     \end{subfigure}
     \begin{subfigure}[b]{0.40\textwidth}
         \centering
         \includegraphics[width=\textwidth]{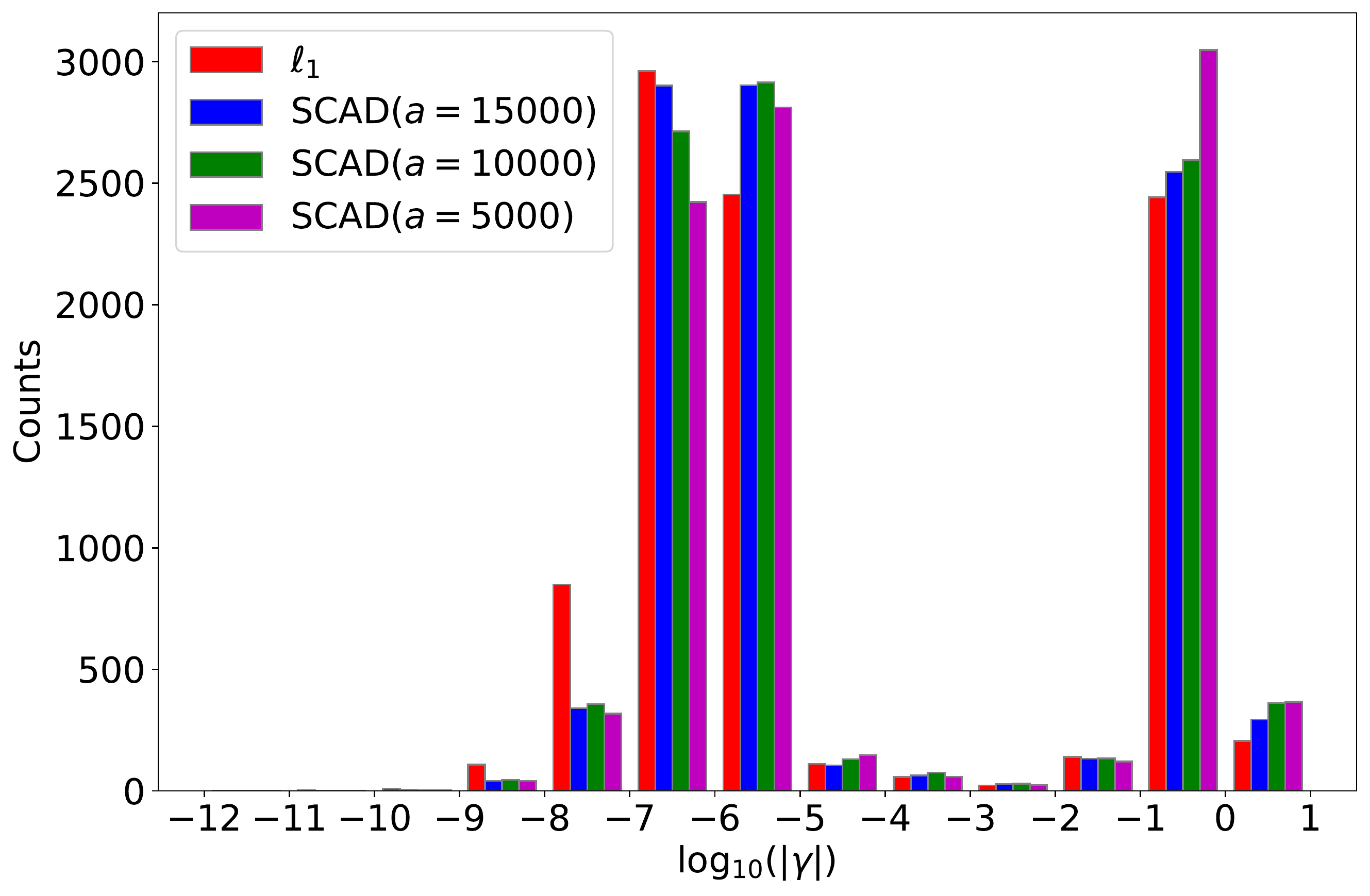}
         \caption{SCAD}
         \label{fig:densenet_SCAD_scaling_cifar100}
     \end{subfigure}
        \caption{Histogram of scaling factors $\gamma$ in DenseNet-40 trained on CIFAR 100. The $x$-axis is $\log_{10}(|\gamma|)$.}
        \label{fig:densenet_scaling_cifar100}
\end{figure*}
\begin{figure*}[h!!!]
\centering
     \begin{subfigure}[b]{0.40\textwidth}
         \centering
         \includegraphics[width=\textwidth]{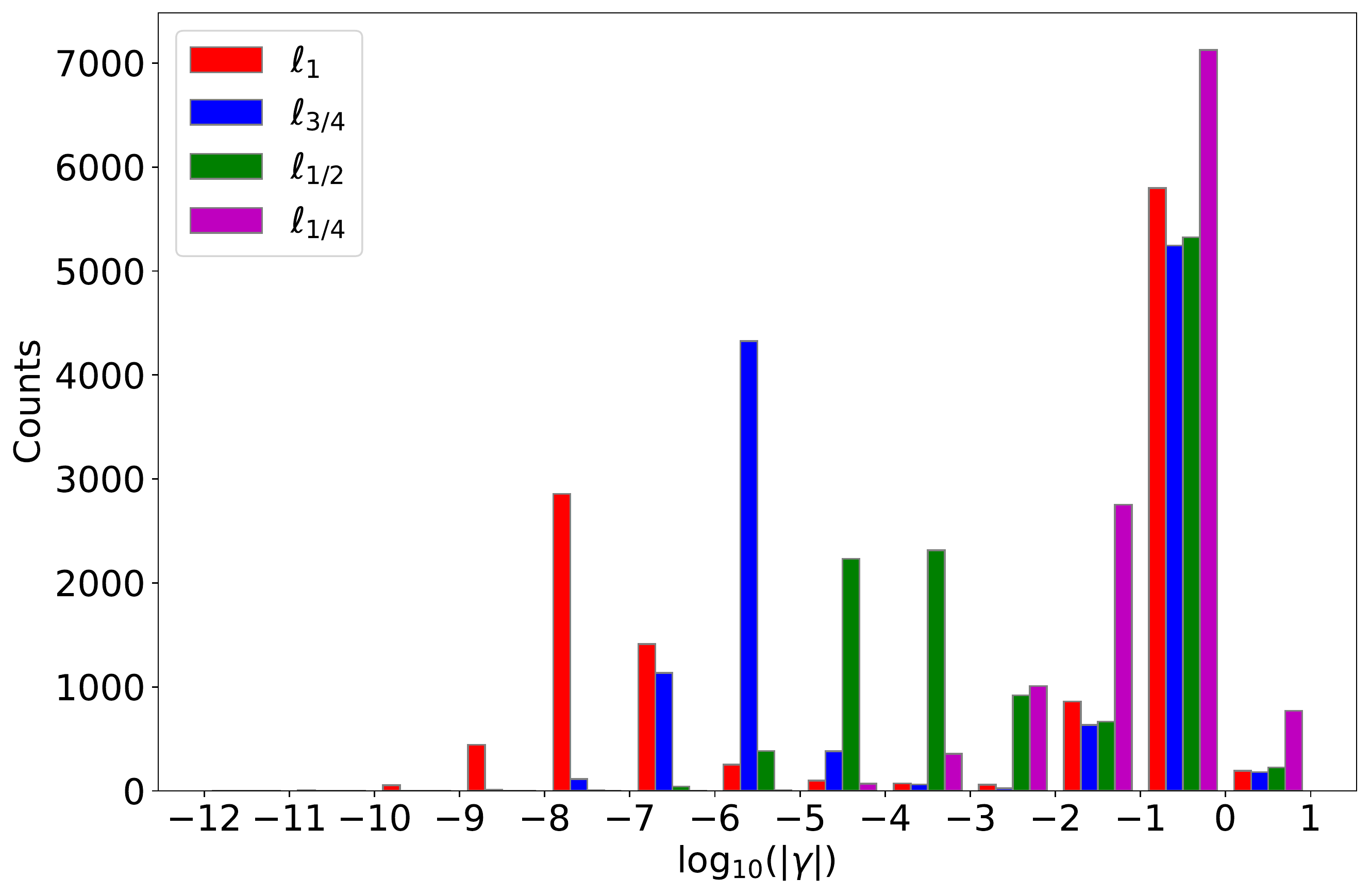}
         \caption{$\ell_p$}
         \label{fig:resnet_lp_scaling_cifar100}
     \end{subfigure}
     \begin{subfigure}[b]{0.40\textwidth}
         \centering
         \includegraphics[width=\textwidth]{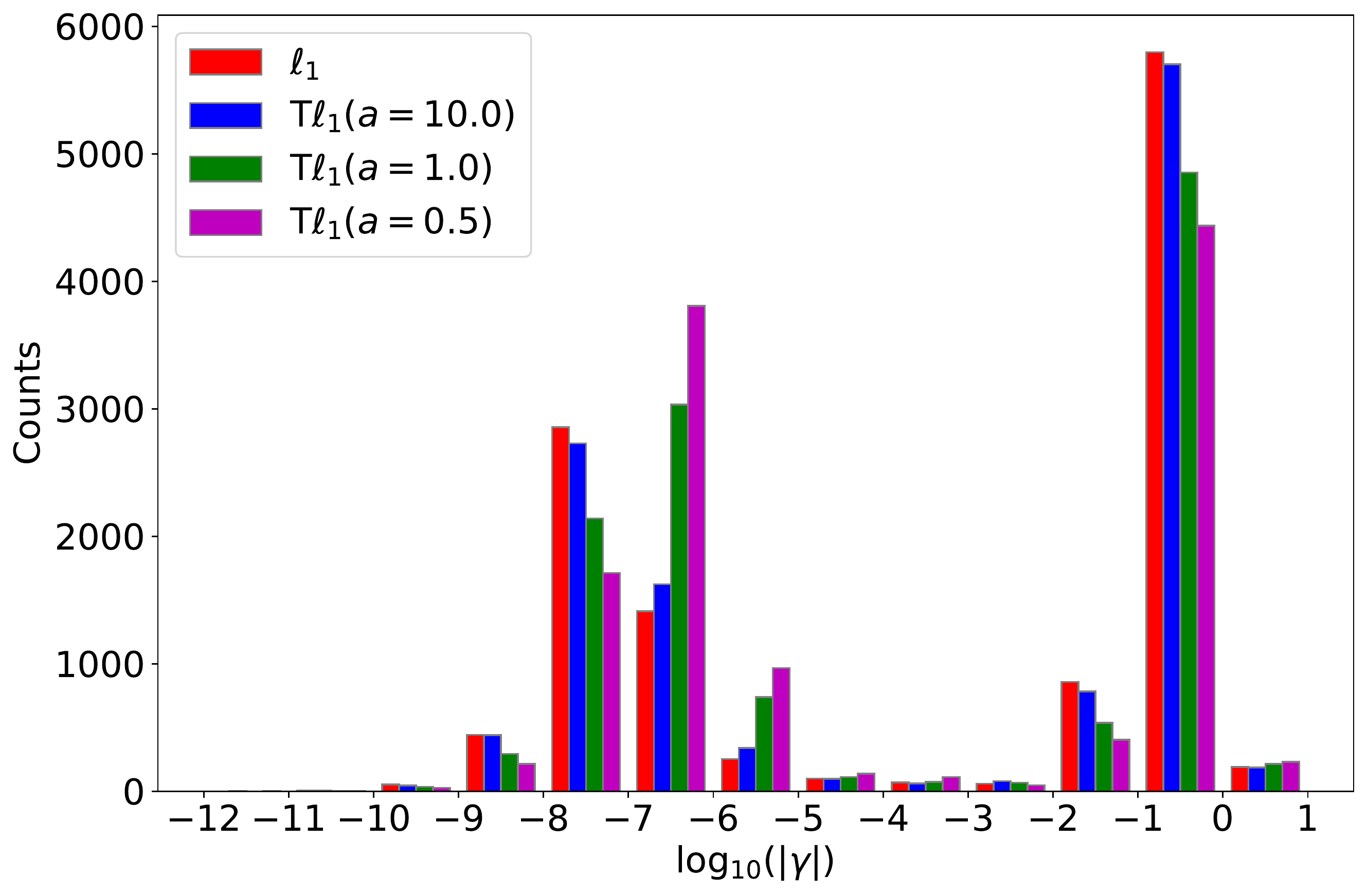}
         \caption{T$\ell_1$}
         \label{fig:resnet_Tl1_scaling_cifar100}
     \end{subfigure}\\
     \begin{subfigure}[b]{0.40\textwidth}
         \centering
         \includegraphics[width=\textwidth]{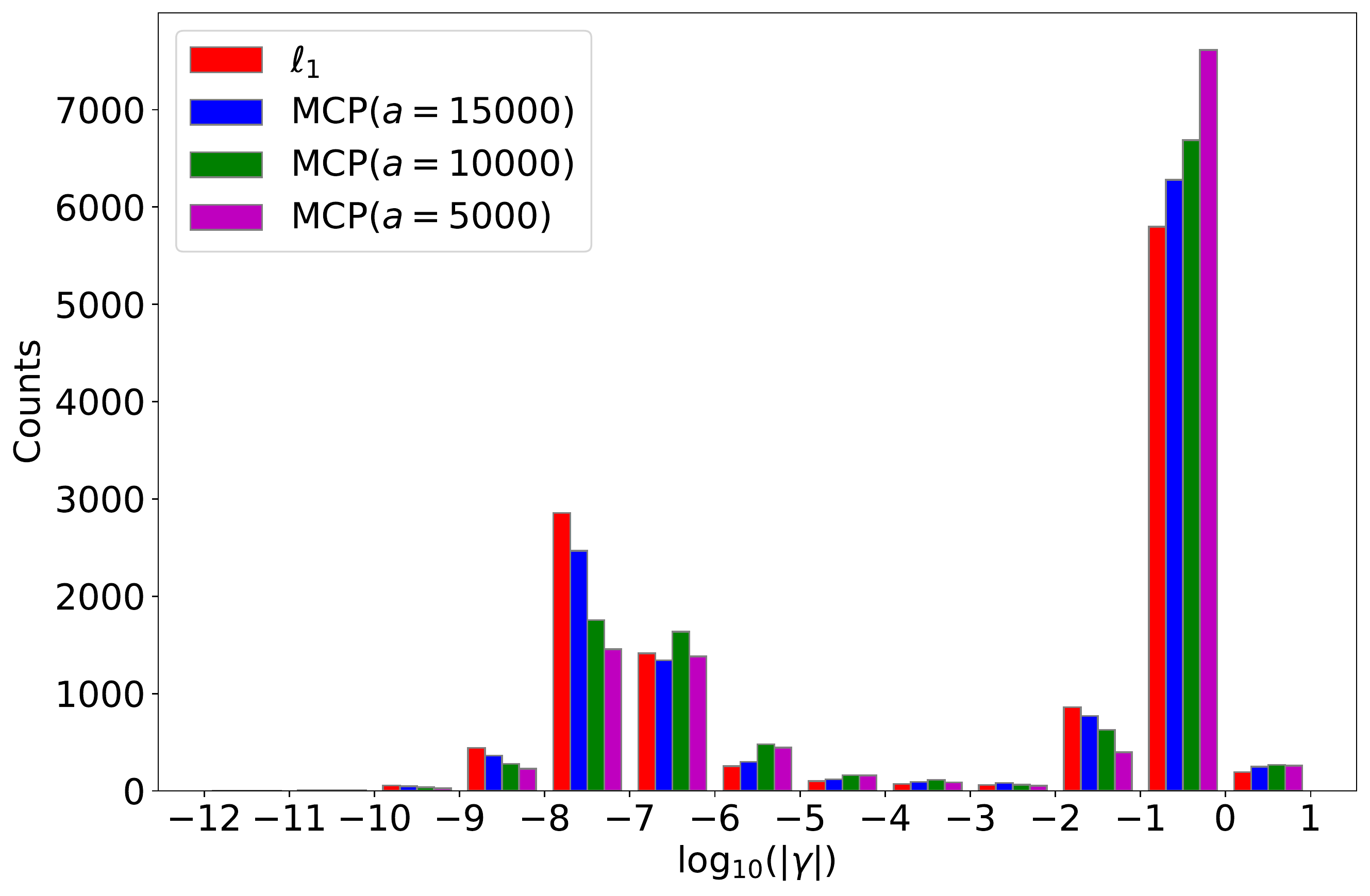}
         \caption{MCP}
         \label{fig:resnet_MCP_scaling_cifar100}
     \end{subfigure}
     \begin{subfigure}[b]{0.40\textwidth}
         \centering
         \includegraphics[width=\textwidth]{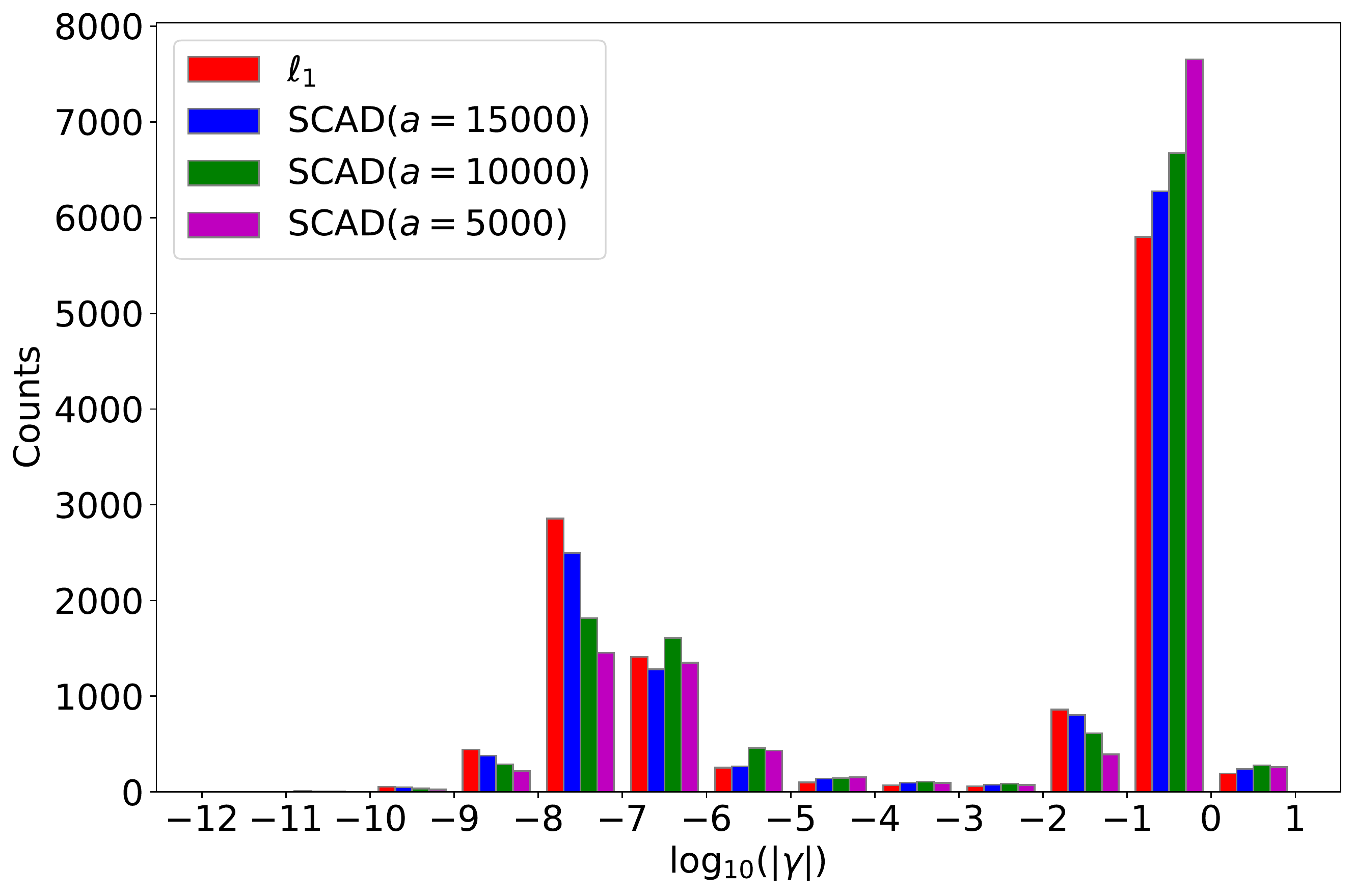}
         \caption{SCAD}
         \label{fig:resnet_SCAD_scaling_cifar100}
     \end{subfigure}
        \caption{Histogram of scaling factors $\gamma$ in ResNet-164 trained on CIFAR 100. The $x$-axis is $\log_{10}(|\gamma|)$.}
        \label{fig:resnet_scaling_cifar100}
\end{figure*}
\begin{figure*}[h!!!]
\centering
     \begin{subfigure}[b]{0.40\textwidth}
         \centering
         \includegraphics[width=\textwidth]{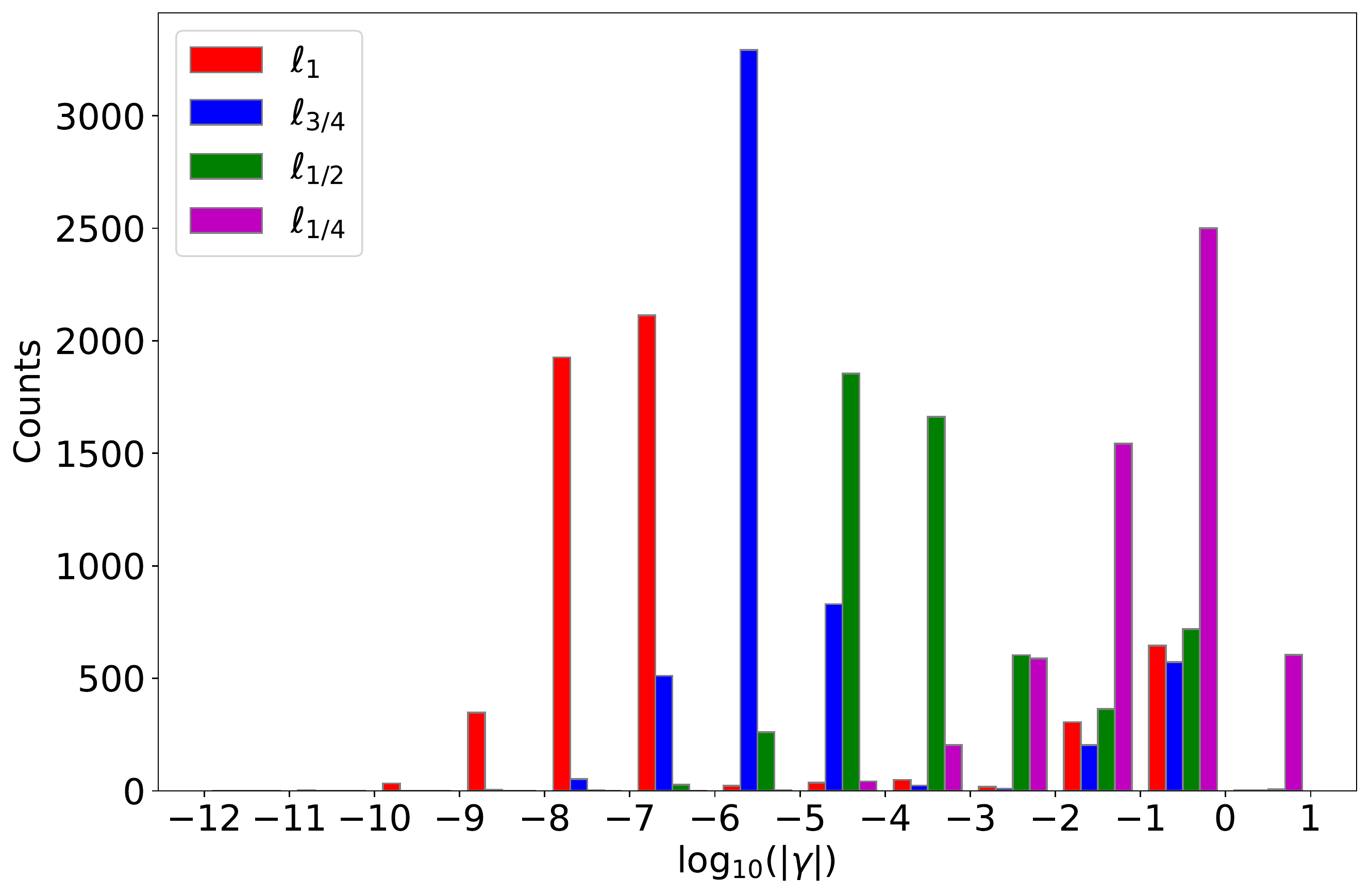}
         \caption{$\ell_p$}
         \label{fig:vgg_lp_scaling_SVHN}
     \end{subfigure}
     \begin{subfigure}[b]{0.40\textwidth}
         \centering
         \includegraphics[width=\textwidth]{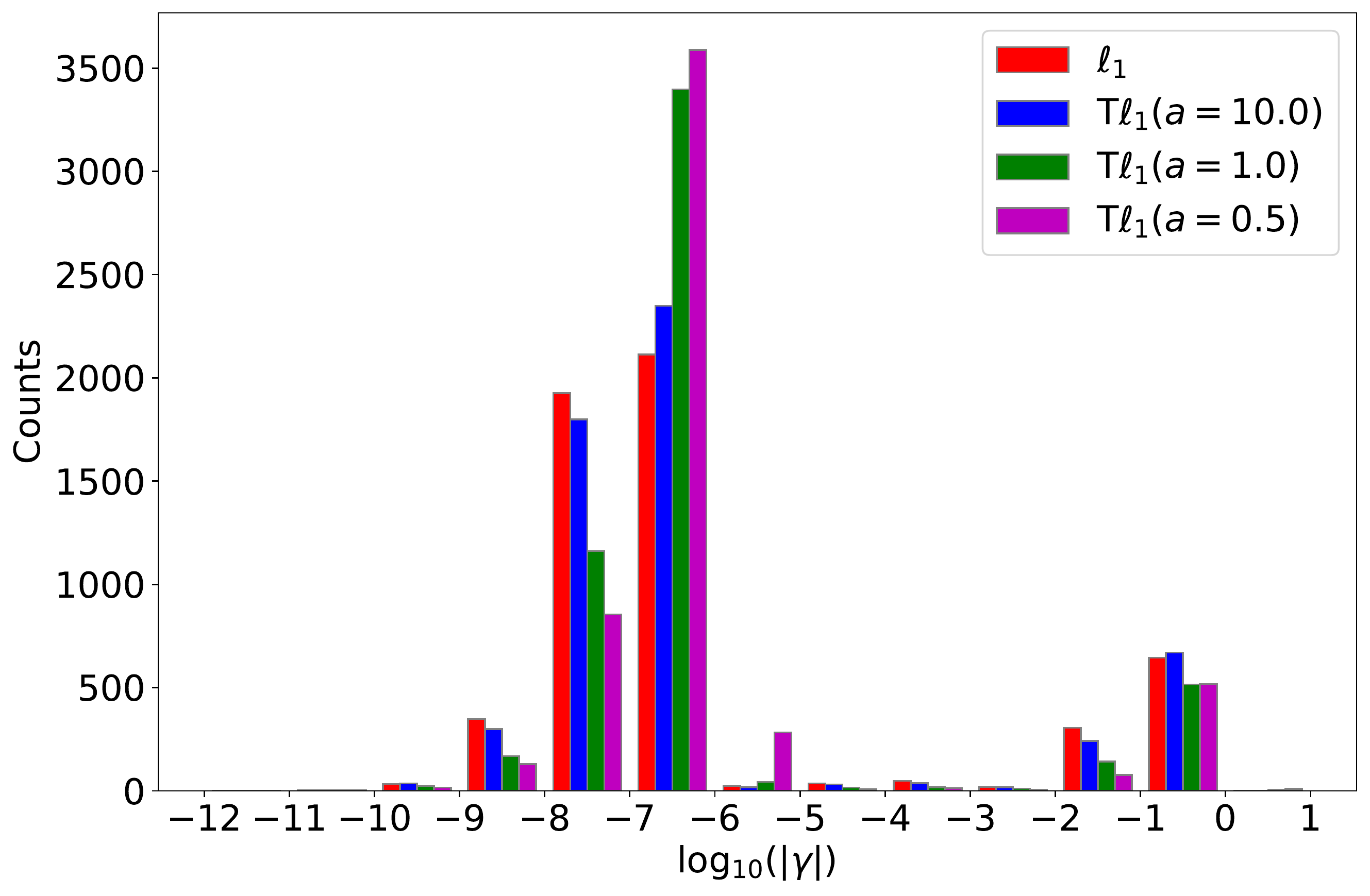}
         \caption{T$\ell_1$}
         \label{fig:vgg_Tl1_scaling_SVHN}
     \end{subfigure}\\
     \begin{subfigure}[b]{0.40\textwidth}
         \centering
         \includegraphics[width=\textwidth]{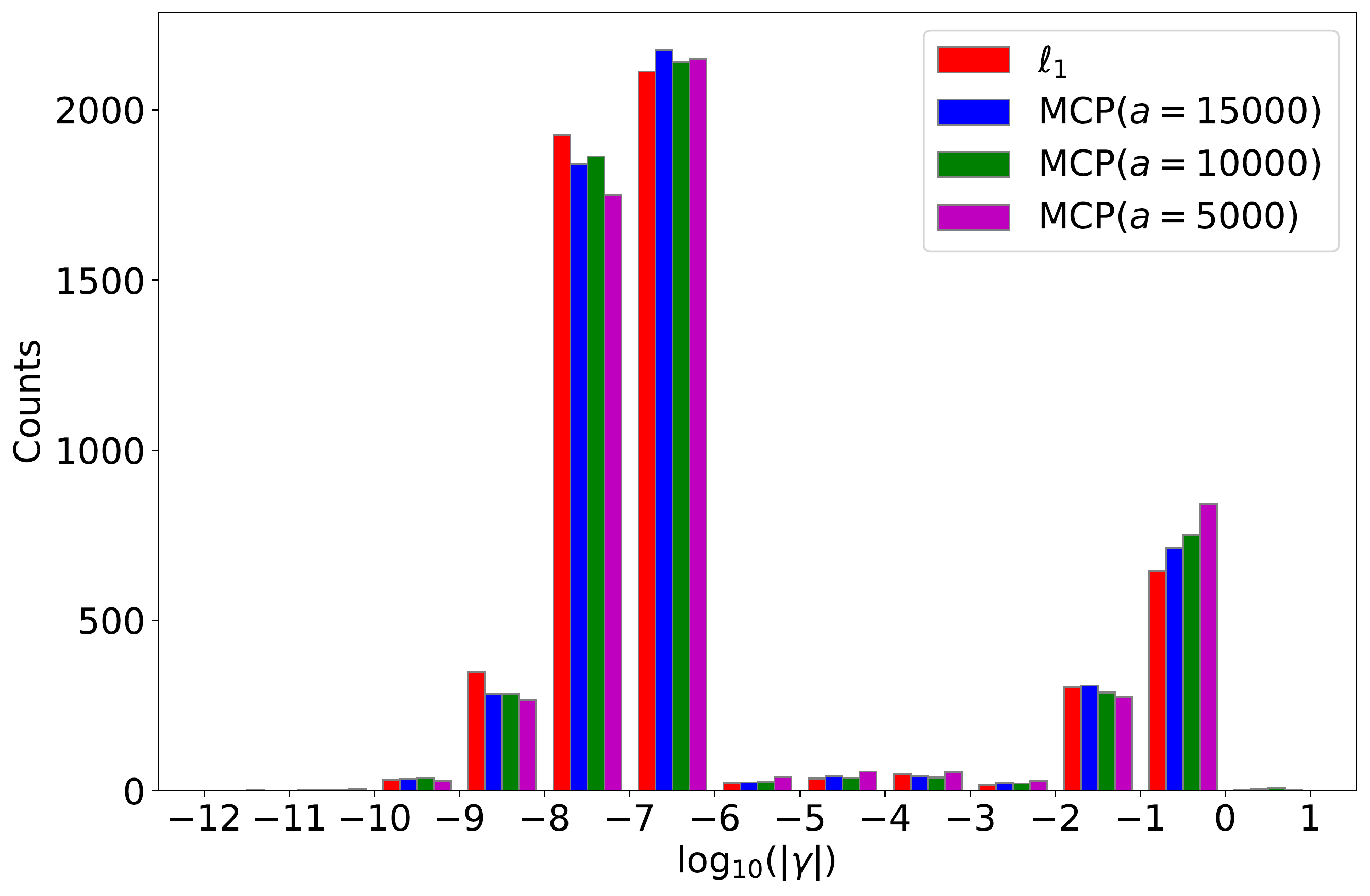}
         \caption{MCP}
         \label{fig:vgg_MCP_scaling_SVHN}
     \end{subfigure}
     \begin{subfigure}[b]{0.40\textwidth}
         \centering
         \includegraphics[width=\textwidth]{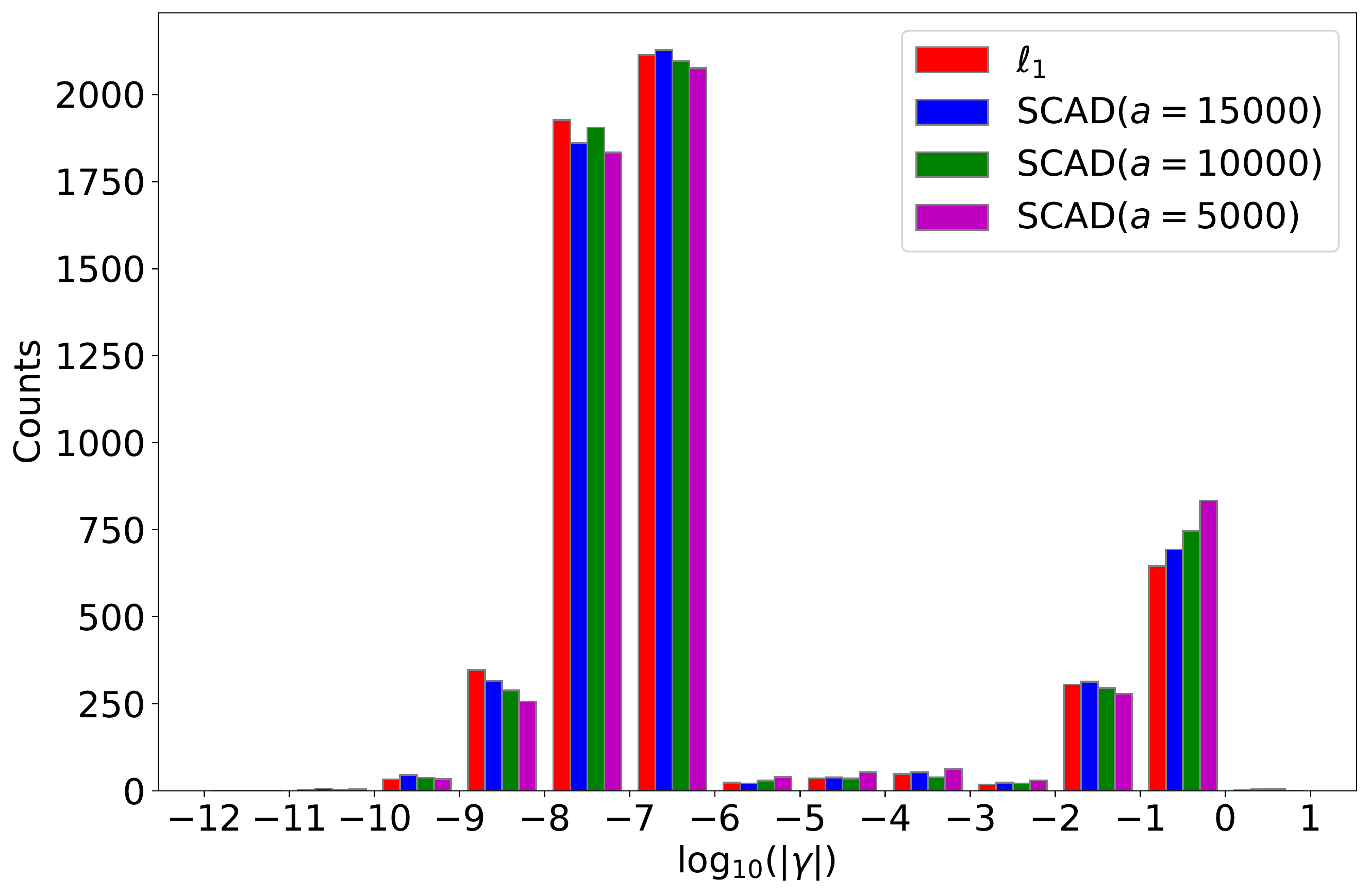}
         \caption{SCAD}
         \label{fig:vgg_SCAD_scaling_SVHN}
     \end{subfigure}
        \caption{Histogram of scaling factors $\gamma$ in VGG-19 trained on SVHN. The $x$-axis is $\log_{10}(|\gamma|)$.}
        \label{fig:vgg_scaling_SVHN}
\end{figure*}
\begin{figure*}[h!!!]
\centering
     \begin{subfigure}[b]{0.40\textwidth}
         \centering
         \includegraphics[width=\textwidth]{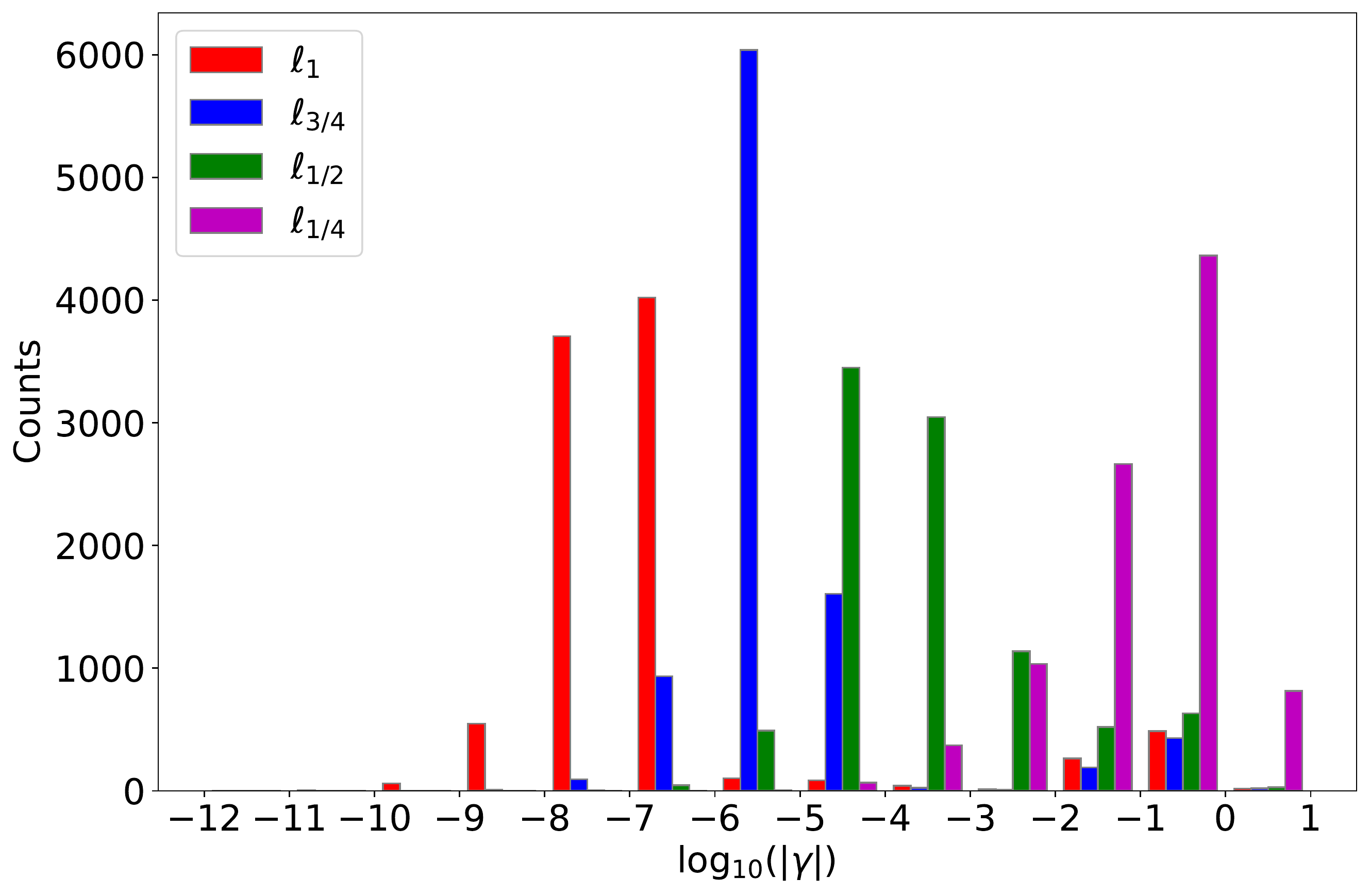}
         \caption{$\ell_p$}
         \label{fig:densenet_lp_scaling_SVHN}
     \end{subfigure}
     \begin{subfigure}[b]{0.40\textwidth}
         \centering
         \includegraphics[width=\textwidth]{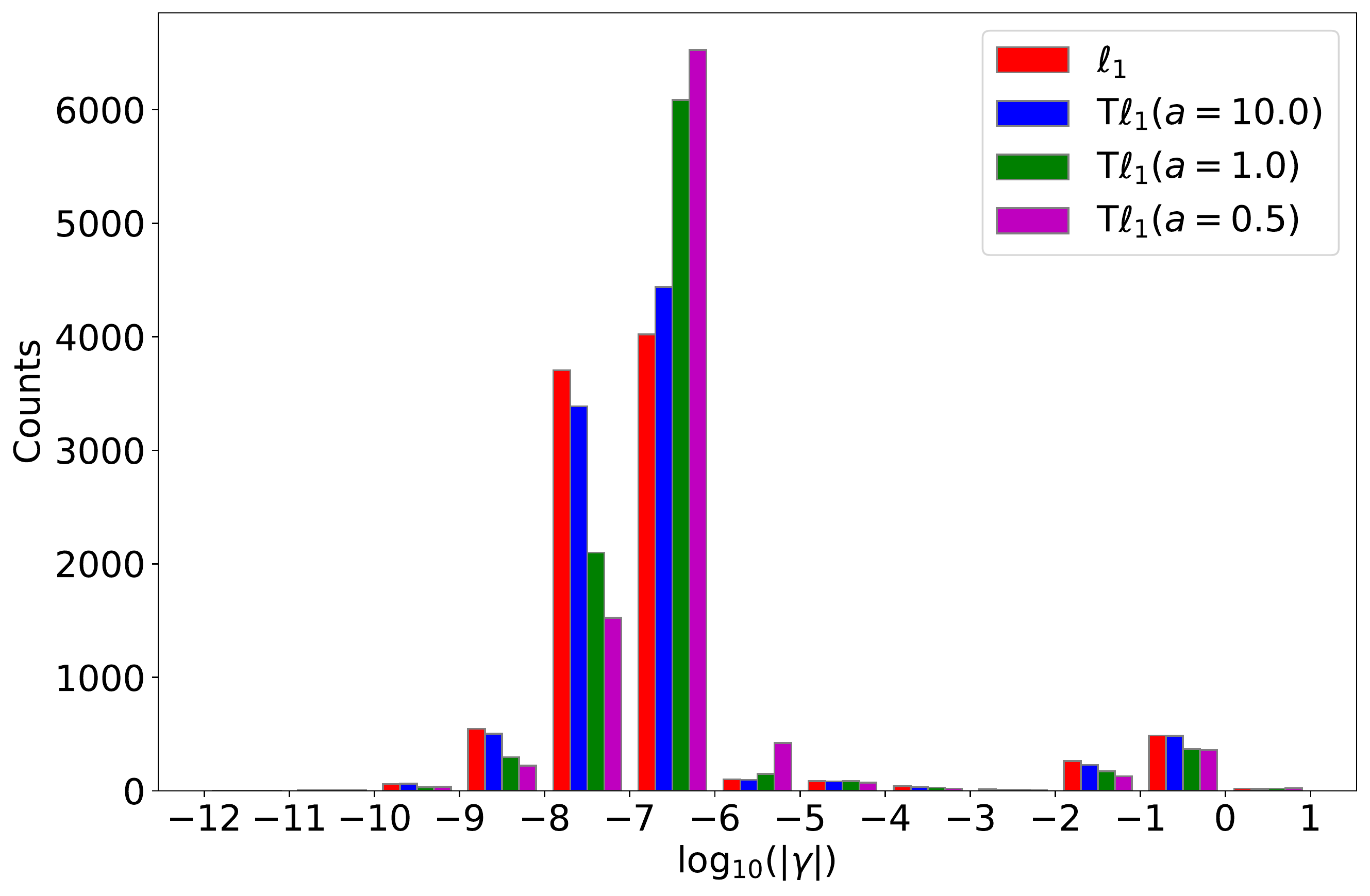}
         \caption{T$\ell_1$}
         \label{fig:densenet_Tl1_scaling_SVHN}
     \end{subfigure}\\
     \begin{subfigure}[b]{0.40\textwidth}
         \centering
         \includegraphics[width=\textwidth]{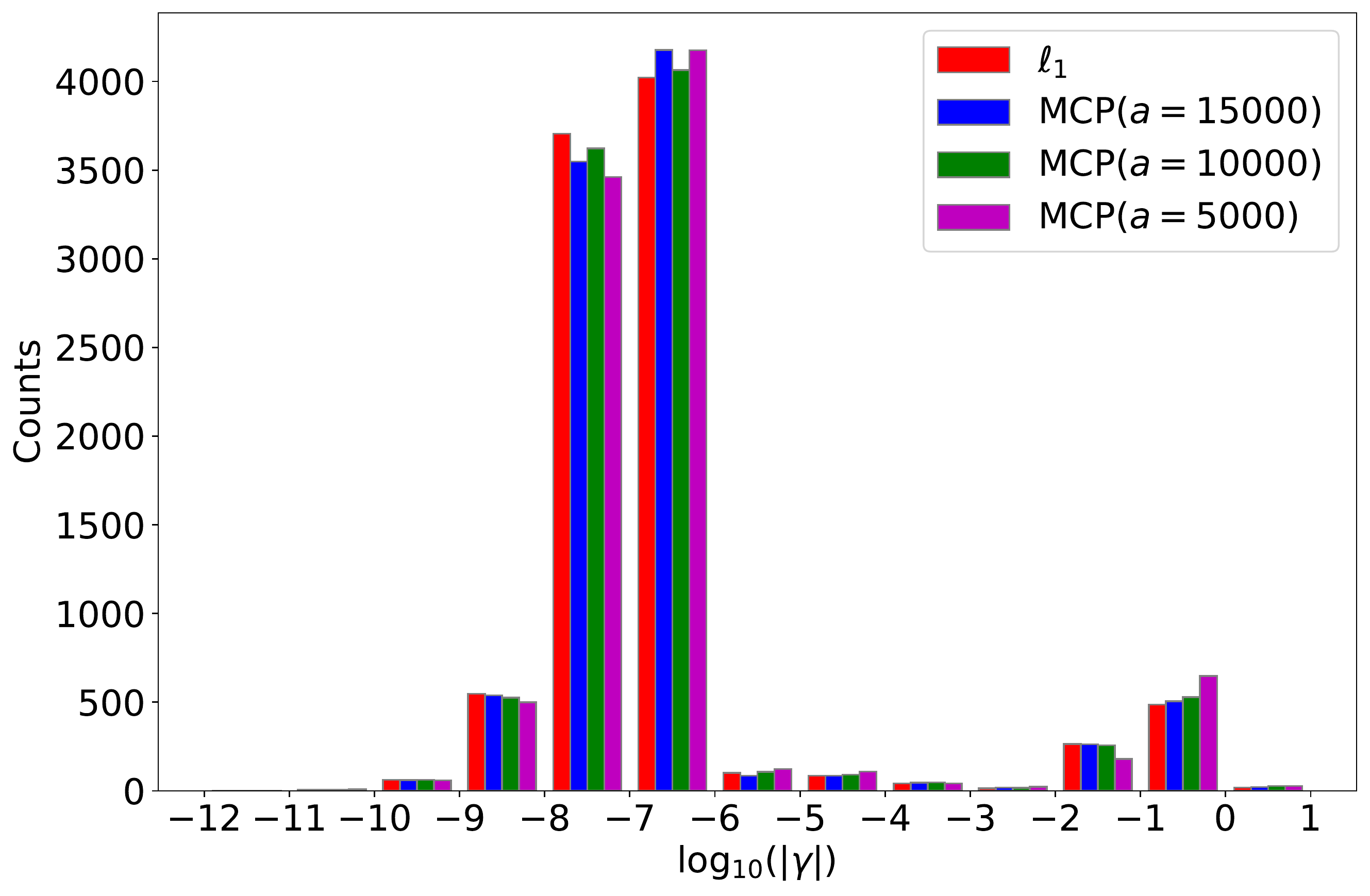}
         \caption{MCP}
         \label{fig:densenet_SVHN_scaling}
     \end{subfigure}
     \begin{subfigure}[b]{0.40\textwidth}
         \centering
         \includegraphics[width=\textwidth]{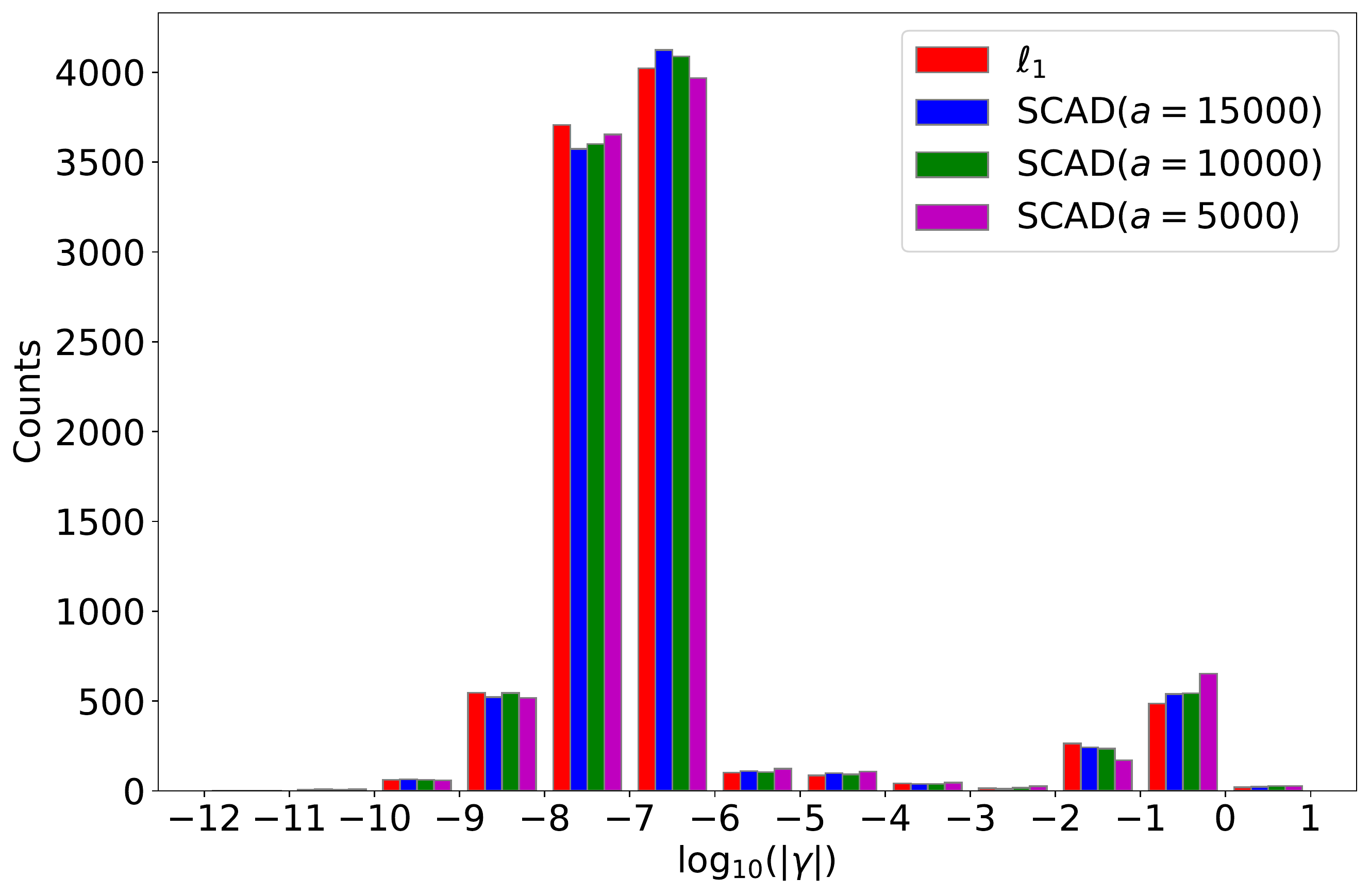}
         \caption{SCAD}
         \label{fig:densenet_SCAD_scaling_SVHN}
     \end{subfigure}
        \caption{Histogram of scaling factors $\gamma$ in DenseNet-40 trained on SVHN. The $x$-axis is $\log_{10}(|\gamma|)$.}
        \label{fig:densenet_scaling_SVHN}
\end{figure*}
\begin{figure*}[h!!!]
\centering
     \begin{subfigure}[b]{0.40\textwidth}
         \centering
         \includegraphics[width=\textwidth]{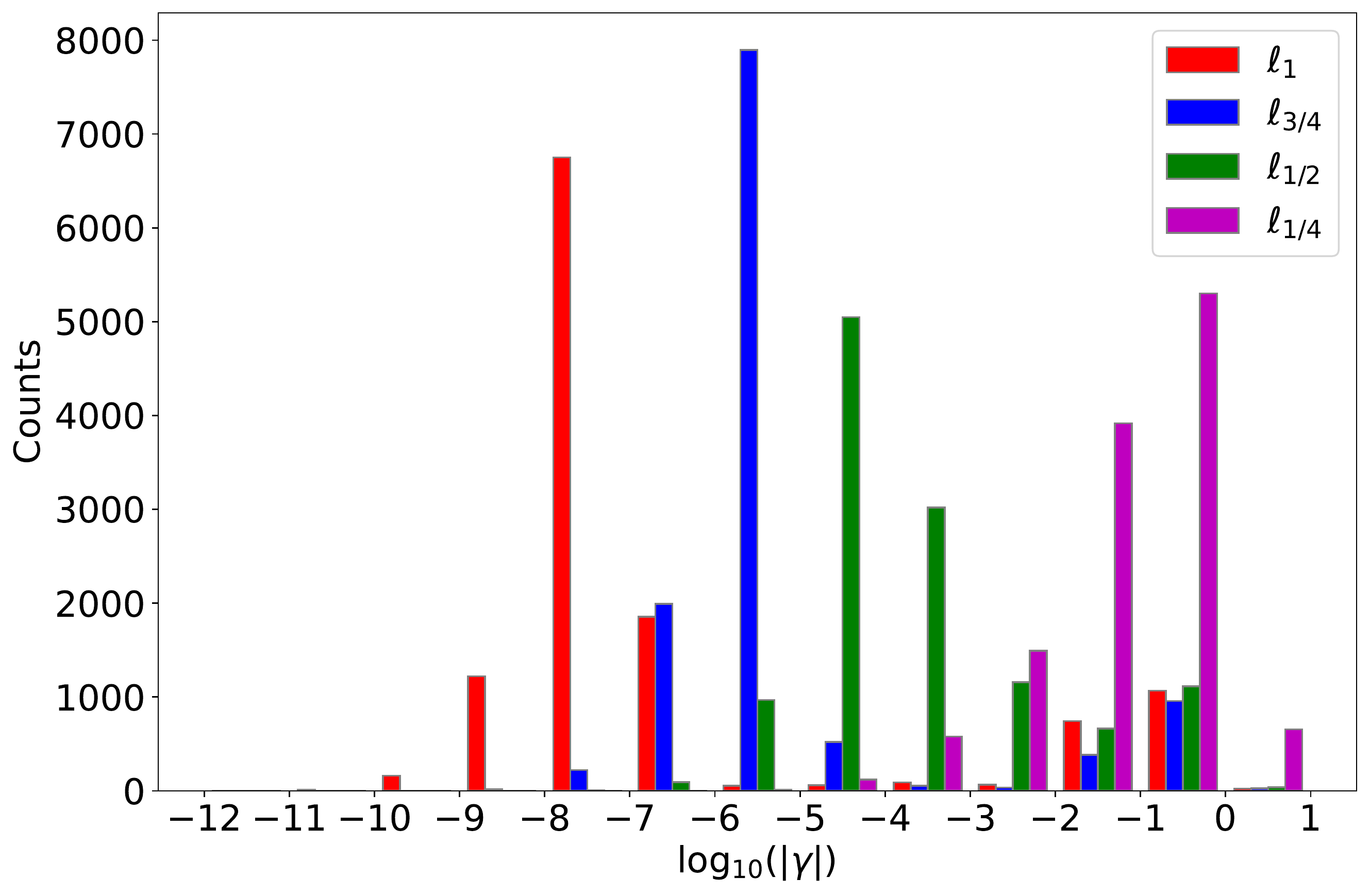}
         \caption{$\ell_p$}
         \label{fig:resnet_lp_scaling_SVHN}
     \end{subfigure}
     \begin{subfigure}[b]{0.40\textwidth}
         \centering
         \includegraphics[width=\textwidth]{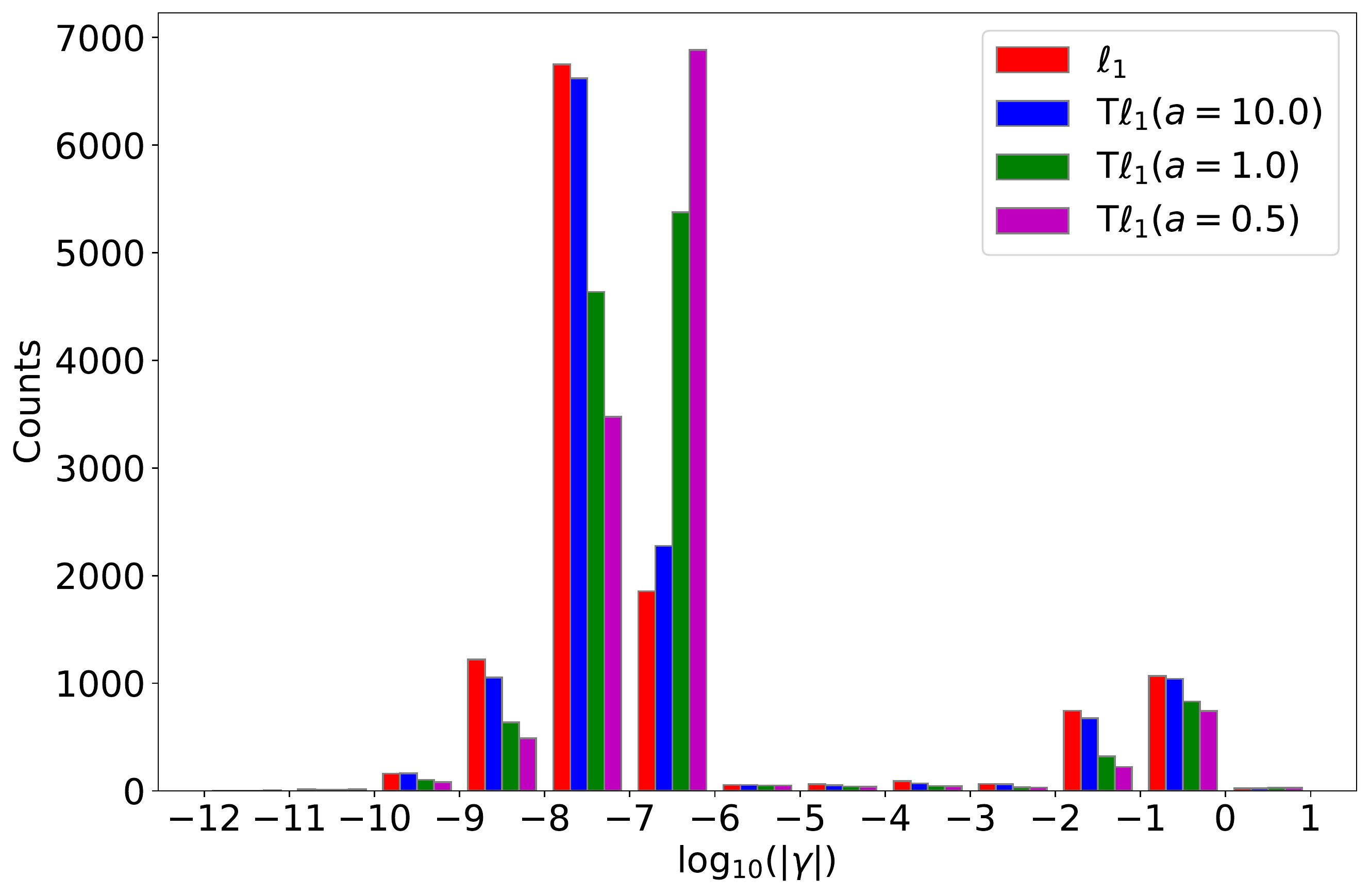}
         \caption{T$\ell_1$}
         \label{fig:resnet_Tl1_scaling_SVHN}
     \end{subfigure}\\
     \begin{subfigure}[b]{0.40\textwidth}
         \centering
         \includegraphics[width=\textwidth]{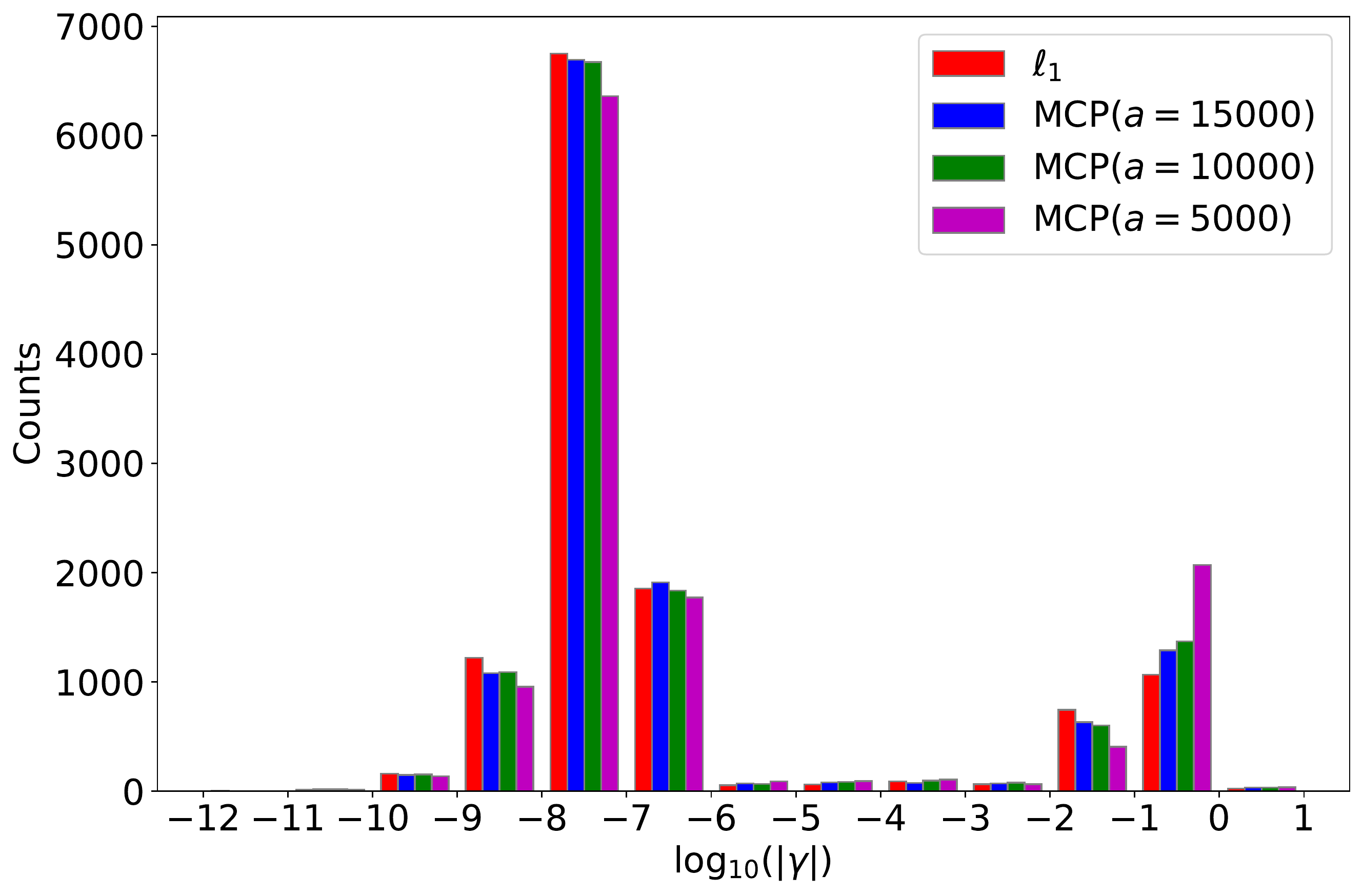}
         \caption{MCP}
         \label{fig:resnet_MCP_scaling_SVHN}
     \end{subfigure}
     \begin{subfigure}[b]{0.40\textwidth}
         \centering
         \includegraphics[width=\textwidth]{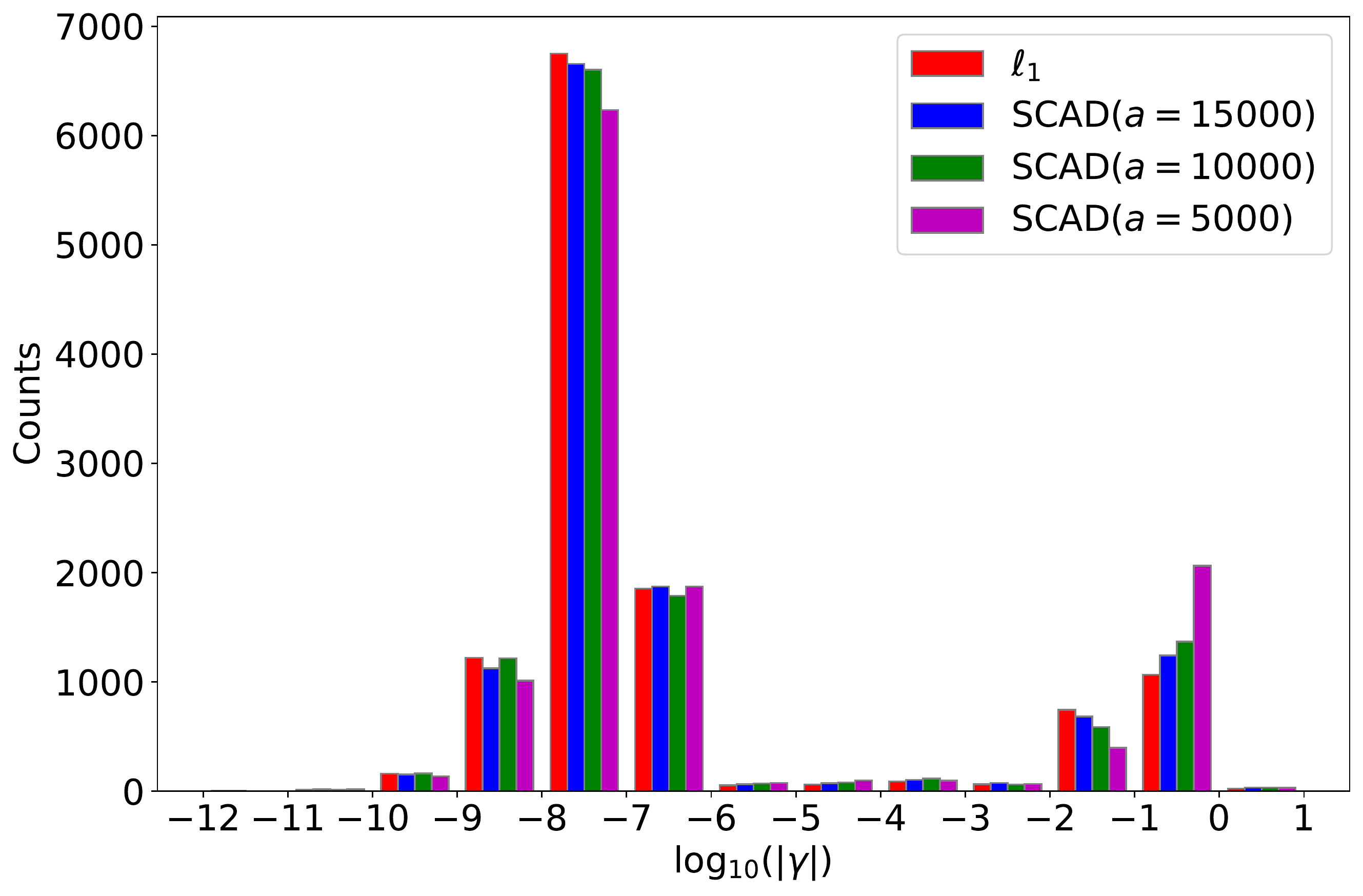}
         \caption{SCAD}
         \label{fig:resnet_SCAD_scaling_SVHN}
     \end{subfigure}
        \caption{Histogram of scaling factors $\gamma$ in ResNet-164 trained on SVHN. The $x$-axis is $\log_{10}(|\gamma|)$.}
        \label{fig:resnet_scaling_SVHN}
\end{figure*}

In order to better understand how $\ell_1$ and the nonconvex regularizers affect the scaling factors $\gamma$, we plot histograms of the counts of the $\log_{10}(|\gamma|)$ averaged from the five models trained for each model and regularizer. Figures \ref{fig:vgg_scaling_cifar10}-\ref{fig:resnet_scaling_SVHN} provide the histograms while Table \ref{tab:scaling_factor} records the average number of scaling factors whose magnitudes are less than $10^{-6}$ and more than $10^{-6}$. The value $10^{-6}$ is chosen because generally, any value below it has negligible effect on the numerical computation \cite{aggarwal2018neural}.

For CIFAR 10, Figures \ref{fig:vgg_scaling_cifar10}-\ref{fig:resnet_scaling_cifar10} show the histograms while Table \ref{tab:cifar10_scale} provides the average counts of the scaling factors based on their magnitudes. For all three networks, we observe the following phenomena. MCP and SCAD have similar scaling factor distributions as $\ell_1$ across all given values of $a$. Moreover, MCP, SCAD, and $\ell_1$ have similar number of scaling factors whose magnitudes are less than $10^{-6}$ as verified by Table \ref{tab:cifar10_scale}. This may explain why their compression rates are similar to $\ell_1$ in our earlier analyses. For $\ell_p$, we see that $\ell_{3/4}$ has most of its scaling factors within the interval $(10^{-6}, 10^{-5})$. As $p$ decreases, the values of the scaling factors tend farther away from 0. In fact, majority of the scaling factors for $\ell_{1/2}$ and $\ell_{1/4}$ are at least $10^{-6}$ in magnitude. Specifically for $\ell_{1/4}$, most of the scaling factors have absolute values at least 0.10. Hence, we can see why $\ell_{1/4}$ is sensitive to channel pruning. Lastly, for T$\ell_1$, more scaling factors decrease towards 0 in magnitude as $a$ decreases. Moreover, we observe that most of the scaling factors are accumulated within the interval $(10^{-7}, 10^{-6})$. Because T$\ell_1$ causes more scaling factors to decrease towards 0 in magnitude, this might explain why T$\ell_1$ is robust against channel pruning.

For CIFAR 100, Figures \ref{fig:vgg_scaling_cifar100}-\ref{fig:resnet_scaling_cifar100} show the histograms of the scaling factors while Table \ref{tab:cifar100_scale} records the average counts by magnitudes. Because CIFAR 100 is a more difficult classification dataset compared to CIFAR 10, most of the scaling factors appear to be within the interval $(10^{-1}, 1)$. However, DenseNet-40 shows bimodal distributions for T$\ell_1$, MCP, and SCAD. Table \ref{tab:cifar100_scale} shows that more than half of the scaling factors are less than $10^{-6}$ in magnitudes in DenseNet-40 for most regularizers, but they are more than $10^{-6}$ in magnitudes in VGG-19 and ResNet-164 for all regularizers. The distributions of the scaling factors convey why DenseNet-40 can be pruned at higher channel pruning ratios than VGG-19 and ResNet-164, as indicated by the middle rows of Figures \ref{fig:vgg_result}-\ref{fig:resnet}. Across the three networks, MCP and SCAD have similar distributions with $\ell_1$. For $\ell_{3/4}$, a considerable amount of scaling factors are within the interval $(10^{-6}, 10^{-5})$, but as $p$ decreases, the magnitudes of most scaling factors increase. Hence, less than a few hundred scaling factors are below $10^{-6}$ in magnitudes. As a result, models regularized with $\ell_{1/2}$ and $\ell_{1/4}$ become more sensitive to channel pruning as demonstrated earlier. For T$\ell_1 (a=10.0)$, its distribution of scaling factors is similar to $\ell_1$. However, when $a=0.5, 1.0$, more scaling factors have magnitudes less than $10^{-5}$, which demonstrates T$\ell_1$'s robustness to channel pruning when $a$ is small enough.  

Figures \ref{fig:vgg_scaling_SVHN}-\ref{fig:resnet_scaling_SVHN} and Table \ref{tab:SVHN_scale} provide statistics about SVHN. For all three networks, T$\ell_1 (a=10.0)$, SCAD, and MCP have similar distributions as $\ell_1$. Similar to CIFAR 10 and 100, $\ell_{3/4}$ has most of its scaling factors to be in the interval $(10^{-6}, 10^{-5})$, but as $p$ decreases for $\ell_p$, the magnitudes of the scaling factors increase, resulting in at least 90\% of the scaling factors to be at least $10^{-6}$ in magnitudes as shown in Table \ref{tab:SVHN_scale}. For T$\ell_1 (a=0.5, 1.0)$ on the other hand, most of the scaling factors are in the interval $(10^{-7}, 10^{-6})$.  

\subsection{Comparison with Variational CNN Pruning}
\begin{table*}
    \centering
        \caption{Comparisons between network slimming with T$\ell_1 (a=0.5,1.0)$ and variational channel pruning. The results are immediately obtained after channel pruning.}
    \label{tab:comparison}
    \scriptsize
    \begin{tabular}{c|c|c|c|c|c}
    Model & Dataset & Method & Test Accuracy & Percentage of Channels Pruned & Percentage of Parameters Pruned\\ \hline
    \multirow{10}{*}{DenseNet-40} & \multirow{5}{*}{CIFAR 10} & VCP \cite{zhao2019variational} & 93.16\% & 60\% & 59.67\%\\
    & & T$\ell_1 (a=1.0)$ & 93.17\% & 60\% & 55.73\% \\
     & & T$\ell_1 (a=1.0)$ & 93.17\% & 80\% & 74.46\%  \\
         & & T$\ell_1 (a=0.5)$ & 92.78\% & 60\% & 56.16\% \\
     & & T$\ell_1 (a=0.5)$ & 92.78\% & 80\% & 74.88\% \\
     \cline{2-6}
     & \multirow{5}{*}{CIFAR 100} & VCP \cite{zhao2019variational} & 72.19\% & 37\% & 37.73\%\\
    & & T$\ell_1 (a=1.0)$ & 72.63\% & 40\% & 36.91\% \\
     & & T$\ell_1 (a=1.0)$ & 72.63\% & 60\% & 55.35\%  \\
         & & T$\ell_1 (a=0.5)$ & 72.57\% & 40\% & 36.98\% \\
     & & T$\ell_1 (a=0.5)$ & 72.58\% & 60\% & 55.46\% \\
     \hline
         \multirow{5}{*}{ResNet-164} & \multirow{2}{*}{CIFAR 10} & VCP \cite{zhao2019variational} & 93.16\% & 74\% & 56.70\%\\
     & & T$\ell_1 (a=0.5)$ & 93.41\% & 75\% & 70.39\% \\
     \cline{2-6}
     & \multirow{3}{*}{CIFAR 100} & VCP \cite{zhao2019variational} & 73.76\% & 47\% & 17.59\%\\
     & & T$\ell_1 (a=1.0)$ & 74.89\% & 45\% & 25.56\% \\
     & & T$\ell_1 (a=0.5)$ & 74.72\% & 45\% & 27.74\% \\
     \hline
    \end{tabular}
\end{table*}

We have shown that network slimming with nonconvex regularizers can outperform the original with $\ell_1$ regularization. Now we compare our proposed method with variational CNN pruning (VCP) proposed in \cite{zhao2019variational}, a Bayesian version of network slimming. VCP is designed to be robust against channel pruning, so we compare it with T$\ell_1 (a=0.5,1.0)$, which is proven to also be robust against channel pruning in our earlier analyses. The comparisons between the two methods are shown in Table \ref{tab:comparison}, using results from DenseNet-40 and ResNet-164 trained on CIFAR 10/100. 

For DenseNet-40 trained on CIFAR 10, T$\ell_1$ has a minimally better accuracy with less parameters pruned than VCP with 60\% channels pruned. However, we can increase the percentage of channels pruned to 80\% for T$\ell_1$ so that the number of parameters are reduced while maintaining the same accuracy. On CIFAR 100, with similar percentages of channels pruned, T$\ell_1$ has a much better accuracy than VCP but again with less parameters pruned. Nevertheless, we can increase the percentage of channels pruned to 60\% and the accuracy will remain the same with more parameters pruned. 

On ResNet-164, with similar percentages of channels pruned, T$\ell_1$ outperforms VCP by a large margin for both test accuracy and percentage of parameters pruned. For CIFAR 10, only T$\ell_1 (a=0.5)$ is able to have 75\% of the channels pruned, and it saves more parameters by almost 24\% with test accuracy better by 0.25\%. For CIFAR 100\%, with 2\% less channels pruned, T$\ell_1$ prunes at least 7.97\% more parameters than VCP while having better accuracy of at least 0.96\%. 

Overall, network slimming with T$\ell_1$ is competitive against the latest variant of network slimming.

\section{Conclusion}
We improve network slimming by replacing the $\ell_1$ regularizer with a sparse, nonconvex regularizer for penalizing the scaling factors in the batch normalization layers. In particular, we investigate $\ell_p (0 < p < 1)$, T$\ell_1$, MCP, and SCAD. We apply the proposed methods onto VGG-19, DenseNet-40, and ResNet-164 trained on CIFAR 10/100 and SVHN. We observe that $\ell_p$ and T$\ell_1$ save more on parameters and FLOPs than $\ell_1$ with a slight decrease in test accuracy. In addition, T$\ell_1$, especially $a=0.5$, preserves model accuracy against channel pruning. Network slimming with T$\ell_1$ is competitive against VCP, another network slimming variant robust against channel pruning. To attain better accuracy than $\ell_1$ while having similar compression, MCP and SCAD perform the best job after their models are pruned and retrained, especially for VGG-19 and DenseNet-40.

For future directions, we plan to develop an optimization algorithm based on the relaxed variable splitting method \cite{dinh2020convergence} in order to use other nonconvex regularizers such as $\ell_1 - \alpha \ell_2$ \cite{lou-2015-cs, yin2015minimization}. Additionally, we aim to generalize nonconvex network slimming to layer normalization \cite{ba2016layer} and group normalization \cite{wu2018group}. Last, we will adapt nonconvex network slimming to the state-of-the-art compact CNNs, such as MobileNetv2 \cite{sandler2018mobilenetv2} and ShuffleNet \cite{zhang2018shufflenet}. 

\section*{Acknowledgment}
The authors would like to thank the associate editor and the two anonymous referees for their careful reading and helpful feedback, which improved the presentation of the paper.

 \bibliographystyle{IEEEtran}
\bibliography{egbib}

\begin{thebibliography}{10}
\providecommand{\url}[1]{#1}
\csname url@samestyle\endcsname
\providecommand{\newblock}{\relax}
\providecommand{\bibinfo}[2]{#2}
\providecommand{\BIBentrySTDinterwordspacing}{\spaceskip=0pt\relax}
\providecommand{\BIBentryALTinterwordstretchfactor}{4}
\providecommand{\BIBentryALTinterwordspacing}{\spaceskip=\fontdimen2\font plus
\BIBentryALTinterwordstretchfactor\fontdimen3\font minus
  \fontdimen4\font\relax}
\providecommand{\BIBforeignlanguage}[2]{{%
\expandafter\ifx\csname l@#1\endcsname\relax
\typeout{** WARNING: IEEEtran.bst: No hyphenation pattern has been}%
\typeout{** loaded for the language `#1'. Using the pattern for}%
\typeout{** the default language instead.}%
\else
\language=\csname l@#1\endcsname
\fi
#2}}
\providecommand{\BIBdecl}{\relax}
\BIBdecl

\bibitem{he2016deep}
K.~He, X.~Zhang, S.~Ren, and J.~Sun, ``Deep residual learning for image
  recognition,'' in \emph{Proceedings of the IEEE Conference on Computer Vision
  and Pattern Recognition}, 2016, pp. 770--778.

\bibitem{krizhevsky2012imagenet}
A.~Krizhevsky, I.~Sutskever, and G.~E. Hinton, ``Imagenet classification with
  deep convolutional neural networks,'' in \emph{Advances in Neural Information
  Processing Systems}, 2012, pp. 1097--1105.

\bibitem{simonyan2014very}
K.~Simonyan and A.~Zisserman, ``Very deep convolutional networks for
  large-scale image recognition,'' \emph{arXiv preprint arXiv:1409.1556}, 2014.

\bibitem{chen2017deeplab}
L.-C. Chen, G.~Papandreou, I.~Kokkinos, K.~Murphy, and A.~L. Yuille, ``Deeplab:
  Semantic image segmentation with deep convolutional nets, atrous convolution,
  and fully connected crfs,'' \emph{IEEE Transactions on Pattern Analysis and
  Machine Intelligence}, vol.~40, no.~4, pp. 834--848, 2017.

\bibitem{long2015fully}
J.~Long, E.~Shelhamer, and T.~Darrell, ``Fully convolutional networks for
  semantic segmentation,'' in \emph{Proceedings of the IEEE Conference on
  Computer Vision and Pattern Recognition}, 2015, pp. 3431--3440.

\bibitem{ronneberger2015u}
O.~Ronneberger, P.~Fischer, and T.~Brox, ``U-net: Convolutional networks for
  biomedical image segmentation,'' in \emph{International Conference on Medical
  image computing and Computer-Assisted Intervention}.\hskip 1em plus 0.5em
  minus 0.4em\relax Springer, 2015, pp. 234--241.

\bibitem{girshick2014rich}
R.~Girshick, J.~Donahue, T.~Darrell, and J.~Malik, ``Rich feature hierarchies
  for accurate object detection and semantic segmentation,'' in
  \emph{Proceedings of the IEEE Conference on Computer Vision and Pattern
  Recognition}, 2014, pp. 580--587.

\bibitem{huang2017speed}
J.~Huang, V.~Rathod, C.~Sun, M.~Zhu, A.~Korattikara, A.~Fathi, I.~Fischer,
  Z.~Wojna, Y.~Song, S.~Guadarrama \emph{et~al.}, ``Speed/accuracy trade-offs
  for modern convolutional object detectors,'' in \emph{Proceedings of the IEEE
  Conference on Computer Vision and Pattern Recognition}, 2017, pp. 7310--7311.

\bibitem{ren2015faster}
S.~Ren, K.~He, R.~Girshick, and J.~Sun, ``Faster {R-CNN}: Towards real-time
  object detection with region proposal networks,'' in \emph{Advances in Neural
  Information Processing Systems}, 2015, pp. 91--99.

\bibitem{denton2014exploiting}
E.~L. Denton, W.~Zaremba, J.~Bruna, Y.~LeCun, and R.~Fergus, ``Exploiting
  linear structure within convolutional networks for efficient evaluation,'' in
  \emph{Advances in Neural Information Processing Systems}, 2014, pp.
  1269--1277.

\bibitem{jaderberg2014speeding}
M.~Jaderberg, A.~Vedaldi, and A.~Zisserman, ``Speeding up convolutional neural
  networks with low rank expansions,'' \emph{arXiv preprint arXiv:1405.3866},
  2014.

\bibitem{wen2017coordinating}
W.~Wen, C.~Xu, C.~Wu, Y.~Wang, Y.~Chen, and H.~Li, ``Coordinating filters for
  faster deep neural networks,'' in \emph{Proceedings of the IEEE Conference on
  Computer Vision and Pattern Recognition}, 2017, pp. 658--666.

\bibitem{xu2018trained}
Y.~Xu, Y.~Li, S.~Zhang, W.~Wen, B.~Wang, Y.~Qi, Y.~Chen, W.~Lin, and H.~Xiong,
  ``Trained rank pruning for efficient deep neural networks,'' \emph{arXiv
  preprint arXiv:1812.02402}, 2018.

\bibitem{xu2020trp}
------, ``{TRP}: Trained rank pruning for efficient deep neural networks,''
  \emph{International Joint Conference on Artificial Intelligence}, 2020.

\bibitem{chen2015compressing}
W.~Chen, J.~Wilson, S.~Tyree, K.~Weinberger, and Y.~Chen, ``Compressing neural
  networks with the hashing trick,'' in \emph{International Conference on
  Machine Learning}, 2015, pp. 2285--2294.

\bibitem{courbariaux2015binaryconnect}
M.~Courbariaux, Y.~Bengio, and J.-P. David, ``Binaryconnect: Training deep
  neural networks with binary weights during propagations,'' in \emph{Advances
  in Neural Information Processing Systems}, 2015, pp. 3123--3131.

\bibitem{li2016ternary}
F.~Li, B.~Zhang, and B.~Liu, ``Ternary weight networks,'' \emph{arXiv preprint
  arXiv:1605.04711}, 2016.

\bibitem{zhu2016trained}
C.~Zhu, S.~Han, H.~Mao, and W.~J. Dally, ``Trained ternary quantization,''
  \emph{arXiv preprint arXiv:1612.01064}, 2016.

\bibitem{yin2018binaryrelax}
P.~Yin, S.~Zhang, J.~Lyu, S.~Osher, Y.~Qi, and J.~Xin, ``Binaryrelax: A
  relaxation approach for training deep neural networks with quantized
  weights,'' \emph{SIAM Journal on Imaging Sciences}, vol.~11, no.~4, pp.
  2205--2223, 2018.

\bibitem{aghasi2017net}
A.~Aghasi, A.~Abdi, N.~Nguyen, and J.~Romberg, ``Net-trim: Convex pruning of
  deep neural networks with performance guarantee,'' in \emph{Advances in
  Neural Information Processing Systems}, 2017, pp. 3177--3186.

\bibitem{han2015learning}
S.~Han, J.~Pool, J.~Tran, and W.~Dally, ``Learning both weights and connections
  for efficient neural network,'' in \emph{Advances in Neural Information
  Processing Systems}, 2015, pp. 1135--1143.

\bibitem{li2016pruning}
H.~Li, A.~Kadav, I.~Durdanovic, H.~Samet, and H.~P. Graf, ``Pruning filters for
  efficient convnets,'' \emph{arXiv preprint arXiv:1608.08710}, 2016.

\bibitem{hu2016network}
H.~Hu, R.~Peng, Y.-W. Tai, and C.-K. Tang, ``Network trimming: A data-driven
  neuron pruning approach towards efficient deep architectures,'' \emph{arXiv
  preprint arXiv:1607.03250}, 2016.

\bibitem{alvarez2016learning}
J.~M. Alvarez and M.~Salzmann, ``Learning the number of neurons in deep
  networks,'' in \emph{Advances in Neural Information Processing Systems},
  2016, pp. 2270--2278.

\bibitem{changpinyo2017power}
S.~Changpinyo, M.~Sandler, and A.~Zhmoginov, ``The power of sparsity in
  convolutional neural networks,'' \emph{arXiv preprint arXiv:1702.06257},
  2017.

\bibitem{scardapane2017group}
S.~Scardapane, D.~Comminiello, A.~Hussain, and A.~Uncini, ``Group sparse
  regularization for deep neural networks,'' \emph{Neurocomputing}, vol. 241,
  pp. 81--89, 2017.

\bibitem{wen2016learning}
W.~Wen, C.~Wu, Y.~Wang, Y.~Chen, and H.~Li, ``Learning structured sparsity in
  deep neural networks,'' in \emph{Advances in Neural Information Processing
  Systems}, 2016, pp. 2074--2082.

\bibitem{liu2017learning}
Z.~Liu, J.~Li, Z.~Shen, G.~Huang, S.~Yan, and C.~Zhang, ``Learning efficient
  convolutional networks through network slimming,'' in \emph{Proceedings of
  the IEEE Conference on Computer Vision and Pattern Recognition}, 2017, pp.
  2736--2744.

\bibitem{zhang2020lightweight}
J.~Zhang, W.~Wang, C.~Lu, J.~Wang, and A.~K. Sangaiah, ``Lightweight deep
  network for traffic sign classification,'' \emph{Annals of
  Telecommunications}, vol.~75, no.~7, pp. 369--379, 2020.

\bibitem{ma2021lightweight}
H.~Ma, T.~Celik, and H.-C. Li, ``Lightweight attention convolutional neural
  network through network slimming for robust facial expression recognition,''
  \emph{Signal, Image and Video Processing}, pp. 1--9, 2021.

\bibitem{he2021cap}
W.~He, M.~Wu, M.~Liang, and S.-K. Lam, ``{CAP}: Context-aware pruning for
  semantic segmentation,'' in \emph{Proceedings of the IEEE/CVF Winter
  Conference on Applications of Computer Vision}, 2021, pp. 960--969.

\bibitem{chartrand2007exact}
R.~Chartrand, ``Exact reconstruction of sparse signals via nonconvex
  minimization,'' \emph{IEEE Signal Processing Letters}, vol.~14, no.~10, pp.
  707--710, 2007.

\bibitem{chartrand2008iteratively}
R.~Chartrand and W.~Yin, ``Iteratively reweighted algorithms for compressive
  sensing,'' in \emph{2008 IEEE International Conference on Acoustics, Speech
  and Signal Processing}.\hskip 1em plus 0.5em minus 0.4em\relax IEEE, 2008,
  pp. 3869--3872.

\bibitem{xu2012l}
Z.~Xu, X.~Chang, F.~Xu, and H.~Zhang, ``${\ell_{1/2}}$ regularization: A
  thresholding representation theory and a fast solver,'' \emph{IEEE
  Transactions on Neural Networks and Learning Systems}, vol.~23, no.~7, pp.
  1013--1027, 2012.

\bibitem{zhang2014minimization}
S.~Zhang and J.~Xin, ``Minimization of transformed $ l_1 $ penalty: Closed form
  representation and iterative thresholding algorithms,'' \emph{Communications
  in Mathematical Sciences}, vol.~15, no.~2, p. 511 – 537, 2017.

\bibitem{zhang2018minimization}
------, ``Minimization of transformed $l_1$ penalty: theory, difference of
  convex function algorithm, and robust application in compressed sensing,''
  \emph{Mathematical Programming}, vol. 169, no.~1, pp. 307--336, 2018.

\bibitem{zhang2010nearly}
C.-H. Zhang, ``Nearly unbiased variable selection under minimax concave
  penalty,'' \emph{The Annals of Statistics}, vol.~38, no.~2, pp. 894--942,
  2010.

\bibitem{fan2001variable}
J.~Fan and R.~Li, ``Variable selection via nonconcave penalized likelihood and
  its oracle properties,'' \emph{Journal of the American Statistical
  Association}, vol.~96, no. 456, pp. 1348--1360, 2001.

\bibitem{shor2012minimization}
N.~Z. Shor, \emph{Minimization methods for non-differentiable functions}.\hskip
  1em plus 0.5em minus 0.4em\relax Springer Science \& Business Media, 2012,
  vol.~3.

\bibitem{bui2020nonconvex}
K.~Bui, F.~Park, S.~Zhang, Y.~Qi, and J.~Xin, ``Nonconvex regularization for
  network slimming: Compressing {CNNs} even more,'' in \emph{International
  Symposium on Visual Computing}.\hskip 1em plus 0.5em minus 0.4em\relax
  Springer, 2020, pp. 39--53.

\bibitem{tai2015convolutional}
C.~Tai, T.~Xiao, Y.~Zhang, X.~Wang \emph{et~al.}, ``Convolutional neural
  networks with low-rank regularization,'' \emph{arXiv preprint
  arXiv:1511.06067}, 2015.

\bibitem{bai2018proxquant}
Y.~Bai, Y.-X. Wang, and E.~Liberty, ``Proxquant: Quantized neural networks via
  proximal operators,'' \emph{arXiv preprint arXiv:1810.00861}, 2018.

\bibitem{aghasi2020fast}
A.~Aghasi, A.~Abdi, and J.~Romberg, ``Fast convex pruning of deep neural
  networks,'' \emph{SIAM Journal on Mathematics of Data Science}, vol.~2,
  no.~1, pp. 158--188, 2020.

\bibitem{zhao2019variational}
C.~Zhao, B.~Ni, J.~Zhang, Q.~Zhao, W.~Zhang, and Q.~Tian, ``Variational
  convolutional neural network pruning,'' in \emph{Proceedings of the IEEE/CVF
  Conference on Computer Vision and Pattern Recognition}, 2019, pp. 2780--2789.

\bibitem{yuan2006model}
M.~Yuan and Y.~Lin, ``Model selection and estimation in regression with grouped
  variables,'' \emph{Journal of the Royal Statistical Society: Series B
  (Statistical Methodology)}, vol.~68, no.~1, pp. 49--67, 2006.

\bibitem{xue2019learning}
F.~Xue and J.~Xin, ``Learning sparse neural networks via {$\ell_0$} and
  {$\text{T}\ell_1$} by a relaxed variable splitting method with application to
  multi-scale curve classification,'' in \emph{World Congress on Global
  Optimization}.\hskip 1em plus 0.5em minus 0.4em\relax Springer, 2019, pp.
  800--809.

\bibitem{ma2019transformed}
R.~Ma, J.~Miao, L.~Niu, and P.~Zhang, ``Transformed {$\ell_1$} regularization
  for learning sparse deep neural networks,'' \emph{Neural Networks}, vol. 119,
  pp. 286--298, 2019.

\bibitem{pandit2021learning}
M.~K. Pandit, R.~Naaz, and M.~A. Chishti, ``Learning sparse neural networks
  using non-convex regularization,'' \emph{IEEE Transactions on Emerging Topics
  in Computational Intelligence}, 2021.

\bibitem{bui2021structured}
K.~Bui, F.~Park, S.~Zhang, Y.~Qi, and J.~Xin, ``Structured sparsity of
  convolutional neural networks via nonconvex sparse group regularization,''
  \emph{Frontiers in Applied Mathematics and Statistics}, 2021.

\bibitem{Li_2020_CVPR}
Y.~Li, S.~Gu, C.~Mayer, L.~V. Gool, and R.~Timofte, ``Group sparsity: The hinge
  between filter pruning and decomposition for network compression,'' in
  \emph{Proceedings of the IEEE/CVF Conference on Computer Vision and Pattern
  Recognition (CVPR)}, June 2020.

\bibitem{candes-2006}
E.~J. Cand{\`e}s, J.~K. Romberg, and T.~Tao, ``Stable signal recovery from
  incomplete and inaccurate measurements,'' \emph{Communications on Pure and
  Applied Mathematics}, vol.~59, no.~8, pp. 1207--1223, 2006.

\bibitem{candes2006robust}
E.~J. Cand{\`e}s, J.~Romberg, and T.~Tao, ``Robust uncertainty principles:
  Exact signal reconstruction from highly incomplete frequency information,''
  \emph{IEEE Transactions on Information Theory}, vol.~52, no.~2, pp. 489--509,
  2006.

\bibitem{yin2008bregman}
W.~Yin, S.~Osher, D.~Goldfarb, and J.~Darbon, ``Bregman iterative algorithms
  for $\ell_1$-minimization with applications to compressed sensing,''
  \emph{SIAM Journal on Imaging sciences}, vol.~1, no.~1, pp. 143--168, 2008.

\bibitem{jung2007improved}
H.~Jung, J.~C. Ye, and E.~Y. Kim, ``Improved k--t blast and k--t sense using
  focuss,'' \emph{Physics in Medicine \& Biology}, vol.~52, no.~11, p. 3201,
  2007.

\bibitem{lustig2007sparse}
M.~Lustig, D.~Donoho, and J.~M. Pauly, ``Sparse {MRI}: The application of
  compressed sensing for rapid mr imaging,'' \emph{Magnetic Resonance in
  Medicine: An Official Journal of the International Society for Magnetic
  Resonance in Medicine}, vol.~58, no.~6, pp. 1182--1195, 2007.

\bibitem{lou-2015-cs}
Y.~Lou, P.~Yin, Q.~He, and J.~Xin, ``Computing sparse representation in a
  highly coherent dictionary based on difference of {$L_1$} and {$L_2$},''
  \emph{Journal of Scientific Computing}, vol.~64, no.~1, pp. 178--196, 2015.

\bibitem{lou2015computational}
Y.~Lou, S.~Osher, and J.~Xin, ``Computational aspects of constrained
  {$L_1-L_2$} minimization for compressive sensing,'' in \emph{Modelling,
  Computation and Optimization in Information Systems and Management
  Sciences}.\hskip 1em plus 0.5em minus 0.4em\relax Springer, 2015, pp.
  169--180.

\bibitem{chartrand2008restricted}
R.~Chartrand and V.~Staneva, ``Restricted isometry properties and nonconvex
  compressive sensing,'' \emph{Inverse Problems}, vol.~24, no.~3, p. 035020,
  2008.

\bibitem{zong2012representative}
Z.~Xu, H.~Guo, Y.~Wang, and Z.~Hai, ``Representative of {$L_{1/2}$}
  regularization among {$L_q (0 \leq q \leq 1)$} regularizations: an
  experimental study based on phase diagram,'' \emph{Acta Automatica Sinica},
  vol.~38, no.~7, pp. 1225--1228, 2012.

\bibitem{krishnan2009fast}
D.~Krishnan and R.~Fergus, ``Fast image deconvolution using hyper-laplacian
  priors,'' in \emph{Advances in Neural Information Processing Systems}, 2009,
  pp. 1033--1041.

\bibitem{cao2013fast}
W.~Cao, J.~Sun, and Z.~Xu, ``Fast image deconvolution using closed-form
  thresholding formulas of {$L_q (q= 1/2, 2/3)$} regularization,''
  \emph{Journal of Visual Communication and Image Representation}, vol.~24,
  no.~1, pp. 31--41, 2013.

\bibitem{qian2011hyperspectral}
Y.~Qian, S.~Jia, J.~Zhou, and A.~Robles-Kelly, ``Hyperspectral unmixing via
  {$L_{1/2}$} sparsity-constrained nonnegative matrix factorization,''
  \emph{IEEE Transactions on Geoscience and Remote Sensing}, vol.~49, no.~11,
  pp. 4282--4297, 2011.

\bibitem{miao2015general}
C.~Miao and H.~Yu, ``A general-thresholding solution for $l_p (0< p< 1)$
  regularized {CT} reconstruction,'' \emph{IEEE Transactions on Image
  Processing}, vol.~24, no.~12, pp. 5455--5468, 2015.

\bibitem{li2020tv}
Y.~Li, C.~Wu, and Y.~Duan, ``The {$\text{TV}p$} regularized {M}umford-{S}hah
  model for image labeling and segmentation,'' \emph{IEEE Transactions on Image
  Processing}, vol.~29, pp. 7061--7075, 2020.

\bibitem{wu2021two}
T.~Wu, J.~Shao, X.~Gu, M.~K. Ng, and T.~Zeng, ``Two-stage image segmentation
  based on nonconvex $\ell_2-\ell_p$ approximation and thresholding,''
  \emph{Applied Mathematics and Computation}, vol. 403, p. 126168, 2021.

\bibitem{fan2004nonconcave}
J.~Fan, H.~Peng \emph{et~al.}, ``Nonconcave penalized likelihood with a
  diverging number of parameters,'' \emph{The Annals of Statistics}, vol.~32,
  no.~3, pp. 928--961, 2004.

\bibitem{lv2009unified}
J.~Lv, Y.~Fan \emph{et~al.}, ``A unified approach to model selection and sparse
  recovery using regularized least squares,'' \emph{The Annals of Statistics},
  vol.~37, no.~6A, pp. 3498--3528, 2009.

\bibitem{zhang2015transformed}
S.~Zhang, P.~Yin, and J.~Xin, ``Transformed {Schatten}-1 iterative thresholding
  algorithms for low rank matrix completion,'' \emph{Communications in
  Mathematical Sciences}, vol.~15, no.~3, p. 839 – 862, 2017.

\bibitem{you2019nonconvex}
J.~You, Y.~Jiao, X.~Lu, and T.~Zeng, ``A nonconvex model with minimax concave
  penalty for image restoration,'' \emph{Journal of Scientific Computing},
  vol.~78, no.~2, pp. 1063--1086, 2019.

\bibitem{jin2016alternating}
Z.-F. Jin, Z.~Wan, Y.~Jiao, and X.~Lu, ``An alternating direction method with
  continuation for nonconvex low rank minimization,'' \emph{Journal of
  Scientific Computing}, vol.~66, no.~2, pp. 849--869, 2016.

\bibitem{mehranian2013smoothly}
A.~Mehranian, H.~S. Rad, A.~Rahmim, M.~R. Ay, and H.~Zaidi, ``Smoothly clipped
  absolute deviation ({SCAD}) regularization for compressed sensing {MRI} using
  an augmented {L}agrangian scheme,'' \emph{Magnetic Resonance Imaging},
  vol.~31, no.~8, pp. 1399--1411, 2013.

\bibitem{breheny2011coordinate}
P.~Breheny and J.~Huang, ``Coordinate descent algorithms for nonconvex
  penalized regression, with applications to biological feature selection,''
  \emph{The Annals of Applied Statistics}, vol.~5, no.~1, p. 232, 2011.

\bibitem{wang2007group}
L.~Wang, G.~Chen, and H.~Li, ``Group {SCAD} regression analysis for microarray
  time course gene expression data,'' \emph{Bioinformatics}, vol.~23, no.~12,
  pp. 1486--1494, 2007.

\bibitem{gu2017tvscad}
G.~Gu, S.~Jiang, and J.~Yang, ``A {TVSCAD} approach for image deblurring with
  impulsive noise,'' \emph{Inverse Problems}, vol.~33, no.~12, p. 125008, 2017.

\bibitem{antoniadis2001regularization}
A.~Antoniadis and J.~Fan, ``Regularization of wavelet approximations,''
  \emph{Journal of the American Statistical Association}, vol.~96, no. 455, pp.
  939--967, 2001.

\bibitem{ahn2017difference}
M.~Ahn, J.-S. Pang, and J.~Xin, ``Difference-of-convex learning: directional
  stationarity, optimality, and sparsity,'' \emph{SIAM Journal on
  Optimization}, vol.~27, no.~3, pp. 1637--1665, 2017.

\bibitem{wen2018survey}
F.~Wen, L.~Chu, P.~Liu, and R.~C. Qiu, ``A survey on nonconvex
  regularization-based sparse and low-rank recovery in signal processing,
  statistics, and machine learning,'' \emph{IEEE Access}, vol.~6, pp.
  69\,883--69\,906, 2018.

\bibitem{ioffe2015batch}
S.~Ioffe and C.~Szegedy, ``Batch normalization: Accelerating deep network
  training by reducing internal covariate shift,'' in \emph{International
  Conference on Machine Learning}, 2015, pp. 448--456.

\bibitem{szegedy2016rethinking}
C.~Szegedy, V.~Vanhoucke, S.~Ioffe, J.~Shlens, and Z.~Wojna, ``Rethinking the
  inception architecture for computer vision,'' in \emph{Proceedings of the
  IEEE Conference on Computer Vision and Pattern Recognition}, 2016, pp.
  2818--2826.

\bibitem{penot2012calculus}
J.-P. Penot, \emph{Calculus without derivatives}.\hskip 1em plus 0.5em minus
  0.4em\relax Springer Science \& Business Media, 2012, vol. 266.

\bibitem{krizhevsky2009learning}
A.~Krizhevsky, G.~Hinton \emph{et~al.}, ``Learning multiple layers of features
  from tiny images,'' University of Toronto, Tech. Rep., 2009.

\bibitem{huang2016deep}
G.~Huang, Y.~Sun, Z.~Liu, D.~Sedra, and K.~Q. Weinberger, ``Deep networks with
  stochastic depth,'' in \emph{European Conference on Computer Vision}.\hskip
  1em plus 0.5em minus 0.4em\relax Springer, 2016, pp. 646--661.

\bibitem{lin2013network}
M.~Lin, Q.~Chen, and S.~Yan, ``Network in network,'' \emph{arXiv preprint
  arXiv:1312.4400}, 2013.

\bibitem{goodfellow2013maxout}
I.~Goodfellow, D.~Warde-Farley, M.~Mirza, A.~Courville, and Y.~Bengio, ``Maxout
  networks,'' in \emph{International Conference on Machine Learning}.\hskip 1em
  plus 0.5em minus 0.4em\relax PMLR, 2013, pp. 1319--1327.

\bibitem{Netzer2011ReadingDI}
Y.~Netzer, T.~Wang, A.~Coates, A.~Bissacco, B.~Wu, and A.~Ng, ``Reading digits
  in natural images with unsupervised feature learning,'' in \emph{NIPS
  Workshop on Deep Learning and Unsupervised Feature Learning}, 2011.

\bibitem{huang2017densely}
G.~Huang, Z.~Liu, L.~Van Der~Maaten, and K.~Q. Weinberger, ``Densely connected
  convolutional networks,'' in \emph{Proceedings of the IEEE Conference on
  Computer Vision and Pattern Recognition}, 2017, pp. 4700--4708.

\bibitem{sutskever2013importance}
I.~Sutskever, J.~Martens, G.~Dahl, and G.~Hinton, ``On the importance of
  initialization and momentum in deep learning,'' in \emph{International
  Conference on Machine Learning}, 2013, pp. 1139--1147.

\bibitem{he2015delving}
K.~He, X.~Zhang, S.~Ren, and J.~Sun, ``Delving deep into rectifiers: Surpassing
  human-level performance on imagenet classification,'' in \emph{Proceedings of
  the IEEE Conference on Computer Vision and Pattern Recognition}, 2015, pp.
  1026--1034.

\bibitem{aggarwal2018neural}
C.~C. Aggarwal, \emph{Neural networks and deep learning}.\hskip 1em plus 0.5em
  minus 0.4em\relax Springer, 2018.

\bibitem{dinh2020convergence}
T.~Dinh and J.~Xin, ``Convergence of a relaxed variable splitting method for
  learning sparse neural networks via $\ell_1$,$\ell_0$, and
  transformed-$\ell_1$ penalties,'' in \emph{Proceedings of SAI Intelligent
  Systems Conference}.\hskip 1em plus 0.5em minus 0.4em\relax Springer, 2020,
  pp. 360--374.

\bibitem{yin2015minimization}
P.~Yin, Y.~Lou, Q.~He, and J.~Xin, ``Minimization of {$\ell_{1-2}$} for
  compressed sensing,'' \emph{SIAM Journal on Scientific Computing}, vol.~37,
  no.~1, pp. A536--A563, 2015.

\bibitem{ba2016layer}
J.~L. Ba, J.~R. Kiros, and G.~E. Hinton, ``Layer normalization,'' \emph{arXiv
  preprint arXiv:1607.06450}, 2016.

\bibitem{wu2018group}
Y.~Wu and K.~He, ``Group normalization,'' in \emph{Proceedings of the European
  Conference on Computer Vision (ECCV)}, 2018, pp. 3--19.

\bibitem{sandler2018mobilenetv2}
M.~Sandler, A.~Howard, M.~Zhu, A.~Zhmoginov, and L.-C. Chen, ``Mobilenetv2:
  Inverted residuals and linear bottlenecks,'' in \emph{Proceedings of the IEEE
  Conference on Computer Vision and Pattern Recognition}, 2018, pp. 4510--4520.

\bibitem{zhang2018shufflenet}
X.~Zhang, X.~Zhou, M.~Lin, and J.~Sun, ``Shufflenet: An extremely efficient
  convolutional neural network for mobile devices,'' in \emph{Proceedings of
  the IEEE Conference on Computer Vision and Pattern Recognition}, 2018, pp.
  6848--6856.

\end{thebibliography}

\end{document}